\documentclass[letterpaper, 10 pt, conference]{ieeeconf} 
\IEEEoverridecommandlockouts    % This command is only needed if 
\usepackage{cite}
\usepackage{amsmath,amssymb,amsfonts}
\usepackage{siunitx}
\usepackage{graphicx}
\usepackage{textcomp}
\usepackage{subfigure}
\usepackage{array}

\DeclareMathOperator*{\argmin}{arg\,min}

\newtheorem{note}{Note}

\begin{document}
    \pagestyle{empty}

    \title{Continuous-time Trajectory Estimation: A Comparative Study Between Gaussian Process and Spline-based Approaches}
    \author{Jacob C. Johnson${}^1$, Joshua G. Mangelson${}^1$, Timothy D. Barfoot${}^2$, Randal W. Beard${}^1$
    \thanks{This work has been funded by the Center for Autonomous Air Mobility and Sensing (CAAMS), a National Science Foundation Industry/University Cooperative Research Center (I/UCRC) under NSF award No. 2139551, along with significant contributions from CAAMS industry members.}
    \thanks{${}^1$J. C. Johnson, J. G. Mangelson, and R. W. Beard are with the Electrical and Computer Engineering Department at Brigham Young University, Provo, UT, 84602 USA.  (e-mail: jjohns99@byu.edu).}
    \thanks{${}^2$T. D. Barfoot is with the University of Toronto Institute for Aerospace Studies, Toronto, ON, CA.}}
    
    \maketitle
    \thispagestyle{empty}

    \begin{abstract}
        Continuous-time trajectory estimation is an attractive alternative to discrete-time batch estimation due to the ability to incorporate high-frequency measurements from asynchronous sensors while keeping the number of optimization parameters bounded. Two types of continuous-time estimation have become prevalent in the literature: Gaussian process regression and spline-based estimation. In this paper, we present a direct comparison between these two methods. We first compare them using a simple linear system, and then compare them in a camera and IMU sensor fusion scenario on SE(3) in both simulation and hardware. Our results show that if the same measurements and motion model are used, the two methods achieve similar trajectory accuracy. In addition, if the spline order is chosen so that the degree-of-differentiability of the two trajectory representations match, then they achieve similar solve times as well.
    \end{abstract}

    \section{Introduction}

State estimation is an essential area of robotics. Without an accurate estimate of a mobile robot's dynamic state, other autonomy tasks such as online path planning and control are not possible. Traditionally, state estimation has most often been accomplished using some variant of the well-known Kalman filter (KF)~\cite{Kalman1960, Smith1962, Jazwinski1972, Julier2004}. More recently, however, discrete-time batch estimation methods such as bundle adjustment~\cite{Triggs2000} have gained a significant amount of popularity~\cite{Dellaert2012, Murartal2015, Qin2018}. In these methods, a discrete set of past robot states is estimated simultaneously using nonlinear optimization. Batch estimation has been shown to offer better performance than traditional filtering~\cite{Strasdat2012}, in large part because the nonlinearities in the measurement models are iteratively relinearized about the current best guess rather than linearizing only once. One drawback of discrete-time batch estimation, however, is that it is difficult to include measurements from multiple sensors that are asynchronous or that come at high frequencies. This is because each measurement used in the optimization must be tied to an optimization parameter, meaning that the number of states that must be optimized goes up linearly with the number of measurements, making the problem intractable when high-frequency sensors are used. Usually to include such sensors, the measurements are grouped and approximated in some way such as IMU preintegration~\cite{Forster2017, Brossard2022}.

Within the last decade or so, continuous-time batch estimation has been introduced~\cite{Furgale2012}. Rather than optimizing a discrete set of robot states, the continuous trajectory of the robot is estimated. The trajectory is represented using a discrete set of parameters and a map from these parameters to the robot state at a given time. The main advantage of continuous-time batch estimation is that the trajectory estimate can be sampled at any time, making it straightforward to incorporate measurements from asynchronous, high-frequency sensors without having to increase the number of optimization parameters. Within the continuous-time estimation literature, there are two major methods that have been proposed to represent continuous robot trajectories: splines (parametric)~\cite{Furgale2012} and Gaussian processes (GPs) (nonparametric)~\cite{Barfoot2014TE}.

\subsubsection{Splines}

Splines~\cite{Cox1972, deboor1978} are piecewise polynomial functions that are characterized by a set of control points on the configuration manifold and a set of knot points in time. The differentiability of the spline is determined by the spline order, which also determines how many control points are used when evaluating the spline at a given time. When performing estimation with splines, the control point locations are optimized such that the resulting trajectory optimally fits the acquired measurements and motion priors~\cite{Furgale2012}. While the original spline formulation is only suited for use on vector spaces, it is possible to generalize them for use with Lie groups~\cite{Kim1995, Sommer2016, Sommer2020}, which is important for state estimation because attitude is most often represented using rotation matrices. Splines have been used for sensor fusion using inertial measurement units (IMUs)~\cite{Furgale2012, Patron2015, Mo2022}, rolling shutter cameras~\cite{Oth2013, Patron2015}, event cameras~\cite{Mueggler2018}, radar~\cite{Ng2021}, and lidar~\cite{Lv2020, Lv2021}. 

\subsubsection{GPs}

Continuous trajectories can also be estimated using GP regression~\cite{Barfoot2014TE, Anderson2015}. In this method, the robot's motion is modeled using a GP and the states at a discrete set of estimation times are optimized. The states at other measurement and query times are evaluated using GP interpolation. The interpolation scheme is derived from the model used to represent the motion of the robot. When performing GP-based estimation on Lie groups, the trajectory is modeled using a locally linear system that evolves in the Lie algebra~\cite{Anderson2015STEAM}. Some models that have been used in the literature include white-noise-on-acceleration (WNOA)~\cite{Anderson2015STEAM}, white-noise-on-jerk (WNOJ)~\cite{Tang2019}, and data-driven motion priors~\cite{Wong2020}. GP regression has been used in a variety of continuous estimation problems, including radar-based localization~\cite{Burnett2022}, lidar odometry using Doppler measurements~\cite{Wu2023}, and even continuum robot pose estimation~\cite{Lilge2022}. The idea has also been extended to continuous-time motion planning problems~\cite{Mukadam2018, Mukadam2019}.

Both splines and GPs are valid ways to represent continuous robot trajectories, and both have been used in a variety of scenarios. Their underlying representations differ in a few significant ways: the estimation parameters lie in different spaces (the configuration manifold for splines versus the state space for GPs), splines do not require the use of a motion model (although one can be used if desired), and the degree-of-differentiability is determined differently (from the chosen order for splines and from the motion model for GPs). It is unclear what advantages and disadvantages each of the methods have, or whether one should be used rather than the other in a given scenario. 

To our knowledge, no formal, direct comparison between these methods exists in the literature. In this paper, we develop such a comparison, with the intent of determining the relative advantages and disadvantages of both. We introduce continuous-time trajectory estimation generally, both for linear systems on vector spaces as well as for nonlinear systems on Lie groups. In the interest of maintaining a fair and accurate comparison, we show how to generalize the motion priors introduced for GP-based estimation in~\cite{Barfoot2014TE} and~\cite{Anderson2015STEAM} so that the same motion model can be used for both GPs and splines. We then compare GP and spline-based estimation directly in two different scenarios: for a simple linear WNOJ system with simulated position measurements, and for camera and IMU sensor fusion on SE(3). For the camera and IMU fusion scenario, we compare the two methods using both simulated data and hardware data in order to validate the comparison against ground truth and also with real sensor measurements.

This paper presents the following contributions:
\begin{itemize}
    \item a direct comparison of GP and spline-based continuous-time trajectory estimation in the case of a linear system with a linear measurement model,
    \item a direct comparison of GP and spline-based estimation on SE(3) using IMU measurements and fiducial pose measurements from a camera, both with simulated and hardware data, and
    \item a generalization of the linear and Lie group GP motion priors in~\cite{Barfoot2014TE, Anderson2015STEAM} for use in spline-based estimation.
\end{itemize}

The outline of this paper is as follows. Section~\ref{sec:cont_traj_est} introduces continuous-time trajectory estimation generally and develops motion models for both linear systems and systems on Lie groups, and Sections~\ref{sec:gp} and~\ref{sec:spline} apply this theory to GP and spline-based estimation, respectively. Section~\ref{sec:example_problems} introduces the comparison scenarios we will be using, and Sections~\ref{sec:lin_sim},~\ref{sec:imu_sim}, and~\ref{sec:imu_hw} present the comparison results for these scenarios. Finally, we finish with some discussion and concluding remarks in Sections~\ref{sec:discussion} and~\ref{sec:conclusion}, respectively.
    \section{Continuous-time Trajectory Estimation}
\label{sec:cont_traj_est}

Let $\mathcal{X}$ represent the state space of a dynamic system and let $\mathcal{W} = [t_\text{min}, t_\text{max}]$ represent a window of time defined by $t_\text{min}$, $t_\text{max}$. Our objective is to find the best estimate of a continuous trajectory $\mathbf{x}(\mathcal{W}) = \{ \mathbf{x}(t) \: | \: t \in \mathcal{W}\}$ taken by the system given a set of measurements collected from a number of different sensors (indexed by the set $\mathcal{S}$) throughout the time window. The sensors do not need to run at similar frequencies or be in phase with one another. Each measurement $\mathbf{z}_i^s$, $s \in S$, $i = 1, \cdots, N_s$ for sensor $s$ is modeled by
\begin{equation}
    \mathbf{z}_i^s = \mathbf{h}^s(\mathbf{x}(t_i^s)) + \boldsymbol{\eta},
    \label{eq:meas_model}
\end{equation}
where $h^s(\cdot)$ is the measurement model, $t_i^s \in \mathcal{W}$ is the measurement time, and $\boldsymbol{\eta} \sim \mathcal{N}(\mathbf{0}, \boldsymbol{\Sigma}_s)$ is zero-mean Gaussian measurement noise\footnote{It is often the case that measurements $\mathbf{z}_i^s$ belong to a Lie group. Rather than using the additive noise model~\eqref{eq:meas_model}, we perturb the measurement with noise in either the left or the right tangent space, e.g., $\mathbf{z}_i^s = \text{Exp}(\boldsymbol{\eta}) h^s(\cdot)$~\cite{Barfoot2014}.} with covariance $\boldsymbol{\Sigma}_s$. 

The goal of continuous-time estimation is to find the continuous state trajectory $\mathbf{x}(\mathcal{W})$ that solves the maximum-a-posteriori problem
\begin{equation}
    \max_{\mathbf{x}(\mathcal{W})} p(\mathbf{x}(\mathcal{W})) | \mathbf{z}) \propto p(\mathbf{z} | \mathbf{x}(\mathcal{W})) p(\mathbf{x}(\mathcal{W})),
    \label{eq:inf_map}
\end{equation}
where $\mathbf{z} = \{\mathbf{z}_i^s \}_{s \in S, i = 1, \cdots, N_s}$, and the prior term $p(\mathbf{x}(\mathcal{W}))$ represents a continuous Markov process. This is an infinite-dimensional estimation problem. Rather than solve it in this form, we will represent $\mathbf{x}(\mathcal{W})$ using a discrete set of estimation parameters $\{ \bar{\mathbf{x}}_n \in \mathcal{E}\}_{n = 1, \cdots, N}$ (where $\mathcal{E}$ is the estimation parameter space) and a function $\iota: \mathcal{W} \times \mathcal{E}^N \rightarrow \mathcal{X}$ such that
\begin{equation}
    \mathbf{x}(t) = \iota(t, \{\bar{\mathbf{x}}_n\}).
    \label{eq:interp}
\end{equation}
We will refer to $\iota(\cdot)$ as the \textit{state interpolation function}. In addition, the prior term $p(\mathbf{x}(\mathcal{W}))$ must be approximated. We will do this using a discrete set of motion priors at the motion sample times $\{t_j^\prime \in \mathcal{W} \}_{j = 1, \cdots, N_m}$. These motion priors will be of the form
\begin{equation}
    \mathbf{m}\left(\mathbf{x}(t_{j}^\prime), \mathbf{x}(t_{j+1}^\prime) \right)
\end{equation}
with covariance ${\mathbf{Q}_{t_{j+1}^\prime}}$. We explain how these motion priors can be derived for certain model types in Section~\ref{sec:motion_priors}.

\begin{figure}
    \centering
    \includegraphics[width=0.45\textwidth]{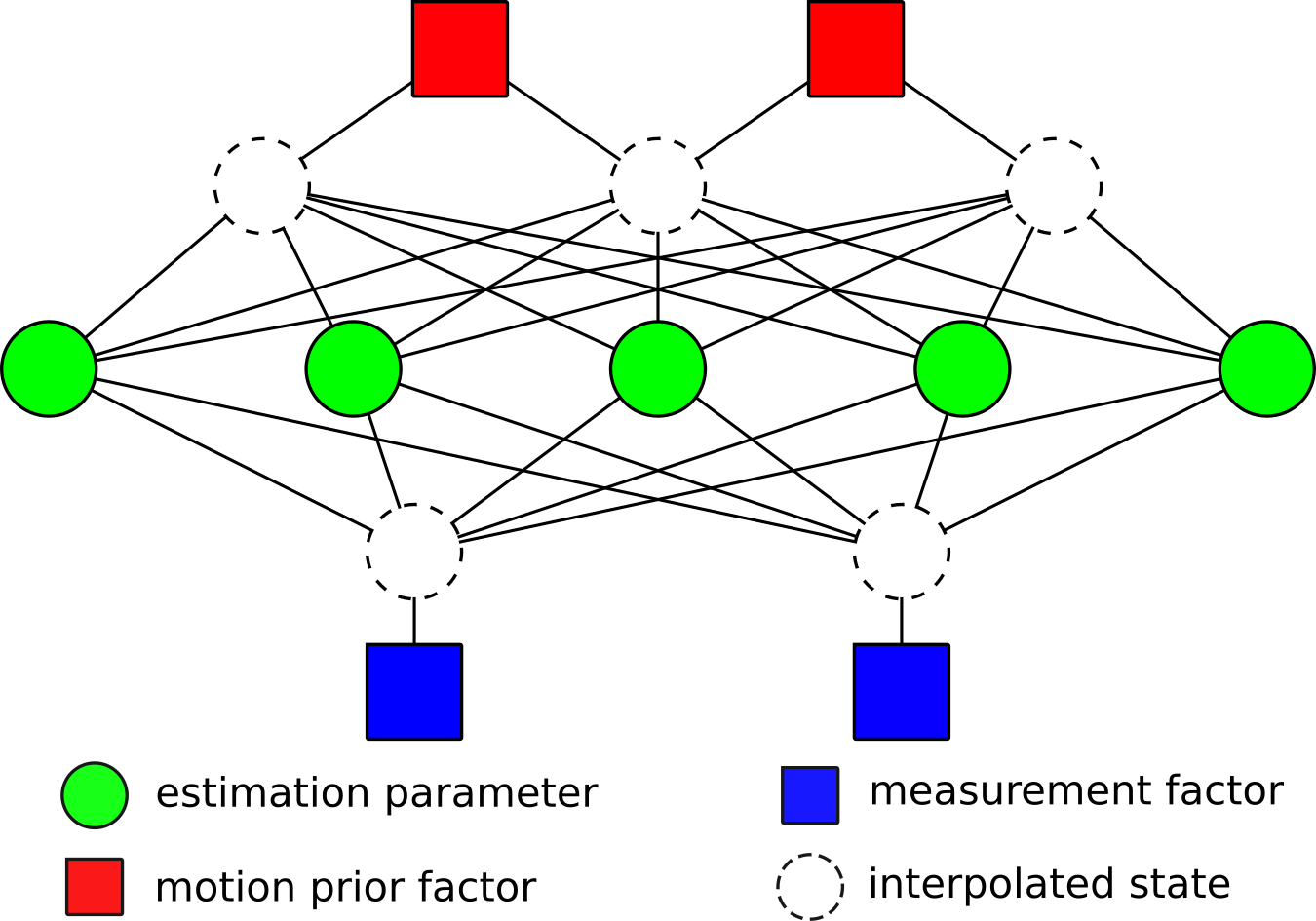}
    \caption{Example factor graph for the continuous-time estimation problem~\eqref{eq:problem}. The interpolated states are not directly optimized, rather they are computed from the estimation parameters using the state interpolation function~\eqref{eq:interp}.}
    \label{fig:ct_factor_graph}
\end{figure}

If the measurements are independent of one another and are only dependent on the state at the time the measurement was acquired, then~\eqref{eq:inf_map} becomes\footnote{In the case that a measurement $\mathbf{z}_i^s$ belongs to a Lie group, we replace the inside of the norm in~\eqref{eq:problem} with $\text{Log}\left( \mathbf{z}_i^s h^s(\cdot)^{-1} \right)$.}
\begin{equation}
\begin{split}
    \{ \bar{\mathbf{x}}^\ast \}_n &= \argmin_{\{ \bar{\mathbf{x}} \}_n} \sum_{s \in S} \sum_{i=1}^{N_s} \lVert \mathbf{z}_i^s - \mathbf{h}^s\left(\iota(t_i^s, \{ \bar{\mathbf{x}}_n \})\right) \rVert^2_{\boldsymbol{\Sigma}_s^{-1}} \\
    &\quad + \sum_{j=1}^{N_m-1} \lVert \mathbf{m}\left(\iota(t_{j}^\prime, \{ \bar{\mathbf{x}}_n \}), \iota(t_{j+1}^\prime, \{ \bar{\mathbf{x}}_n \}) \right) \rVert^2_{\mathbf{Q}_{t_{j+1}^\prime}^{-1}},
    \label{eq:problem}
\end{split}
\end{equation}
where $\lVert \cdot \rVert_\mathbf{W}$ indicates the 2-norm weighted by the matrix $\mathbf{W}$. An example factor graph for this problem is shown in Figure~\ref{fig:ct_factor_graph}.

\begin{note}
    The graph in Figure~\ref{fig:ct_factor_graph} is dense with respect to the estimation parameters because the general interpolation function~\eqref{eq:interp} is dependent on every parameter. This need not be the case, however. The state interpolation methods used in Sections~\ref{sec:gp} and~\ref{sec:spline} are chosen such that the interpolation is only dependent on a few of the parameters in order to make the problem more sparse.
\end{note}

\subsection{Motion Priors}
\label{sec:motion_priors}

To approximate the prior term $p(\mathbf{x}(\mathcal{W}))$ in~\eqref{eq:inf_map}, we will assume that the state $\mathbf{x}$ evolves according to a certain stochastic model. We will first consider linear models and then show how this formulation can be extended to Lie groups. These motion priors were originally presented in~\cite{Barfoot2014TE, Anderson2015STEAM, Tang2019} for GP-based estimation. Here we generalize them so they can be used in spline-based estimation as well.

\subsubsection{Linear systems}

We restrict our attention to time-invariant systems for simplicity. Assume that the state evolves according to
\begin{equation}
    \dot{\mathbf{x}}(t) = \mathbf{A} \mathbf{x}(t) + \mathbf{B} \mathbf{u(t)} + \mathbf{L}\mathbf{w}(t),
    \label{eq:lin_sys}
\end{equation}
where $\mathbf{A}$, $\mathbf{B}$, and $\mathbf{L}$ are the system matrices, $\mathbf{u}(t)$ is a known system input, and $\mathbf{w}(t) \sim \mathcal{GP}(\mathbf{0}, \mathbf{Q} \delta(t - \tau))$ is a stationary GP with power spectral density $\mathbf{Q}$. The state transition matrix of the system is
\begin{equation}
    \boldsymbol{\Phi}(t, \tau) = \text{exp}(\mathbf{A}(t - \tau)),
\end{equation}
where $\text{exp}(\cdot)$ is the matrix exponential. If we have a guess for an initial state $\check{\mathbf{x}}(t_0)$, then the \textit{prior mean function} is
\begin{equation}
    \check{\mathbf{x}}(t) = \boldsymbol{\Phi}(t, t_0) \check{\mathbf{x}}_0 + \int_{t_0}^t \boldsymbol{\Phi}(t, s) \mathbf{B} \mathbf{u}(s) ds.
    \label{eq:gp_prior_mean}
\end{equation}
Because $\mathbf{x}(t)$ is a GP, the prior term $p(\mathbf{x}(t))$ sampled at any time $\tau$ is
\begin{equation}
\begin{split}
    &p(\mathbf{x}(\tau)) \propto \\
    &\quad \text{exp}\left( \frac{1}{2} (\mathbf{x}(\tau) - \check{\mathbf{x}}(\tau))^\top \check{\mathbf{P}}(\tau, \tau)^{-1} (\mathbf{x}(\tau) - \check{\mathbf{x}}(\tau))\right),
    \label{eq:lin_prior_samp}
\end{split}
\end{equation}
where $\check{\mathbf{P}}(\tau, \tau)$ is the prior covariance at $\tau$. It can be shown~\cite{Barfoot2014TE} that, for $\tau = t_{j+1}^\prime$,~\eqref{eq:lin_prior_samp} is equivalent to
\begin{equation}
    \text{exp}\left( \frac{1}{2} \mathbf{m}\left(\mathbf{x}(t_{j}^\prime), \mathbf{x}(t_{j+1}^\prime) \right)^\top \mathbf{Q}_{t_{j+1}^\prime}^{-1} \mathbf{m}\left(\mathbf{x}(t_{j}^\prime), \mathbf{x}(t_{j+1}^\prime) \right)\right),
\end{equation}
where\footnote{If there is no known input $\mathbf{u}(t)$ to the system, then the prior terms in~\eqref{eq:lin_motion_priors} cancel out.}
\begin{equation}
\begin{split}
    &\mathbf{m}\left(\mathbf{x}(t_{j}^\prime), \mathbf{x}(t_{j+1}^\prime) \right) = \\
    &\qquad \mathbf{x}(t_{j+1}^\prime) - \check{\mathbf{x}}(t_{j+1}^\prime) - \boldsymbol{\Phi}(t_{j+1}^\prime, t_j^\prime) (\mathbf{x}(t_{j}^\prime) - \check{\mathbf{x}}(t_{j}^\prime))
\end{split}
\label{eq:lin_motion_priors}
\end{equation}
and, for $\nu \in [t_{j}^\prime, t_{j+1}^\prime]$,
\begin{equation}
    \mathbf{Q}_\nu = \int_0^{\nu - t_{j}^\prime} \boldsymbol{\Phi}(\nu - t_{j}^\prime, s) \mathbf{L} \mathbf{Q} \mathbf{L}^\top \boldsymbol{\Phi}(\nu - t_{j}^\prime, s)^\top ds.
    \label{eq:q_gp}
\end{equation}

\subsubsection{Lie groups}
\label{sec:lie_prior}

Let $G$ be a matrix Lie group of dimension $d$ with Lie algebra $\mathfrak{g}$. We will break the trajectory $g(\mathcal{W}) \in G \times \mathcal{W}$ into segments between motion sample times and assume that the system obeys a linear stochastic model in the Lie algebra in each of these segments~\cite{Anderson2015STEAM}. Specifically, we restrict our attention to a WNOJ model for simplicity\footnote{This can easily be simplified to a WNOA model by ignoring the second derivative terms and applying noise to the second derivative rather than the third.}~\cite{Tang2019}. Let the state be $\mathbf{x}(t) = \{g(t), \dot{g}(t), \ddot{g}(t)\} \in G \times \mathbb{R}^d \times \mathbb{R}^d$, where $\dot{g}(t)$ is represented as an infinitesimal perturbation in the left tangent space of $g(t)$ and $\ddot{g}(t) = \frac{d}{dt} \dot{g}(t)$. Without loss of generality assume that $t \in [t_{j}^\prime, t_{j+1}^\prime]$, and let
\begin{equation}
    \boldsymbol{\xi}_j(t) = \text{Log}\left( g(t) g(t_j^\prime)^{-1} \right),
    \label{eq:mp_xi}
\end{equation}
where $\text{Log} = \vee \circ \text{log}$, $\vee: \mathfrak{g} \rightarrow \mathbb{R}^d$ is the vee map from the Lie algebra, and $\text{log}: G \rightarrow \mathfrak{g}$ is the logarithmic map of $G$. If $\boldsymbol{\xi}_j(t)$ evolves according to a WNOJ model, then
\begin{equation}
    \dot{\boldsymbol{\gamma}}_j(t) = \begin{bmatrix} \mathbf{0} & \mathbf{I} & \mathbf{0} \\ \mathbf{0} & \mathbf{0} & \mathbf{I} \\ \mathbf{0} & \mathbf{0} & \mathbf{0} \end{bmatrix} \boldsymbol{\gamma}_j(t) + \begin{bmatrix} \mathbf{0} \\ \mathbf{0} \\ \mathbf{I} \end{bmatrix} \mathbf{w}(t),
    \label{eq:gamma_dot}
\end{equation}
where $\boldsymbol{\gamma}_j(t) = \begin{bmatrix} \boldsymbol{\xi}_j(t)^\top & \dot{\boldsymbol{\xi}}_j(t)^\top & \ddot{\boldsymbol{\xi}}_j(t)^\top \end{bmatrix}^\top$ and $\mathbf{w}(t) \sim \mathcal{GP}(\mathbf{0}, \mathbf{Q} \delta(t - \tau))$. We can express this system in terms of global variables using~\eqref{eq:mp_xi} and its time derivatives
\begin{equation}
\begin{gathered}
    \dot{\boldsymbol{\xi}}_j(t) = J_l^{-1}(\boldsymbol{\xi}_j(t)) \dot{g}(t) \\
\begin{aligned}
    \ddot{\boldsymbol{\xi}}_j(t) &= J_l^{-1}(\boldsymbol{\xi}_j(t)) \ddot{g}(t) + \frac{d}{dt} \left( J_l^{-1}(\boldsymbol{\xi}_j(t)) \right) \dot{g}(t) \\
    &\approx J_l^{-1}(\boldsymbol{\xi}_j(t)) \ddot{g}(t) - \frac{1}{2} \dot{\boldsymbol{\xi}}_j(t)^\curlywedge \dot{g}(t),
\end{aligned}
\end{gathered}
\label{eq:gp_local_var}
\end{equation}
(see Appendix~\ref{sec:lie_gp_derivatives} or~\cite{Tang2019} for the derivation), where $J_l(\cdot)$ is the left Jacobian of $G$, $J_l^{-1}(\cdot)$ is its inverse, and $\curlywedge: \mathfrak{g} \to \text{ad}(\mathfrak{g})$ maps elements of $\mathfrak{g}$ to their adjoint representation.

In the Lie algebra, the motion priors take the same form as~\eqref{eq:lin_motion_priors}, with
\begin{gather}
    \boldsymbol{\Phi}(t, t_j^\prime) = \begin{bmatrix} \mathbf{I} & (t - t_j^\prime)\mathbf{I} & \frac{1}{2}{(t - t_j^\prime)}^2\mathbf{I} \\ \mathbf{0} & \mathbf{I} & (t - t_j^\prime)\mathbf{I} \\ \mathbf{0} & \mathbf{0} & \mathbf{I} \end{bmatrix}, \label{eq:wnoj_phi}\\
    \mathbf{Q}_\nu = \begin{bmatrix} \frac{1}{20} {(\nu - t_j^\prime)}^5 \mathbf{Q} & \frac{1}{8} {(\nu - t_j^\prime)}^4 \mathbf{Q} & \frac{1}{6} {(\nu - t_j^\prime)}^3 \mathbf{Q} \\ \frac{1}{8} {(\nu - t_j^\prime)}^4 \mathbf{Q} & \frac{1}{3} {(\nu - t_j^\prime)}^3 \mathbf{Q} & \frac{1}{2} {(\nu - t_j^\prime)}^2 \mathbf{Q} \\ \frac{1}{6} {(\nu - t_j^\prime)}^3 \mathbf{Q} & \frac{1}{2} {(\nu - t_j^\prime)}^2 \mathbf{Q} & (\nu - t_j^\prime) \mathbf{Q} \end{bmatrix}, \label{eq:wnoj_q}
\end{gather}
for $\nu \in [t_j^\prime, t_{j+1}^\prime]$. Converting to global variables, we get
\begin{equation}
\begin{split}
    &\mathbf{m}\left(\mathbf{x}(t_{j}^\prime), \mathbf{x}(t_{j+1}^\prime) \right) = \\
    &\quad\begin{bmatrix} \boldsymbol{\xi}_j(t_{j+1}^\prime) - \Delta t \dot{g}(t_j^\prime) - \frac{1}{2} \Delta t^2 \ddot{g}(t_j^\prime) \\ \dot{\boldsymbol{\xi}}_j(t_{j+1}^\prime) - \dot{g}(t_j^\prime) - \Delta t \ddot{g}(t_j^\prime) \\ J_l^{-1}(\boldsymbol{\xi}_j(t_{j+1}^\prime))\ddot{g}(t_{j+1}^\prime) - \frac{1}{2} \dot{\boldsymbol{\xi}}_j(t_{j+1}^\prime)^\curlywedge \dot{g}(t_{j+1}^\prime) - \ddot{g}(t_j^\prime) \end{bmatrix}
    \label{eq:mp_lie}
\end{split}
\end{equation}
where $\Delta t = t_{j+1}^\prime - t_j^\prime$, $\boldsymbol{\xi}_j(t_{j+1}^\prime) = \text{Log}\left( g(t_{j+1}^\prime) g(t_j^\prime)^{-1} \right)$, and $\dot{\boldsymbol{\xi}}_j(t_{j+1}^\prime) = J_l^{-1}(\boldsymbol{\xi}_j(t_{j+1}^\prime)) \dot{g}(t_{j+1}^\prime)$. The Jacobians for these motion priors are derived in Appendix~\ref{sec:mp_jacs}.
    \section{Gaussian Process Regression}
\label{sec:gp}

We will first look at Gaussian process (GP) regression as a method for performing continuous-time estimation. We introduce the state interpolation function for GPs on linear systems in Section~\ref{sec:gp_linear}, and then for Lie groups in Section~\ref{sec:gp_lie}. Finally, we explain how to use GPs in estimation problems in Section~\ref{sec:gp_est}. For a more in-depth treatment of GP regression for continuous-time estimation, see~\cite{Barfoot2017}.

The GP interpolation approach is similar to spline interpolation in that it can be derived by representing the state trajectory using an arbitrary set of basis functions (i.e., the weight-space formulation in~\cite{Rasmussen2005}); however, the interpolation equation can be manipulated into a form that is non-parametric. We will not show this derivation here for the sake of brevity, but invite the reader to see~\cite{Rasmussen2005, Tong2013} for more details.

\subsection{Linear Systems}
\label{sec:gp_linear}

Choose a set of \textit{estimation times} $\{ t_n \in \mathcal{W} \}_{n=0,\cdots,N-1}$. The state estimates at these times $\{ \bar{\mathbf{x}}_n = \mathbf{x}(t_n) \}_{n=0, \cdots, N-1}$ will be the estimation parameters that govern the trajectory estimate. Assume that the state $\mathbf{x}(t)$ evolves according to~\eqref{eq:lin_sys}. We would like to sample the trajectory at some arbitrary time $\mathbf{\tau} \in [t_n, t_{n+1}]$ using the estimation parameters. It can be shown~\cite{Barfoot2014TE} that the posterior mean function of the GP evaluated at $\tau$ given the estimation parameters is
\begin{equation}
    \mathbf{x}(\tau) = \check{\mathbf{x}}(\tau) + \boldsymbol{\Lambda}(\tau) (\bar{\mathbf{x}}_n - \check{\mathbf{x}}_n) + \boldsymbol{\Omega}(\tau) (\bar{\mathbf{x}}_{n+1} - \check{\mathbf{x}}_{n+1}),
    \label{eq:gp_interp_prior}
\end{equation}
where
\begin{equation}
\begin{gathered}
    \boldsymbol{\Omega}(\tau) = \mathbf{Q}_\tau \boldsymbol{\Phi}(t_{n+1}, \tau)^\top \mathbf{Q}_{t_{n+1}}^{-1}, \\
    \boldsymbol{\Lambda}(\tau) = \boldsymbol{\Phi}(\tau, t_n) - \boldsymbol{\Omega}(\tau) \boldsymbol{\Phi}(t_{n+1}, t_n),
\end{gathered}
\label{eq:om_lam}
\end{equation}
$\mathbf{Q}_\tau$ and $\mathbf{Q}_{t_{n+1}}$ are defined as in~\eqref{eq:q_gp}, and the prior mean $\check{\mathbf{x}}_n = \check{\mathbf{x}}(t_n)$ was defined in~\eqref{eq:gp_prior_mean}. This can be simplified if no information is known about the input $\mathbf{u}(t)$. Then the integral term in the prior mean function~\eqref{eq:gp_prior_mean} vanishes, the prior terms in the interpolation function~\eqref{eq:gp_interp_prior} cancel out, and~\eqref{eq:gp_interp_prior} becomes
\begin{equation}
    \mathbf{x}(\tau) = \boldsymbol{\Lambda}(\tau) \bar{\mathbf{x}}_n + \boldsymbol{\Omega}(\tau) \bar{\mathbf{x}}_{n+1}.
    \label{eq:gp_interp}
\end{equation}

\begin{note}
    Interestingly, it is also possible to arrive at~\eqref{eq:gp_interp_prior} using an optimal control perspective. See Appendix~\ref{sec:gp_optimal_control}.
\end{note}

% An example mean function of a one-dimensional GP with a constant velocity model is shown in Figure~\ref{fig:gp_ex}.

% \begin{figure}
%     \centering
%     \includegraphics{}
%     \caption{Caption}
%     \label{fig:gp_ex}
% \end{figure}

\subsection{Lie Groups}
\label{sec:gp_lie}

For an arbitrary Lie group $G$, we will again use the WNOJ motion model introduced in Section~\ref{sec:lie_prior}. Given the set of estimation times $\{ t_n \in \mathcal{W} \}_{n=0,\cdots,N-1}$, the estimation parameters will be the states at these times $\left\{ \bar{\mathbf{x}}_n = \{ \bar{g}_n, \dot{\bar{g}}_n, \ddot{\bar{g}}_n \} \in G \times \mathbb{R}^d \times \mathbb{R}^d \right\}_{n = 0, \cdots, N-1}$, where $\bar{g}_n = g(t_n)$, $\dot{\bar{g}}_n = \dot{g}(t_n)$, and $\ddot{\bar{g}}_n = \ddot{g}(t_n)$. Let $t \in [t_n, t_{n+1}]$, and define
\begin{equation}
    \boldsymbol{\xi}_n(t) = \text{Log}\left( g(t)\bar{g}_n^{-1} \right),
    \label{eq:gp_xi}
\end{equation}
as was done in~\eqref{eq:mp_xi}. Then $\boldsymbol{\gamma}_n(t) = \begin{bmatrix} \boldsymbol{\xi}_n(t)^\top & \dot{\boldsymbol{\xi}}_n(t)^\top & \ddot{\boldsymbol{\xi}}_n(t)^\top \end{bmatrix}^\top$ evolves according to~\eqref{eq:gamma_dot}, and its state interpolation function can be defined using~\eqref{eq:gp_interp},
\begin{equation}
    \boldsymbol{\gamma}_n(\tau) = \boldsymbol{\Lambda}(\tau) \boldsymbol{\gamma}_n(t_n) + \boldsymbol{\Omega}(\tau) \boldsymbol{\gamma}_{n}(t_{n+1}),
    \label{eq:gp_interp_lie}
\end{equation}
where $\boldsymbol{\Lambda}(\tau)$ and $\boldsymbol{\Omega}(\tau)$ were defined in~\eqref{eq:om_lam}, and $\boldsymbol{\Phi}(t, \tau)$ and $\mathbf{Q}_\nu$ were defined in~\eqref{eq:wnoj_phi} and~\eqref{eq:wnoj_q} respectively. We can convert these local interpolated variables back into the global representation by inverting~\eqref{eq:gp_xi} and its derivatives~\eqref{eq:gp_local_var} (with the index $j$ replaced with $n$), i.e.,~\cite{Tang2019}
\begin{equation}
\begin{gathered}
    g(t) = \text{Exp}(\boldsymbol{\xi}_n(t))\bar{g}_n \\
    \dot{g}(t) = J_l(\boldsymbol{\xi}_n(t)) \dot{\boldsymbol{\xi}}_n(t) \\
    \ddot{g}(t) \approx J_l(\boldsymbol{\xi}_n(t)) \left(\ddot{\boldsymbol{\xi}}_n(t) + \frac{1}{2}\dot{\boldsymbol{\xi}}_n(t)^\curlywedge \dot{g}(t)\right),
\end{gathered}
\label{eq:gp_global_var}
\end{equation}
where $\text{Exp}: \text{exp} \circ \wedge$, $\wedge: \mathbb{R}^d \rightarrow \mathfrak{g}$ is the wedge map to the Lie algebra, and $\text{exp}: \mathfrak{g} \rightarrow G$ is the exponential map.

\subsection{Estimation}
\label{sec:gp_est}

\begin{figure}
    \centering
    \includegraphics[width=0.45\textwidth]{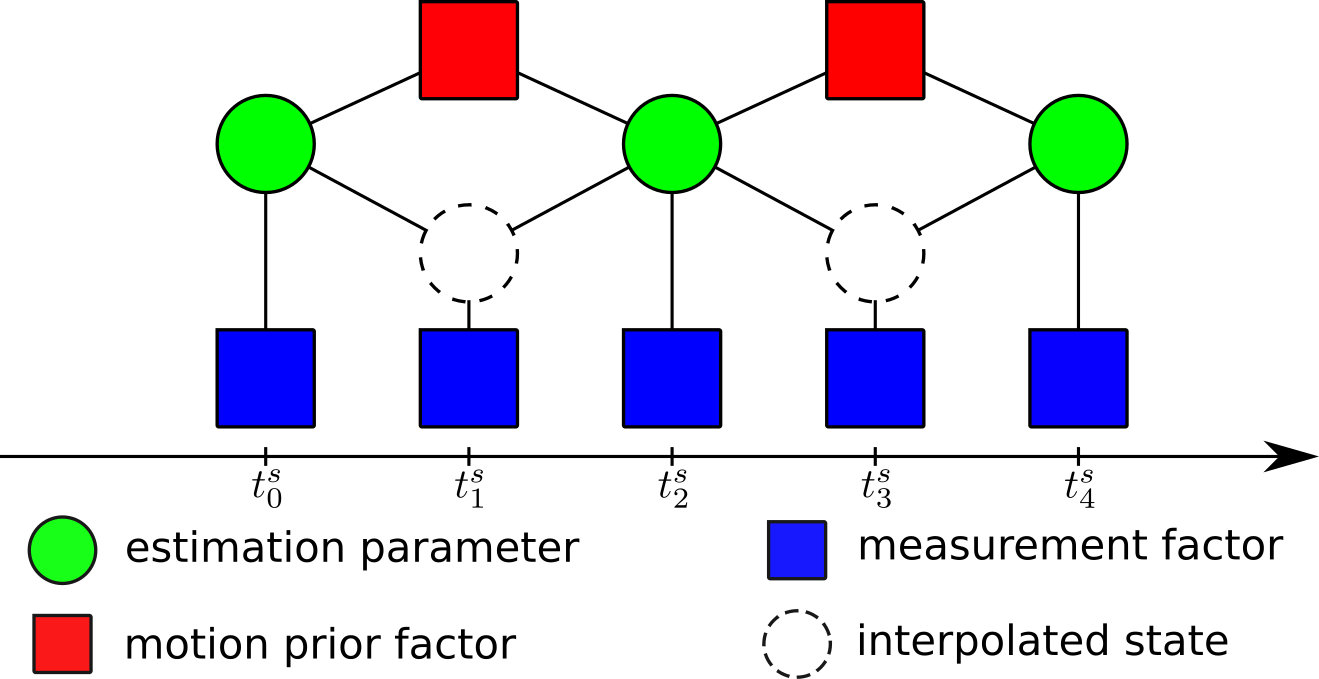}
    \caption{An example factor graph for a GP-based estimator.}
    \label{fig:gp_factor_graph}
\end{figure}

When performing estimation using GP regression, the states at the estimation times $\bar{\mathbf{x}}_n \in \mathcal{X}$ are the estimation parameters. It is usually preferable (although not necessary) to choose estimation times that correlate with a subset of the measurement times. Then, when evaluating residuals for these measurements, no state interpolation is necessary. However, it is often the case (e.g., when using high-frequency sensors) that there are too many measurements to associate a new estimation parameter with every measurement time. For measurement times that are not associated with an estimation parameter, the state is computed using the GP interpolation equation~\eqref{eq:gp_interp} (or the composition of equations~\eqref{eq:gp_xi},~\eqref{eq:gp_local_var},~\eqref{eq:gp_interp_lie}, and~\eqref{eq:gp_global_var} for Lie groups). The motion sample times $\{t_j^\prime \}$ for the motion priors should coincide with the estimation times to avoid performing GP interpolation when evaluating them. Note that while the GP interpolation equation~\eqref{eq:gp_interp} is derived using a motion model, it is still possible to exclude the motion prior terms in~\eqref{eq:problem} during estimation. We will refer to the inclusion (\textit{resp.} exclusion) of these terms as ``GP-based estimation \textit{with} (\textit{resp.} \textit{without}) motion priors''. An example factor graph for GP-based estimation is shown in Figure~\ref{fig:gp_factor_graph}.

In order to perform this estimation, we will need to compute the Jacobians of the interpolation equations as well as the motion priors with respect to the estimation parameters\footnote{For linear systems the interpolation function is linear, so these Jacobians are constant.}. Unfortunately, we are only able to do this approximately for Lie groups. These Jacobians are derived in Appendix~\ref{sec:lie_gp_jacs}.

    \section{Spline-Based Estimation}
\label{sec:spline}

Splines are piece-wise polynomial functions that are $\mathcal{C}^{k-2}$-continuous (where $k$ is the \textit{spline order}) at every point. They are an appealing way to model continuous trajectories because their evaluation at any point is only dependent on $k$ different parameters known as \textit{control points}. The total number of control points used to represent the spline trajectory is a hyper-parameter that the user is able to specify. We first explain how splines are defined on vector spaces in Section~\ref{sec:spl_euclid}, and then show how this definition can be extended to Lie groups in Section~\ref{sec:spl_lie}. Finally, we discuss how splines are used in continuous-time trajectory estimation problems in Section~\ref{sec:spl_est}.

\subsection{Vector Spaces}
\label{sec:spl_euclid}

\begin{figure}
    \centering
    \includegraphics[width=0.45\textwidth]{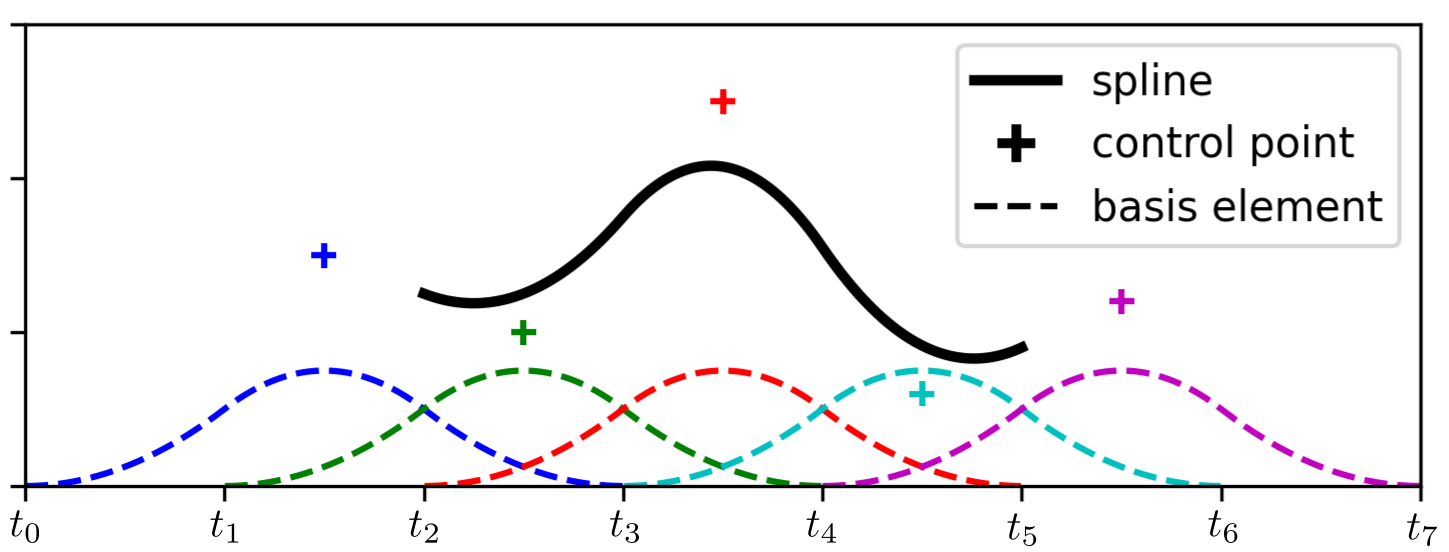}
    \caption{One-dimensional spline example, with $k = 3$.}
    \label{fig:spline_ex}
\end{figure}

A spline of dimension $d$ and order $k$ is completely characterized by a set of control points $\{ \bar{\mathbf{p}}_n \in \mathbb{R}^d \}_{n = 1, \cdots, N}$ and monotonically increasing knot points $\{t_m^\dagger \in \mathbb{R} \}_{m = 1, \cdots, N + k}$. In our treatment, we will assume that the spacing between the knot points is uniform, i.e., $t_m^\dagger - t_{m-1}^\dagger = \delta t \ \forall \  m$, for simplicity\footnote{It is possible to use non-uniform splines in estimation problems as well~\cite{Anderson2014, Dube2016, Hug2020}. However, this adds an additional level of complexity and is beyond the scope of this paper.}. The spline can be sampled at any point $t \in [t_k^\dagger, t_{N+1}^\dagger]$ using
\begin{equation}
    \mathbf{p}(t) = \sum_{n=1}^N B_{n,k}(t) \bar{\mathbf{p}}_n,
    \label{eq:spline_eval_sum}
\end{equation}
where $B_{n,k}(t)$ are the spline basis functions of order $k$~\cite{Cox1972}. An example of a one-dimensional spline with $k = 3$ is shown in Figure~\ref{fig:spline_ex}.

While~\eqref{eq:spline_eval_sum} is sufficient for spline evaluation, it requires evaluating all of the spline basis functions, only $k$ of which will be nonzero for any value of $t$, and it is unwieldy to use in estimation problems. We can re-express~\eqref{eq:spline_eval_sum} in a  cumulative form
\begin{equation}
    \mathbf{p}(t) = \tilde{B}_{1,k}(t) \bar{\mathbf{p}}_1 + \sum_{n = 2}^M \tilde{B}_{n,k}(t) (\bar{\mathbf{p}}_n - \bar{\mathbf{p}}_{n-1}),
\end{equation}
where $\tilde{B}_{n,k}(t) = \sum_{l=n}^M B_{l,k}(t)$. If we remove terms that evaluate to zero, we get
\begin{equation}
    \mathbf{p}(t) = \bar{\mathbf{p}}_{n-k+1} + \sum_{j=1}^{k-1} b_j(u(t)) (\bar{\mathbf{p}}_{n+j-k+1} - \bar{\mathbf{p}}_{n+j-k}),
    \label{eq:spline_eval_cumu}
\end{equation}
where the index $n$ is chosen such that $t \in (t_n^\dagger, t_{n+1}^\dagger]$, $u(t) = \frac{t - t_n^\dagger}{t_{n+1}^\dagger - t_n^\dagger} \in (0, 1]$, and $b_j(u)$ is the $j$-th element of $\mathbf{b}(u) = \mathbf{C}_k \boldsymbol{\mu}(u)$, where $\mathbf{C}_k \in \mathbb{R}^{k\times k}$ encodes the continuity constraints of the spline~\cite{Qin1998} and 
\begin{equation}
    \boldsymbol{\mu}(u) = \begin{bmatrix} 1 & u(t) & u(t)^2 & \cdots & u(t)^{k-1} \end{bmatrix}^\top.
\end{equation}
Finally, we can rewrite~\eqref{eq:spline_eval_cumu} as
\begin{equation}
    \mathbf{p}(t) = \Theta(t) \bar{\mathbf{p}},
    \label{eq:spline_eval_mat}
\end{equation}
where
\begin{equation}
    \Theta(t) = \begin{bmatrix} (b_0 - b_1) \mathbf{I} & (b_1 - b_2) \mathbf{I} & \cdots & b_{k-1} \mathbf{I} \end{bmatrix}
\end{equation}
(the dependence of $b_j$ on $u(t)$ has been omitted for brevity) and $\bar{\mathbf{p}} = \begin{bmatrix} \bar{\mathbf{p}}_{n-k+1}^\top & \bar{\mathbf{p}}_{n-k+2}^\top & \cdots & \bar{\mathbf{p}}_{n}^\top \end{bmatrix}^\top$. Computing time-derivatives of Euclidean B-splines is trivial with~\eqref{eq:spline_eval_mat},
\begin{equation}
    \dot{\mathbf{p}}(t) = \dot{\Theta}(t) \bar{\mathbf{p}}, \quad \ddot{\mathbf{p}}(t) = \ddot{\Theta}(t) \bar{\mathbf{p}}, \quad \cdots.
    \label{eq:spl_deriv}
\end{equation}

\subsection{Lie Groups}
\label{sec:spl_lie}

It is possible to extend this representation of vector space splines to matrix Lie groups. We begin with the cumulative representation~\eqref{eq:spline_eval_cumu}. Recall that the vector space $\mathbb{R}^d$ is itself a Lie group, where the group operation is vector addition, inversion is vector negation, and the exponential and logarithmic maps are identity. We can generalize~\eqref{eq:spline_eval_cumu} to any Lie group by substituting these operations. Let $G$ be an arbitrary Lie group with Lie algebra $\mathfrak{g}$ and dimension $d$, and let $\{ \bar{g}_n \in G \}_{n=1, \cdots, N}$ be a set of control points. Then a spline on $G$ can be sampled at the point $t$ using
\begin{equation}
\begin{gathered}
    g(t) = \bar{g}_{n-k+1} \prod_{j=1}^M \text{Exp}\left( b_j(u(t)) \boldsymbol{\Omega}_{n+j-k+1}\right), \\
    \boldsymbol{\Omega}_i = \text{Log}(\bar{g}_{i-1}^{-1} \bar{g}_i).
    \label{eq:spl_lie}
\end{gathered}
\end{equation}
This can be understood as taking products of scaled geodesics between consecutive control points. Note that~\eqref{eq:spline_eval_cumu} and~\eqref{eq:spl_lie} are identical for the specific Lie group $\mathbb{R}^d$.

We will need to compute time derivatives $\dot{g}(t), \ddot{g}(t) \in \mathbb{R}^d$ of~\eqref{eq:spl_lie} (which represent the same quantities as those in Section~\ref{sec:lie_prior}). This process is much more complicated than it was for vector space splines~\eqref{eq:spl_deriv}. We show how to compute the first two time derivatives in Appendix~\ref{sec:spl_derivs}.

\subsection{Estimation}
\label{sec:spl_est}

When performing estimation using splines, the control points are usually defined on the configuration manifold $\mathcal{M}$ of the system rather than on the state space $\mathcal{X}$. Let the state of the system be $\mathbf{x}(t) = \{ g(t), \dot{g}(t), \ddot{g}(t), \cdots \} \in \mathcal{M} \times \mathbb{R}^d \times \mathbb{R}^d \times \cdots = \mathcal{X}$. The state interpolation function~\eqref{eq:interp} for splines is then computed using~\eqref{eq:spline_eval_mat} and~\eqref{eq:spl_deriv} for vector space splines or~\eqref{eq:spl_lie} and its derivatives in Appendix~\ref{sec:spl_derivs} for Lie group splines. 

% As opposed to GPs, splines are not motivated by any dynamic model, thus there is no obvious way to constrain the motion defined by the spline. However, it is possible to impose a dynamic model on $\mathbf{x}(t)$ and form motion priors that will encourage the spline control points to satisfy the model. In this paper, we elect to use the same dynamic models that motivated GP-based estimation in Section~\ref{sec:gp}. To form the motion priors~\eqref{eq:motion_priors}, we choose a set of spline sample times $\{ t_j^\diamond \in \mathcal{W} \}_{j = 1, \cdots, N_m}$, evaluate the state at each, and form the same motion residuals for GPs in~\eqref{eq:gp_mp_lin} and~\eqref{eq:gp_mp_lie}. Unlike the GP formulation, there is not an obvious way to choose the motion sample times in order to fully constrain the control points. In our analysis for Euclidean splines in Section~\ref{sec:lin_sim}, we will show the effect that the motion sample period $\delta t^\diamond = t_j^\diamond - t_{j-1}^\diamond$ has on estimation results.

As opposed to the GP formulation, splines are not directly motivated by a specific dynamic model. However, it is still possible to assume that the motion of the system obeys one of the models in Section~\ref{sec:motion_priors} and to constrain the spline using the motion priors~\eqref{eq:lin_motion_priors},~\eqref{eq:mp_lie}. We have chosen to use these motion priors rather than those derived in~\cite{Furgale2012} in order to more directly compare splines with GPs.

\begin{figure}
    \centering
    \includegraphics[width=0.45\textwidth]{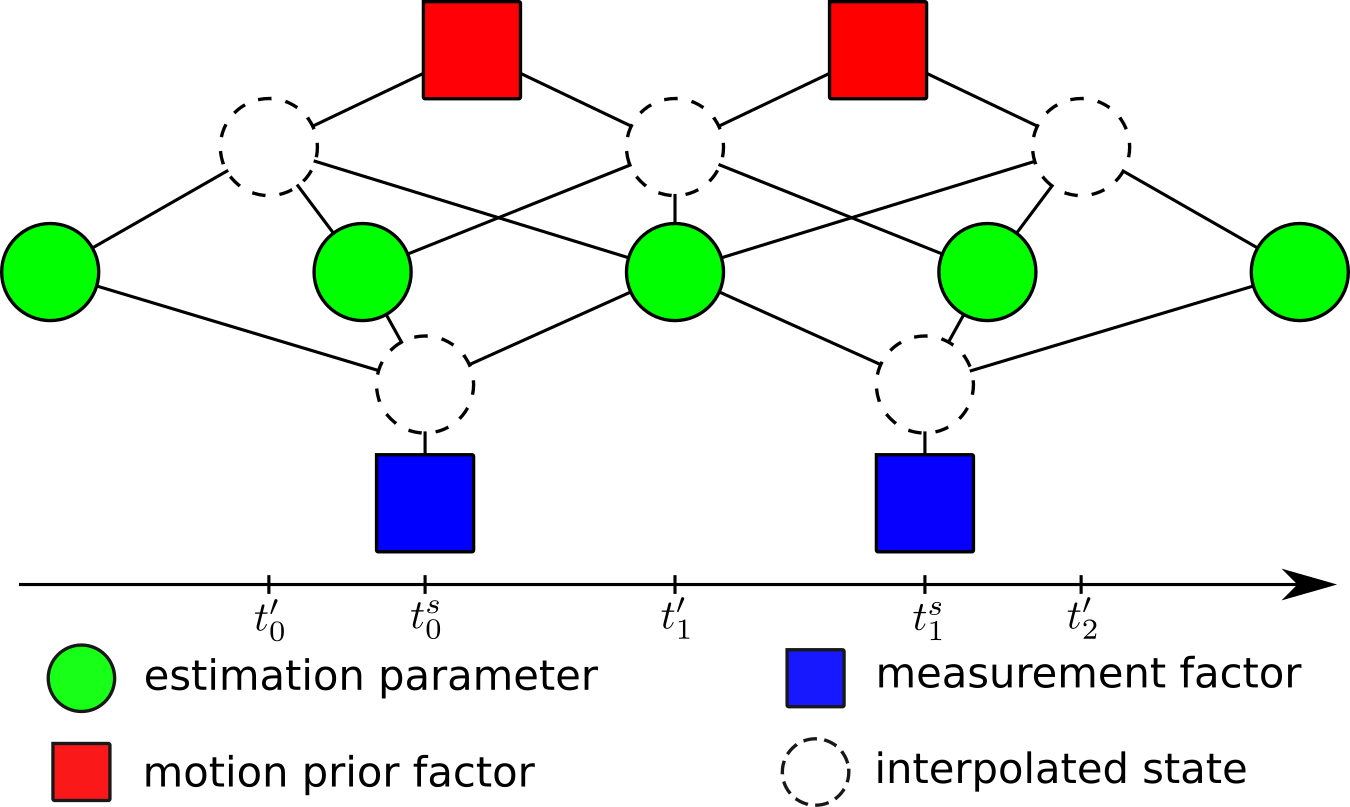}
    \caption{Example factor graph for a spline-based estimator with $k=3$. Note the relative sparsity between this graph and the general graph in Figure~\ref{fig:ct_factor_graph}.}
    \label{fig:spline_factor_graph}
\end{figure}

There is not an obvious way to choose the motion sample times $\{ t_j^\prime \}$ to ensure that the control points are fully constrained. In our analysis for vector space splines in Section~\ref{sec:lin_sim}, we will show the effect that the motion sample period $\delta t^\prime = t_j^\prime - t_{j-1}^\prime$ has on estimation results. An example factor graph for spline-based estimation with $k=3$ is given in Figure~\ref{fig:spline_factor_graph}.

To perform estimation using splines, we will need to compute the Jacobians of the interpolation function with respect to the control points\footnote{In the case of vector space splines, the interpolation function is linear, so these Jacobians are constant.}. These Jacobians are derived in Appendix~\ref{sec:spl_jacs}.

    \section{Estimation Scenarios}
\label{sec:example_problems}

We aim to compare GP-based estimation against spline-based estimation in a couple of scenarios. Our first scenario is a simple linear model introduced in Section~\ref{sec:lin_est}. This will allow us to understand how the various hyper-parameters of each estimation scheme affect the estimated solution at a high level. Next, we will look at a more complicated system that fuses low frequency fiducial pose measurements obtained from a monocular camera with high frequency IMU measurements. The configuration manifold of the camera is the Lie group SE(3). This will allow us to compare the GP and spline Lie group adaptations directly. This scenario is introduced in Section~\ref{sec:cam_imu_est}.

\subsection{Linear Estimation}
\label{sec:lin_est}

Let the state of a linear system be $\mathbf{x}(t) = \begin{bmatrix} d(t)^\top & v(t)^\top & a(t)^\top \end{bmatrix}^\top$, where $d(t), v(t), a(t) \in \mathbb{R}^2$ are respectively the position, velocity, and acceleration of the system, and assume that it evolves according to
\begin{equation}
    \dot{\mathbf{x}}(t) = \begin{bmatrix} \mathbf{0} & \mathbf{I} & \mathbf{0} \\ \mathbf{0} & \mathbf{0} & \mathbf{I} \\ \mathbf{0} & \mathbf{0} & \mathbf{0} \end{bmatrix} \mathbf{x}(t) + \begin{bmatrix} \mathbf{0} \\ \mathbf{0} \\ \mathbf{I} \end{bmatrix}\mathbf{w}(t)
\end{equation}
(i.e., a WNOJ model), where $\mathbf{w}(t) \sim \mathcal{GP}(\mathbf{0}, \delta (t - t^\prime) \mathbf{Q})$. The system receives measurements 
\begin{equation}
    \mathbf{z}_i^s = d(t_i^s) + \boldsymbol{\eta},
\end{equation}
where $t_i^s$ is the measurement time and $\boldsymbol{\eta} \sim \mathcal{N}(\mathbf{0}, \mathbf{R})$ is zero-mean Gaussian noise with covariance $\mathbf{R}$. Because the measurement function and the system are both linear and defined on a vector space, the estimation problem~\eqref{eq:problem} can be solved analytically using linear least squares for both GPs and splines. In Section~\ref{sec:lin_sim} we will provide an in-depth comparison between the performance of GP-based and spline-based estimation for this system.

\subsection{Camera and IMU Estimation}
\label{sec:cam_imu_est}

Let the pose of an IMU be $\mathbf{T}_\mathcal{I}^\mathcal{B}(t) \in \text{SE}(3)$, where $\mathcal{I}$ represents an inertial coordinate frame, $\mathcal{B}$ represents the IMU coordinate frame, and $\mathbf{T}_\mathcal{I}^\mathcal{B}$ is the homogeneous transformation from $\mathcal{I}$ to $\mathcal{B}$. The IMU pose has dynamics
\begin{equation}
\begin{gathered}
    \dot{\mathbf{T}}_\mathcal{I}^\mathcal{B}(t) = - \boldsymbol{\varpi}_{\mathcal{B}/\mathcal{I}}^\mathcal{B} (t)^\wedge \mathbf{T}_\mathcal{I}^\mathcal{B} (t), \\
    \dot{\boldsymbol{\varpi}}_{\mathcal{B}/\mathcal{I}}^\mathcal{B} (t) = \boldsymbol{\alpha}_{\mathcal{B}/\mathcal{I}}^\mathcal{B} (t), \\
    \dot{\boldsymbol{\alpha}}_{\mathcal{B}/\mathcal{I}}^\mathcal{B} (t) = \mathbf{w}(t),
\end{gathered}
\label{eq:se3_wnoj}
\end{equation}
where 
$\boldsymbol{\varpi}_{\mathcal{B}/\mathcal{I}}^\mathcal{B} = \begin{bmatrix} {\mathbf{v}_{\mathcal{B}/\mathcal{I}}^{\mathcal{B}^\top}} & {\boldsymbol{\omega}_{\mathcal{B}/\mathcal{I}}^{\mathcal{B}^\top}} \end{bmatrix}^\top \in \mathbb{R}^6$ 
is the twist of the IMU with velocity 
$\mathbf{v}_{\mathcal{B}/\mathcal{I}}^\mathcal{B}$ 
and angular velocity 
$\boldsymbol{\omega}_{\mathcal{B}/\mathcal{I}}^\mathcal{B}$, 
and where 
$\boldsymbol{\alpha}_{\mathcal{B}/\mathcal{I}}^\mathcal{B} = \begin{bmatrix} {\mathbf{a}_{\mathcal{B}/\mathcal{I}}^{\mathcal{B}^\top}} & {\dot{\boldsymbol{\omega}}_{\mathcal{B}/\mathcal{I}}^{\mathcal{B}^\top}} \end{bmatrix}^\top \in \mathbb{R}^6$ 
with acceleration 
$\mathbf{a}_{\mathcal{B}/\mathcal{I}}^\mathcal{B}$, and $\mathbf{w}(t) \sim \mathcal{GP}(\mathbf{0}, \delta(t - t') \mathbf{Q})$.

\begin{figure}
    \centering
    \includegraphics[width=0.45\textwidth]{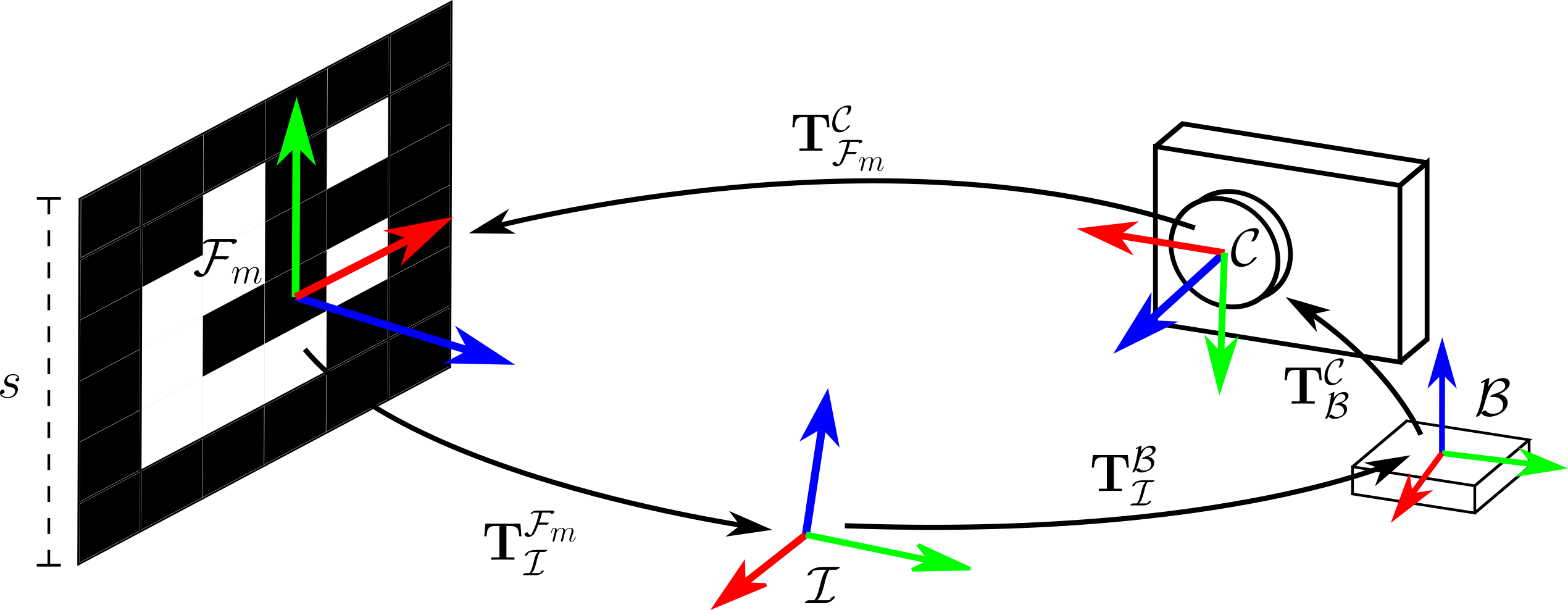}
    \caption{Relevant coordinate frames for the camera and IMU scenario.}
    \label{fig:cam_imu_est}
\end{figure}

Additionally, there is a monocular camera with coordinate frame $\mathcal{C}$ that is rigidly attached to the IMU with precalibrated relative transformation $\mathbf{T}_\mathcal{B}^\mathcal{C} \in \text{SE}(3)$. The camera views fiducial markers with coordinate frames $\mathcal{F}_m, m = 1, \cdots, M$ and known inertial poses $\mathbf{T}_\mathcal{I}^{\mathcal{F}_m}$ and, using the Perspective-N-Point (PnP) algorithm~\cite{Fischler1981}, finds the relative pose between the marker and the camera $\mathbf{T}_{\mathcal{F}_m}^\mathcal{C}$. This scenario is depicted in Figure~\ref{fig:cam_imu_est}.

The measurement models for the gyroscope and accelerometer are
\begin{equation}
    \mathbf{z}_i^g = \boldsymbol{\omega}_{\mathcal{B}/\mathcal{I}}^\mathcal{B}(t_i^g) + \mathbf{b}^g + \boldsymbol{\eta}^g
\end{equation}
and
\begin{equation}
    \mathbf{z}_i^a = \mathbf{a}_{\mathcal{B}/\mathcal{I}}^\mathcal{B}(t_i^a) - g \mathbf{R}_\mathcal{I}^\mathcal{B}(t_i^a) \mathbf{e}_3 + \mathbf{b}^a + \boldsymbol{\eta}^a,
\end{equation}
respectively, where $t_i^g$ and $t_i^a$ are the measurement timestamps, $\boldsymbol{\eta}^g \sim \mathcal{N}(\mathbf{0}, \boldsymbol{\Sigma}^g)$ and $\boldsymbol{\eta}^a \sim \mathcal{N}(\mathbf{0}, \boldsymbol{\Sigma}^a)$ are additive Gaussian noise with covariance $\boldsymbol{\Sigma}^g$ and $\boldsymbol{\Sigma}^a$ respectively, $\mathbf{b}^g, \mathbf{b}^a \in \mathbb{R}^3$ are walking bias terms with dynamics 
\begin{equation}
    \dot{\mathbf{b}}^g = \boldsymbol{\eta}^{bg}, \quad \dot{\mathbf{b}}^a = \boldsymbol{\eta}^{ba},
\end{equation} 
where $\boldsymbol{\eta}^{bg} \sim \mathcal{N}(\mathbf{0}, \boldsymbol{\Sigma}^{bg}), \ \boldsymbol{\eta}^{ba} \sim \mathcal{N}(\mathbf{0}, \boldsymbol{\Sigma}^{ba})$, and where $g$ is the constant of gravitational acceleration. The fiducial pose measurement model is
\begin{equation}
    \mathbf{z}_{i_m}^c = \text{Exp}(\boldsymbol{\eta}^c) \mathbf{T}_\mathcal{B}^\mathcal{C} \mathbf{T}_\mathcal{I}^\mathcal{B}(t_{i_m}^c) {\mathbf{T}_\mathcal{I}^{\mathcal{F}_m}}^{-1},
\end{equation}
where $t_{i_m}^c$ is the measurement timestamp and $\boldsymbol{\eta^c} \sim \mathcal{N}(\mathbf{0}, \boldsymbol{\Sigma}^c)$. The fiducial pose covariance $\boldsymbol{\Sigma}^c$ varies depending on the viewing angle and distance. We compute it for each individual measurement by propagating the pixel projection covariance $\boldsymbol{\Sigma}^p \in \mathbb{R}^{2\times 2}$ through the PnP model using the fiducial side length $s$ and the camera projection matrix $\mathbf{K} \in \mathbb{R}^{3 \times 3}$. 

The dynamic state of this system resides in the Lie group $\text{SE(3)}$, so it will be necessary to use a Lie group GP or a Lie group spline to solve this problem. We provide in-depth comparisons for these two estimation methods using simulated measurements and real measurements in Sections~\ref{sec:imu_sim} and~\ref{sec:imu_hw}, respectively.
    \section{Linear Simulation Comparison}
\label{sec:lin_sim}

For our linear simulation comparison, we implemented the system in Section~\ref{sec:lin_est} with the following parameters:
\begin{equation}
    \mathbf{Q} = \begin{bmatrix} 1.0 & 0 \\ 0 & 0.01\end{bmatrix} \si{(m/s^3)^2}, \quad \mathbf{R} = 0.01^2 \mathbf{I} \: \si{m^2}.
\end{equation}
Measurements were simulated at 100~\si{Hz} for 20~\si{s} with initial conditions $d(t_0) = \mathbf{0}$~\si{m}, $v(t_0) = \begin{bmatrix} 1.0, 0.0\end{bmatrix}^\top$~\si{m/s}, and $a(t_0) = \mathbf{0}$~\si{m/s^2}. 

\begin{figure}
    \centering
    \includegraphics[width=0.45\textwidth]{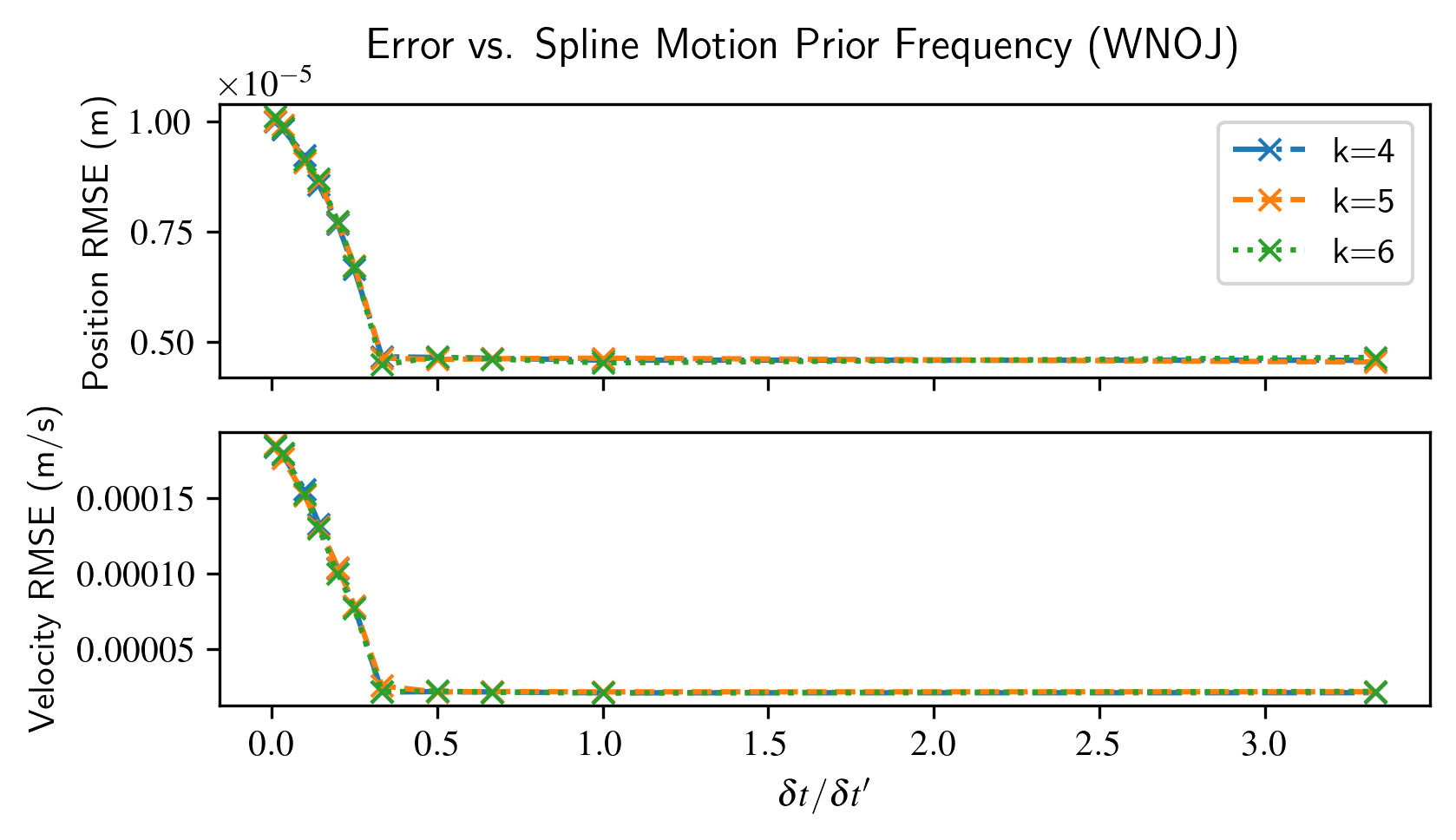}
    \caption{Spline-based trajectory error versus spline motion prior frequency for several values of $k$. Both the trajectory and motion priors used a WNOJ model. The values shown are the medians over 50 Monte Carlo trials.}
    \label{fig:spline_mp_freq_wnoj}
\end{figure}

\begin{figure}
    \centering
    \includegraphics[width=0.45\textwidth]{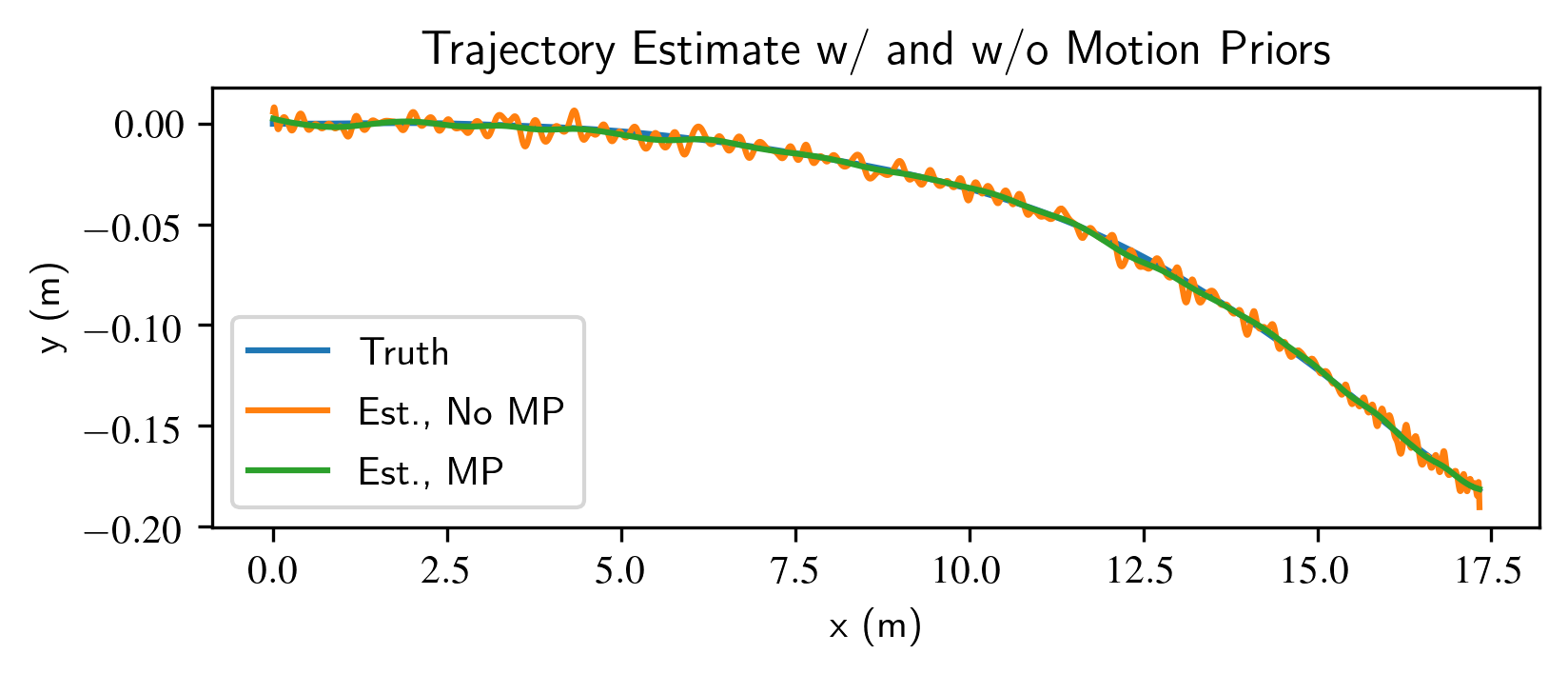}
    \caption{Spline-based trajectory estimation with and without motion priors.}
    \label{fig:traj_comp_prior}
\end{figure}

We wish to compare the performance of GP-based estimation to spline-based estimation for this scenario. In order to do this, we first need to know how to select the spline motion sample period $\delta t^\prime$ to fully constrain the motion of the system. Figure~\ref{fig:spline_mp_freq_wnoj} shows the position and velocity root-mean-square-error (RMSE) for the spline trajectory estimate for varying values of the spline order $k$ and the ratio between the knot period $\delta t$ and $\delta t^\prime$. The values shown are the medians over 50 Monte Carlo trials. As can be seen, the RMSE stops improving as soon as $\delta t / \delta t^\prime = \frac{1}{3}$, showing that the motion is fully constrained at this value. This ratio means that there is one motion prior for every three control points. This makes sense because each WNOJ motion prior introduces three constraints on $\mathbf{x}(t)$, so the number of constraints and control points is one-to-one. Figure~\ref{fig:traj_comp_prior} qualitatively shows the effect that the motion priors have on the trajectory estimate.

% \begin{figure}
%     \centering
%     \includegraphics[width=0.45\textwidth]{figs/spline_mp_freq_wnoa.png}
%     \caption{Trajectory error versus spline motion prior frequency for several values of $k$. Both the trajectory and motion priors used a WNOA model. The values shown are the medians over 50 Monte Carlo trials.}
%     \label{fig:spline_mp_freq_wnoa}
% \end{figure}

\begin{figure}
    \centering
    \includegraphics[width=0.45\textwidth]{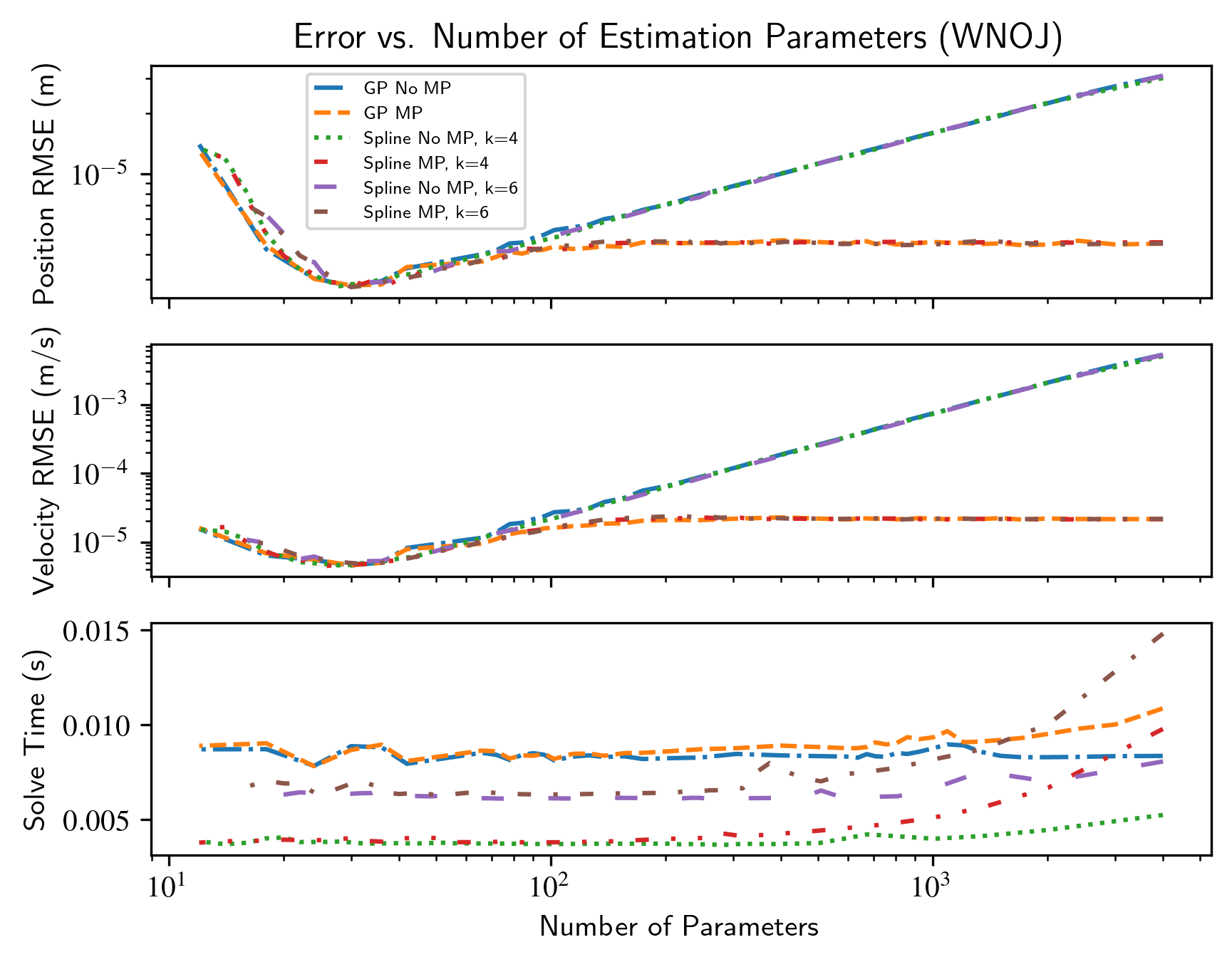}
    \caption{Trajectory error and solve time versus number of floating point estimation parameters for a WNOJ trajectory, for both splines and GPs with and without motion priors. The values shown are the medians over 50 Monte Carlo trials.}
    \label{fig:error_v_params_wnoj}
\end{figure}

With this knowledge in hand, we performed GP-based and spline-based estimation for varying values of $k$, $\delta t$, and $n_\text{GP}$, where $n_\text{GP}$ means that a GP estimation parameter was placed at every $n_\text{GP}$-th measurement time. We compared each method with and without motion priors to show the effect that the cost terms~\eqref{eq:lin_motion_priors} have on the solution and to determine if one of the methods is less prone to over-fitting measurement noise when no regularization is present. The value of $\delta t^\prime$ was chosen such that $\delta t / \delta t^\prime = \frac{1}{3}$, and the GP estimator placed one motion prior between each consecutive set of estimation times. The results are shown in Figure~\ref{fig:error_v_params_wnoj}. The horizontal axis shows the total number of floating point values used in the estimation parameters, and the vertical axes show the position RMSE, velocity RMSE, and solve time\footnote{Note that because the system model is linear and Euclidean, the estimation problem has an analytic solution for both splines and GPs. The solve times shown indicate the amount of compute time required to solve the linear system of equations via sparse Cholesky decomposition.}. Without motion priors, the RMSE increased as the number of parameters increased for both GPs and splines, indicating that the estimated trajectories overfit the measurement noise. When motion priors were used, the error leveled out as the number of parameters increased, meaning that the priors prevented the trajectory from fitting the measurement noise\footnote{Note the slight increase in trajectory RMSE in Figure~\ref{fig:error_v_params_wnoj} before leveling out as the number of parameters increases when using motion priors. This behavior is caused by the relative weighting in the motion prior terms versus the measurement terms in~\eqref{eq:problem}. When a low number of parameters are used, the time between parameters is large and the inverse covariance $\mathbf{Q}_{t^\prime_{j+1}}^{-1}$ is significantly smaller than $\boldsymbol{\Sigma}_s^{-1}$, thus the motion priors have little to no effect on the solution. As the number of parameters increases the effect of the motion priors becomes more significant. In the case of Figure~\ref{fig:error_v_params_wnoj}, the trajectory has enough freedom to fit the measurement noise slightly before the cost of the motion prior terms becomes significant compared to the cost of the measurement terms.}. It is especially interesting to note that the trajectory errors are approximately equivalent for GPs and splines regardless of the number of parameters used or the value of the spline order $k$. The solve time tended to increase with the number of estimation parameters, especially when using motion priors. The spline estimate for $k = 6$ using motion priors had the longest solve time, likely because the time complexity of the spline interpolation equation~\eqref{eq:spline_eval_mat} is linear in $k$. However, the increase in order did not provide superior performance in this case, thus it seems that picking a lower value for $k$ would be reasonable.

% There is enough freedom in the motion model to tolerate a little bit of measurement overfitting, and there is a point where you would be better off picking a low number of parameters than using motion priors.

\begin{figure}
    \centering
    \includegraphics[width=0.45\textwidth]{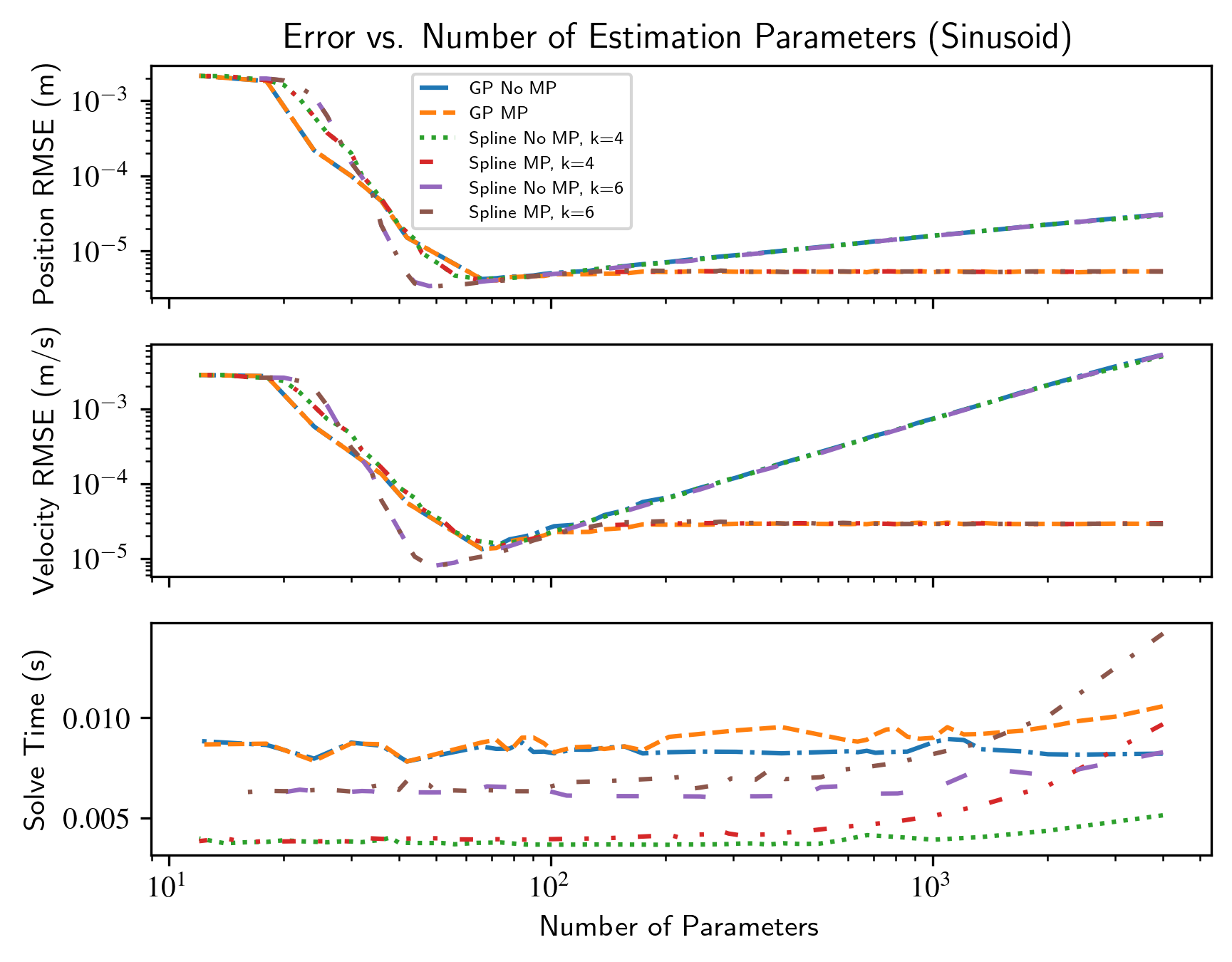}
    \caption{Trajectory error and solve time versus number of floating point estimation parameters for a sinusoidal trajectory, for both splines and GPs with and without motion priors. The motion priors use a WNOJ model. The values shown are the medians over 50 Monte Carlo trials.,}
    \label{fig:error_v_params_sin}
\end{figure}

In addition to WNOJ trajectories, we simulated a circular sinusoidal trajectory in order to show the performance of the two estimation methods when the assumed motion model does not perfectly match the true motion of the system. We used WNOJ motion priors and chose $\mathbf{Q} = \mathbf{I}\si{(m/s^3)^2}$. The results are shown in Figure~\ref{fig:error_v_params_sin}. When motion priors were not used, the error decreased at first as the number of estimation parameters increased, and then increased when the estimated trajectory began to fit the measurement noise. When using motion priors, the error decreased in a similar manner, but then leveled out, indicating that the motion priors restricted the trajectory from overfitting the measurements. Once again, GPs and splines had similar accuracy, regardless of the number of parameters used or the value of $k$. The solve time followed a similar trend to the previous experiment.

    \section{IMU Simulation Comparison}
\label{sec:imu_sim}

We additionally simulated the camera and IMU scenario described in Section~\ref{sec:cam_imu_est}. The following parameters were used:
\begin{equation}
\begin{gathered}
    \mathbf{T}_\mathcal{B}^\mathcal{C} = \begin{bmatrix} \mathbf{I} & \mathbf{p}_{\mathcal{B}/\mathcal{C}}^\mathcal{C} \\ \mathbf{0} & 1 \end{bmatrix}, \quad \mathbf{p}_{\mathcal{B}/\mathcal{C}}^\mathcal{C} = \begin{bmatrix} 
0.0 \\ 0.0 \\ 0.1 \end{bmatrix}~\si{m} \\
    \mathbf{Q} = \begin{bmatrix} 5.0 \mathbf{I} & \mathbf{0} \\ \mathbf{0} & 1.5 \mathbf{I} \end{bmatrix}, \quad \boldsymbol{\Sigma}_g = \sigma_g^2 \mathbf{I}, \quad \sigma_g = 0.005~\si{rad/s} \\
    \boldsymbol{\Sigma}_a = \sigma_a^2 \mathbf{I}, \quad \sigma_a = 0.005~\si{m/s^2}, \\
    \boldsymbol{\Sigma}_{bg} = \sigma_{bg}^2 \mathbf{I}, \quad \sigma_{bg} = 0.005~\si{rad/s^2}, \\
    \boldsymbol{\Sigma}_{ba} = \sigma_{ba}^2 \mathbf{I}, \quad \sigma_{ba} = 0.005~\si{m/s^3}.
\end{gathered}
\end{equation}
We placed a single fiducial marker at the origin of the inertial frame. When simulating fiducial pose measurements from the camera, we either used a constant pose covariance 
\begin{equation}
    \boldsymbol{\Sigma}_c = 0.05 \mathbf{I}
\end{equation}
or we simulated the way noise from the camera propagates to the PnP solution using
\begin{equation}
\begin{gathered}
    \boldsymbol{\Sigma}_p = \sigma_p^2 \mathbf{I}, \quad \sigma_p = 0.5~\si{px}, \\
    \mathbf{K} = \begin{bmatrix} f_x & 0 & c_x \\ 0 & f_y & c_y \\ 0 & 0 & 1 \end{bmatrix}, \quad f_x = f_y = 620~\si{px}, \\ 
    c_x = 320~\si{px}, \quad c_y = 240~\si{px}, \quad s = 0.4~\si{m}.
\end{gathered}
\end{equation}
When the camera trajectory was constrained such that the fiducial marker stayed in the field-of-view (FOV) of the simulated camera, the variable pose covariance was used. Otherwise the constant pose covariance was used, and pose measurements were simulated regardless of whether the fiducial marker was in the FOV. The camera was simulated at 10~\si{Hz} and the IMU was simulated at 100~\si{Hz}.

\begin{figure}
    \centering
    \includegraphics[width=0.4\textwidth]{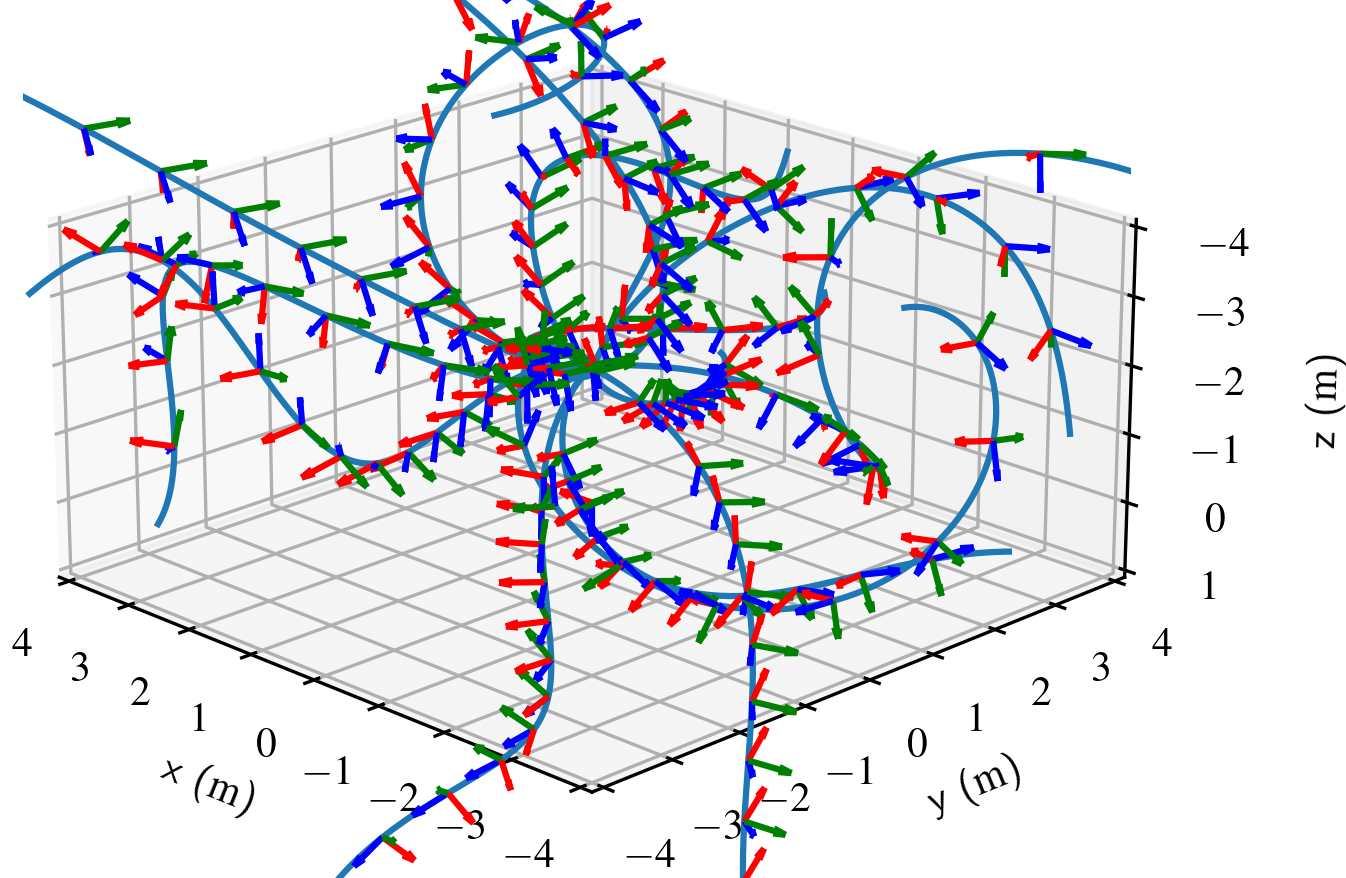}
    \caption{A collection of random WNOJ trajectories on SE(3).}
    \label{fig:wnoj_trajs}
\end{figure}

We simulated 10~\si{s} WNOJ trajectories using the model in~\eqref{eq:se3_wnoj}. A collection of these WNOJ trajectories is shown in Figure~\ref{fig:wnoj_trajs}. Because these trajectories were not constrained to have the fiducial marker be in the camera FOV, the constant pose covariance model was used when simulating pose measurements.

We tested the ability of six different continuous-time trajectory representations to estimate the camera trajectory from the acquired fiducial pose measurements: splines on SE(3) with $k=4$ and $k=6$, GPs on SE(3), splines on SO(3)$\times \mathbb{R}^3$ with $k=4$ and $k=6$, and GPs on SO(3)$\times \mathbb{R}^3$. When estimating on SO(3)$\times \mathbb{R}^3$, the translation and rotation were represented using separate splines or GPs. We included these representations in our analysis because previous works have found that decoupling rotation and translation in the estimation process can improve computational performance~\cite{Ovren2019}. In addition, we varied the type of trajectory regularization that was used during estimation: motion priors, IMU measurements, both, or neither. When including IMU and/or motion priors, we ran estimation in two steps, first without regularization and then with regularization. This is because the estimator struggled to converge with poor initial conditions when including IMU or motion prior residuals. All estimation parameters were initialized to zero or identity. In all cases IMU biases were estimated as constant values (this is reasonable because the time window is relatively short). We used Levenberg-Marquardt optimization implemented with the Ceres Solver~\cite{Agarwal2022}. 

When including motion priors with spline-based estimation, we set the motion prior period $\delta t^\prime$ to three times the knot point period $\delta t$ to be consistent with our findings in Section~\ref{sec:lin_sim}. Rather than vary the number of estimation parameters, we chose fixed values for the knot point period $\delta t = 0.1~\si{s}$ and GP estimation frequency $n_\text{GP} = 1$ (meaning that an estimation time was placed at every $n_\text{GP}$-th fiducial pose measurement). We found that these values provided accurate and consistent results for each of our test scenarios.

\begin{figure*}
    \centering
    \includegraphics[width=0.98\textwidth]{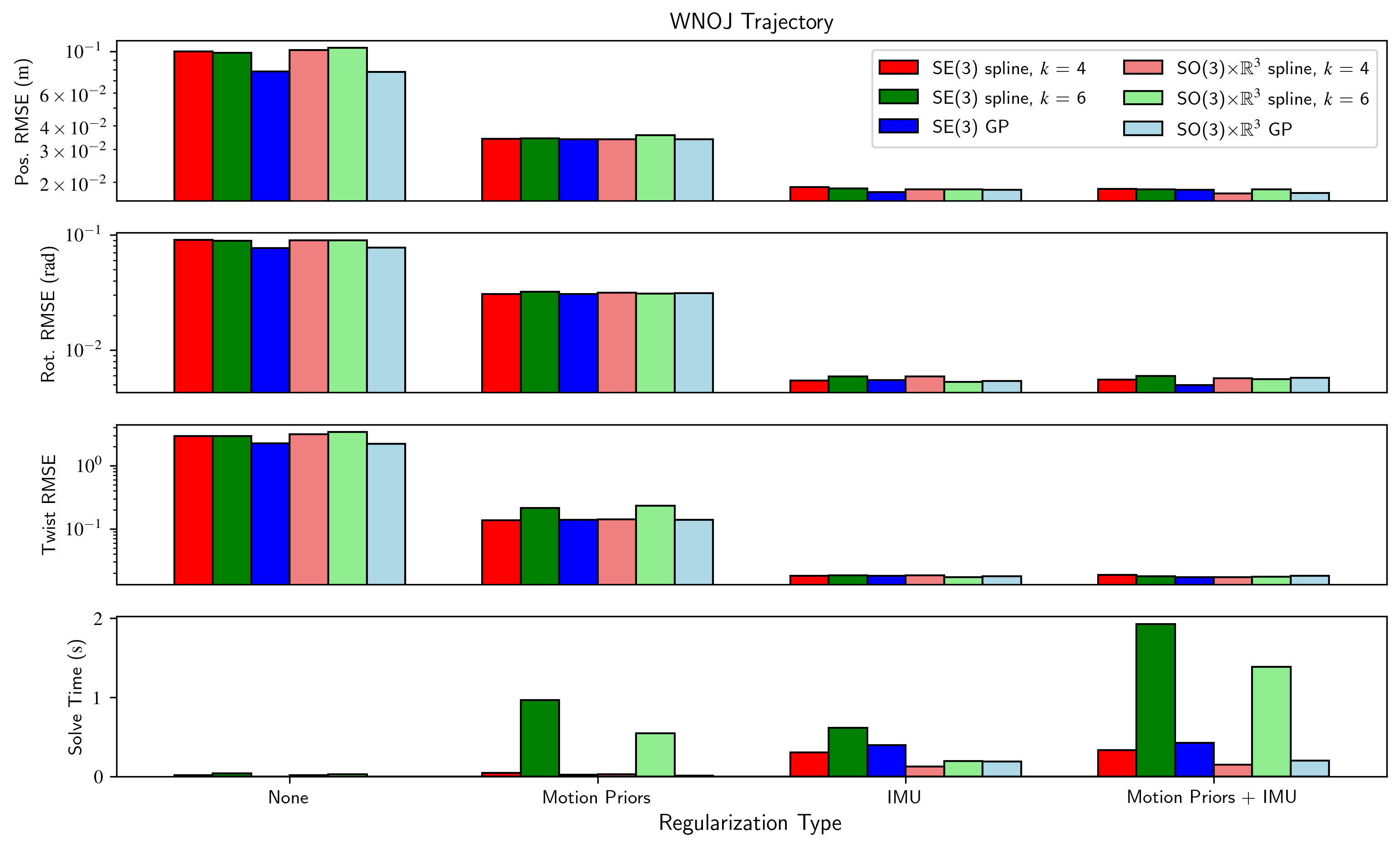}
    \caption{GP and spline trajectory error and solve time for the WNOJ trajectories.}
    \label{fig:wnoj_mc_cam_imu}
\end{figure*}

In each of these scenarios, we ran 100 Monte Carlo trials. The median pose, velocity, and twist RMSE, as well as the computation times for each of these scenarios are shown in Figure~\ref{fig:wnoj_mc_cam_imu}. There are several observations that can be made from this figure. First, all trajectory representations had better accuracy when including motion priors, and the best trajectory accuracy was achieved when using IMU measurements. Using motion priors in addition to the IMU measurements did not improve accuracy. We also note that when using the same type of regularization, each method achieved similar trajectory accuracy. The only exception is that the GP representations appear to have slightly lower error when no regularization is used, although the difference is not overly significant.

\begin{figure}
    \centering
    \includegraphics[width=0.46\textwidth]{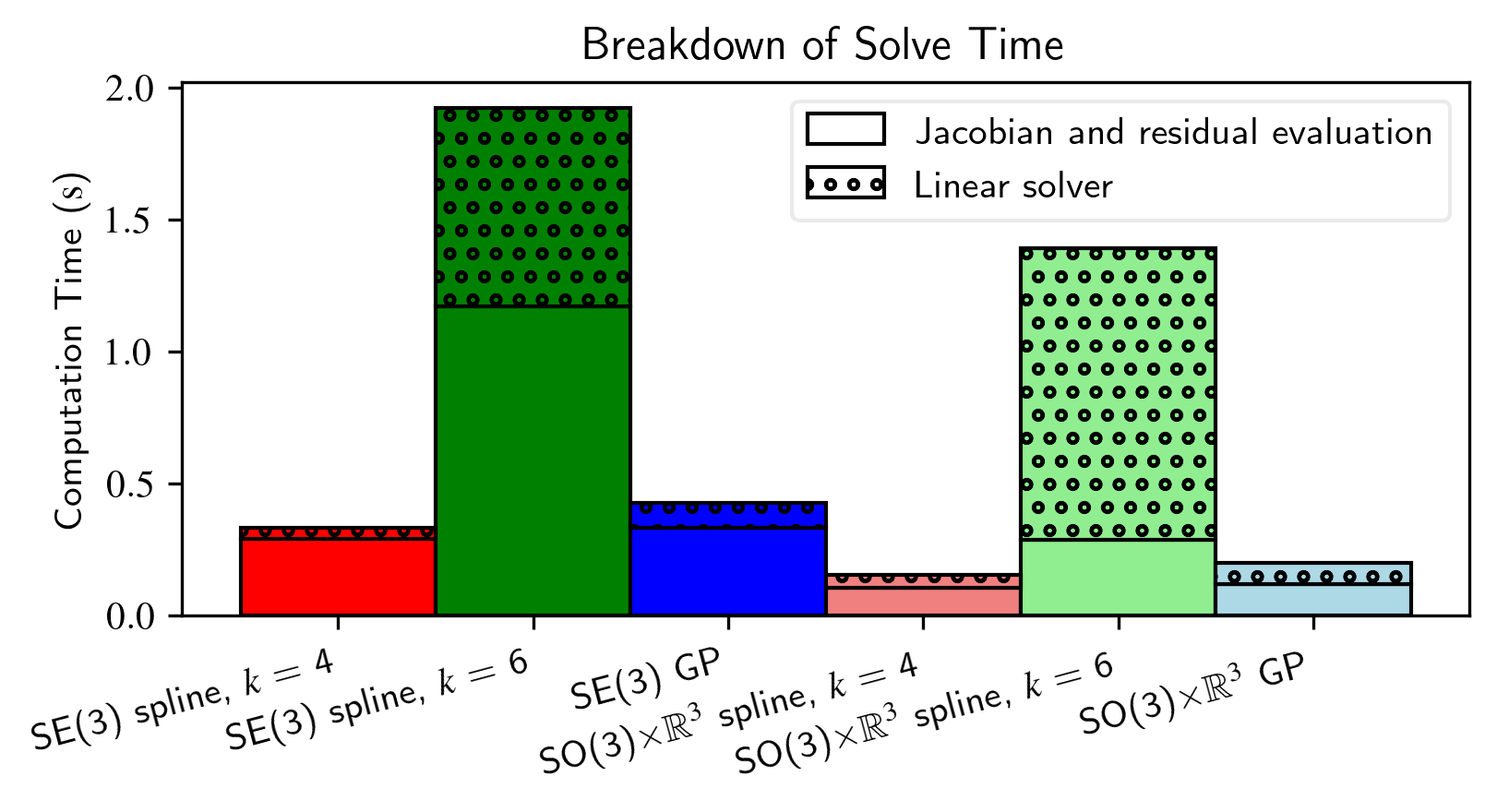}
    \caption{Solve time breakdown for the WNOJ trajectory results when using both motion priors and IMU.}
    \label{fig:solve_time_breakdown}
\end{figure}

\begin{figure}
    \centering
    \includegraphics[width=0.46\textwidth]{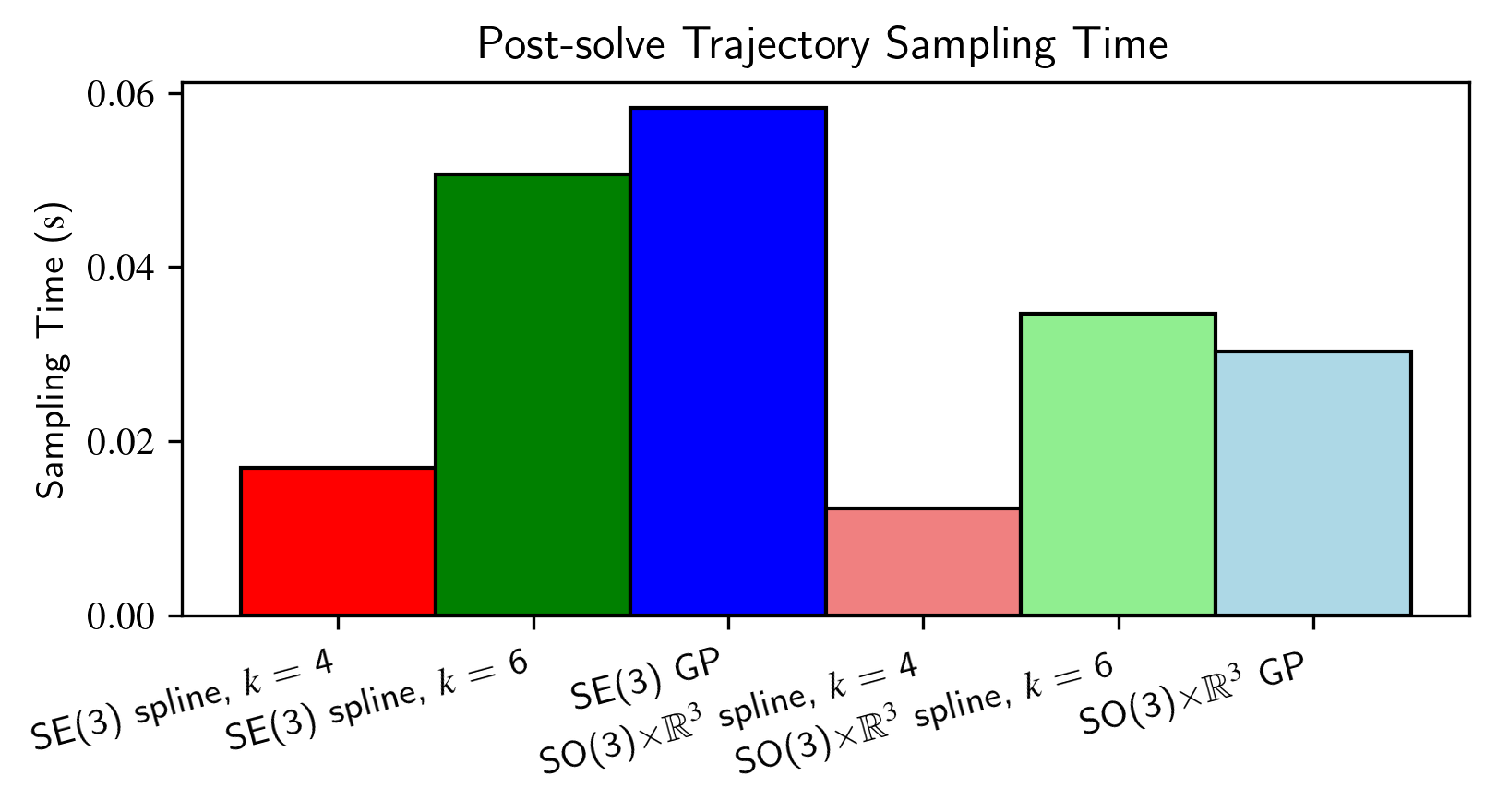}
    \caption{Total post-solve trajectory sampling times. Pose and twist were sampled uniformly along the 10 second trajectories at a rate of 1~\si{kHz}.}
    \label{fig:sampling_time}
\end{figure}

There were, however, significant differences in computation time between the methods. The GP-based methods required less computation time than the spline-based methods when not using IMU measurements. This is because the GP was only evaluated at the estimation times during optimization in this case, so no GP state interpolation~\eqref{eq:gp_global_var} was required\footnote{In this case, the problem is equivalent to discrete-time estimation.}. When including IMU measurements, the solve times for GPs and splines with $k=4$ were comparable. However, splines with $k=6$ required significantly more solve time than the other methods, especially when using motion priors. Figure~\ref{fig:solve_time_breakdown} gives a breakdown of the solve times for WNOJ trajectories when using both motion priors and IMU. The bottom part of the bars shows the amount of time spent evaluating residuals and their Jacobians, and the top part shows the amount of time spent solving the linear system of equations. For splines with $k=6$ most of the time is spent evaluating residuals and Jacobians, but there was also a large amount of time spent solving the linear system of equations. We elaborate more on these results in Section~\ref{sec:discussion}. Finally, Figure~\ref{fig:sampling_time} shows the amount of time required to query the entire trajectory (pose and twist) at a rate of 1~\si{kHz} after optimization. GPs on SE(3) required the most time, and splines with $k=4$ required the least.

\begin{figure*}
     \centering
     \subfigure[None]
     {
         \centering
         \includegraphics[width=0.23\textwidth]{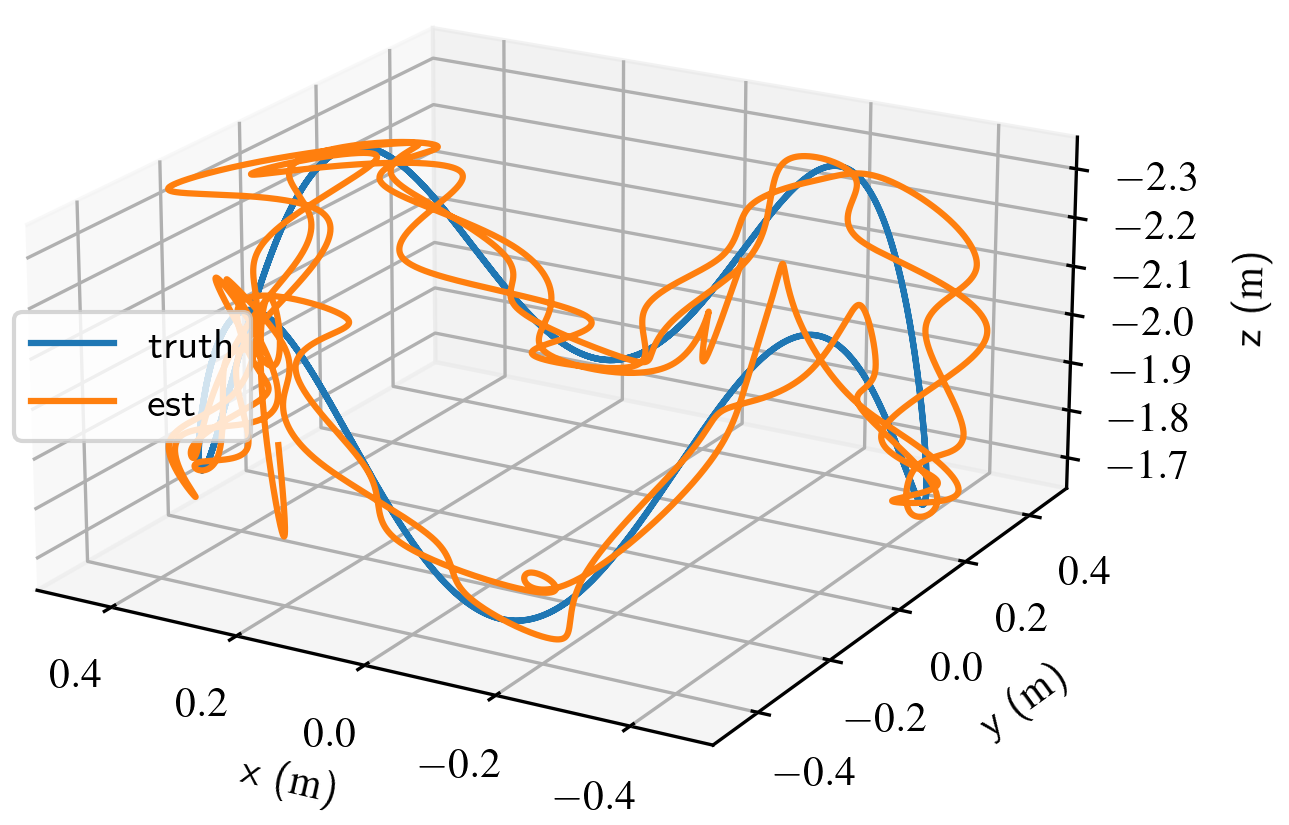}
         \label{fig:sin_spline_none}
     }
     \hfill
     \subfigure[Motion priors]
     {
         \centering
         \includegraphics[width=0.23\textwidth]{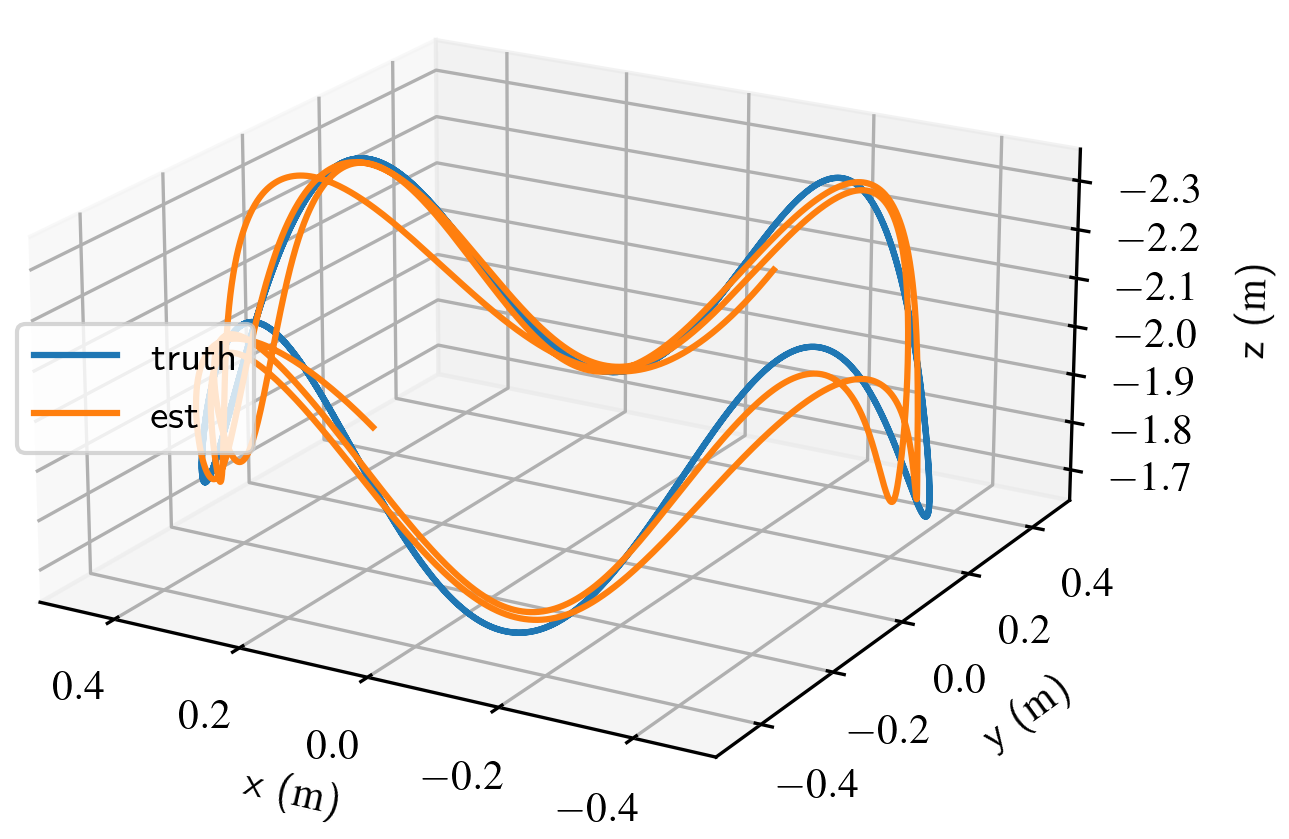}
         \label{fig:sin_spline_mp}
     }
     \hfill
     \subfigure[IMU]
     {
         \centering
         \includegraphics[width=0.23\textwidth]{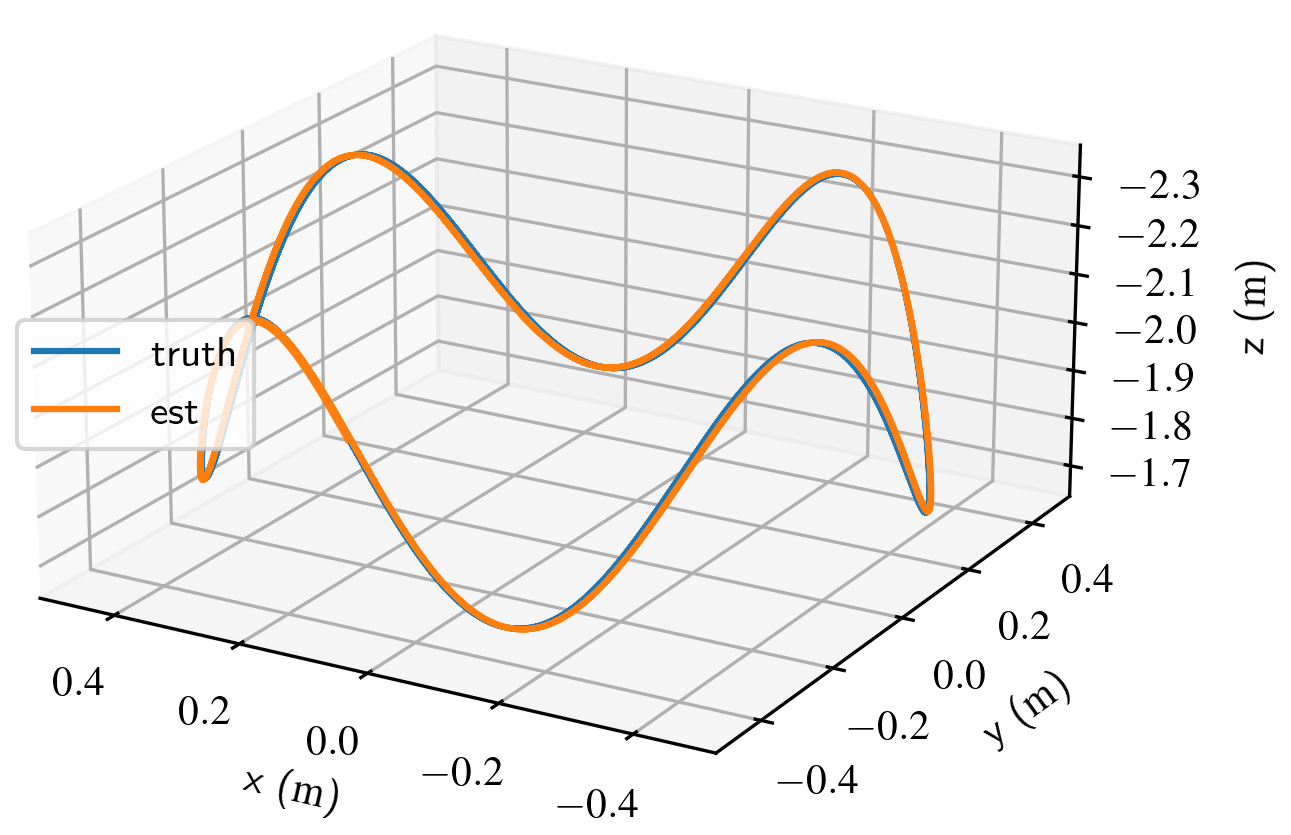}
         \label{fig:sin_spline_imu}
     }
     \hfill
     \subfigure[Motion priors + IMU]
     {
         \centering
         \includegraphics[width=0.23\textwidth]{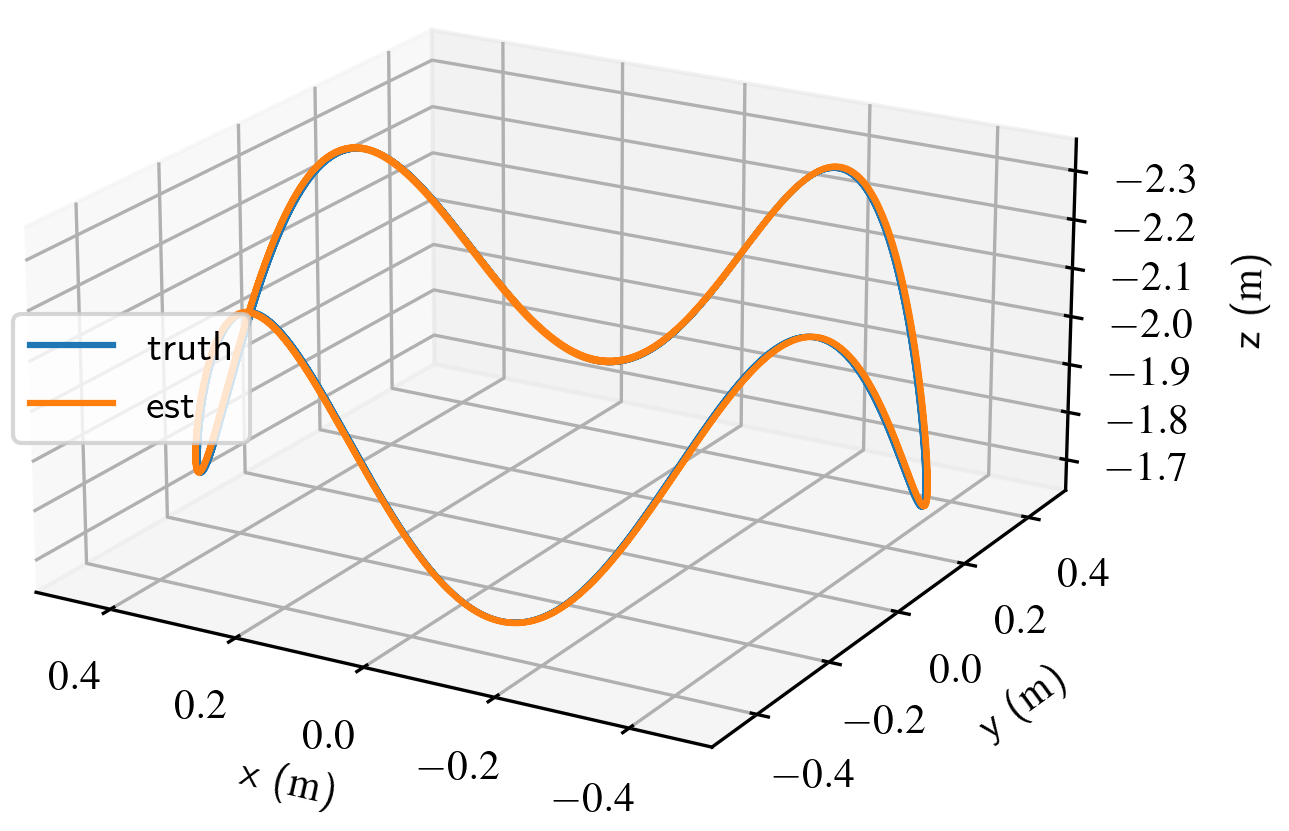}
         \label{fig:sin_spline_imump}
     }
    \caption{SE(3) spline ($k=4$) trajectory estimate for the sinusoidal trajectory with each type of regularization.}
    \label{fig:sin_spline_traj}
\end{figure*}

\begin{figure*}
     \centering
     \subfigure[None]
     {
         \centering
         \includegraphics[width=0.23\textwidth]{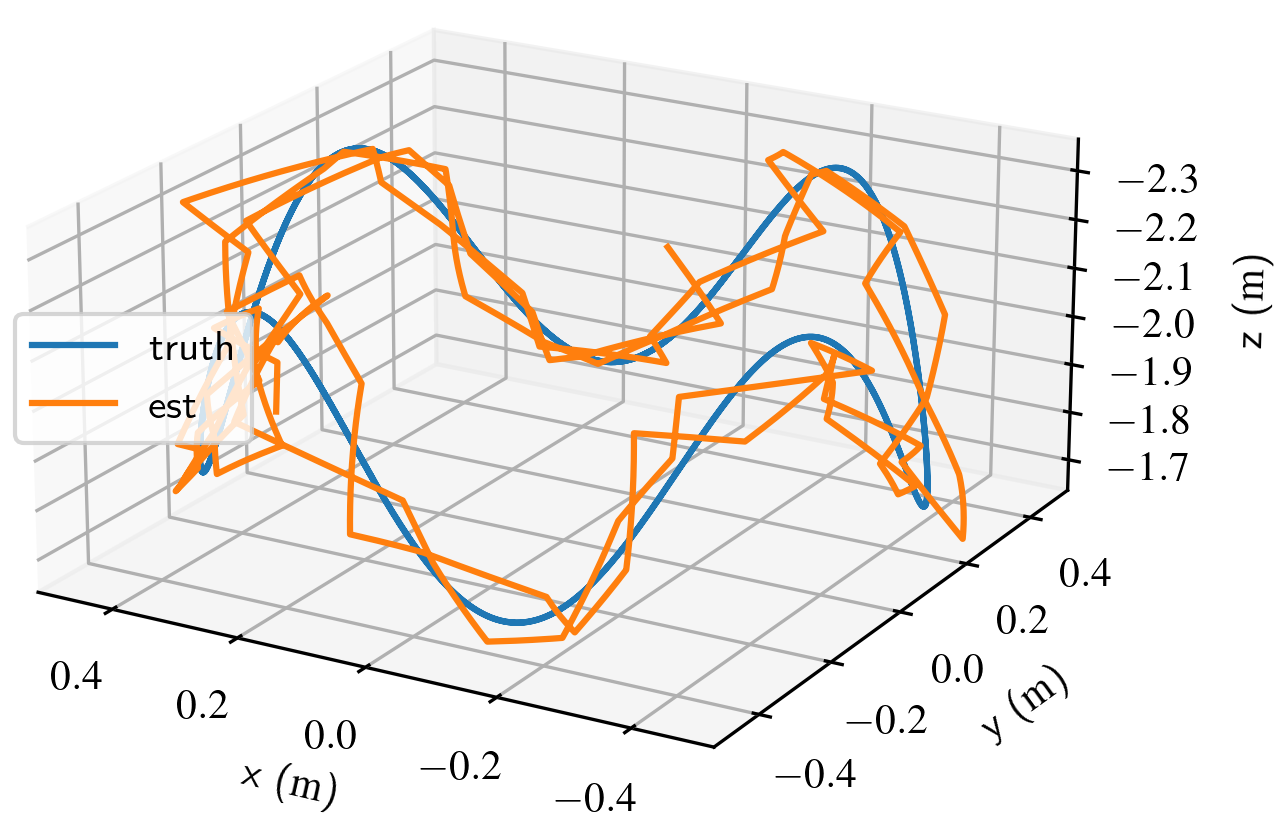}
         \label{fig:sin_gp_none}
     }
     \hfill
     \subfigure[Motion priors]
     {
         \centering
         \includegraphics[width=0.23\textwidth]{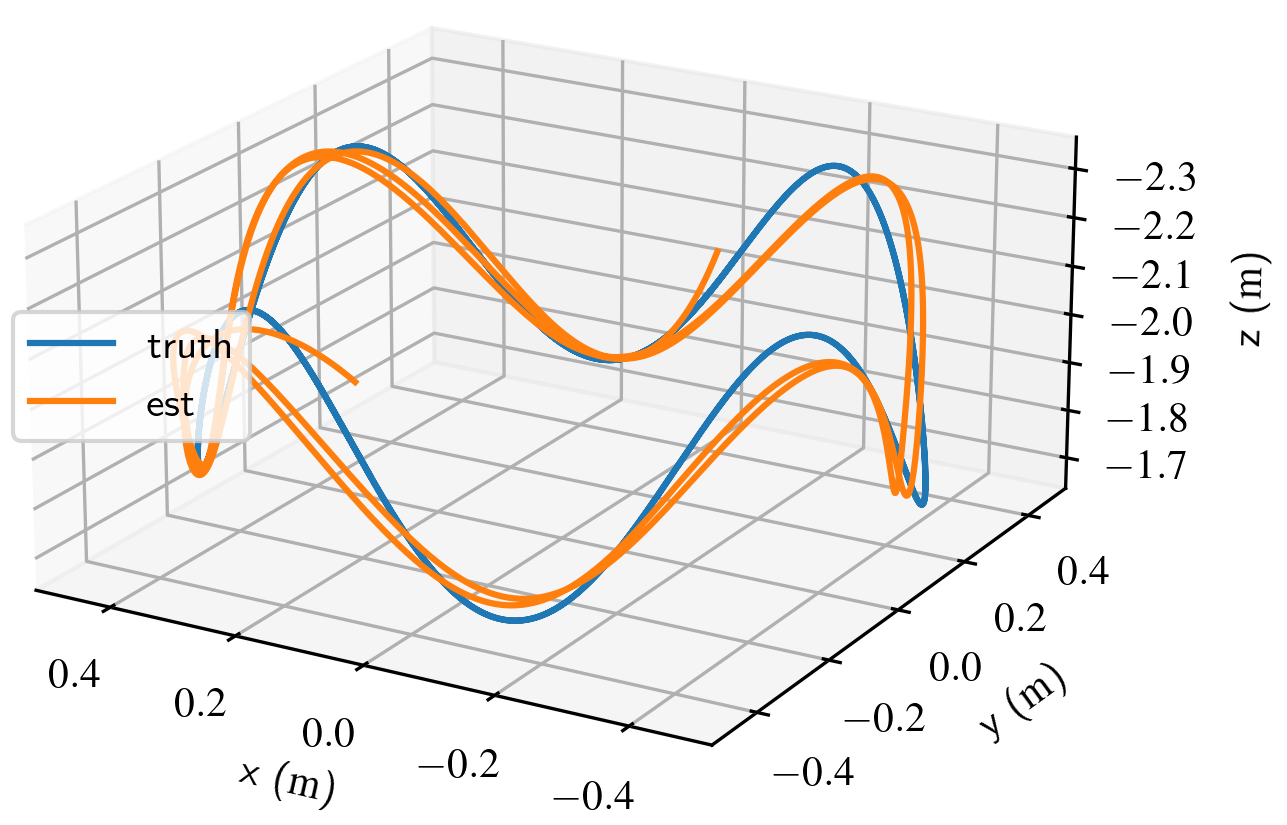}
         \label{fig:sin_gp_mp}
     }
     \hfill
     \subfigure[IMU]
     {
         \centering
         \includegraphics[width=0.23\textwidth]{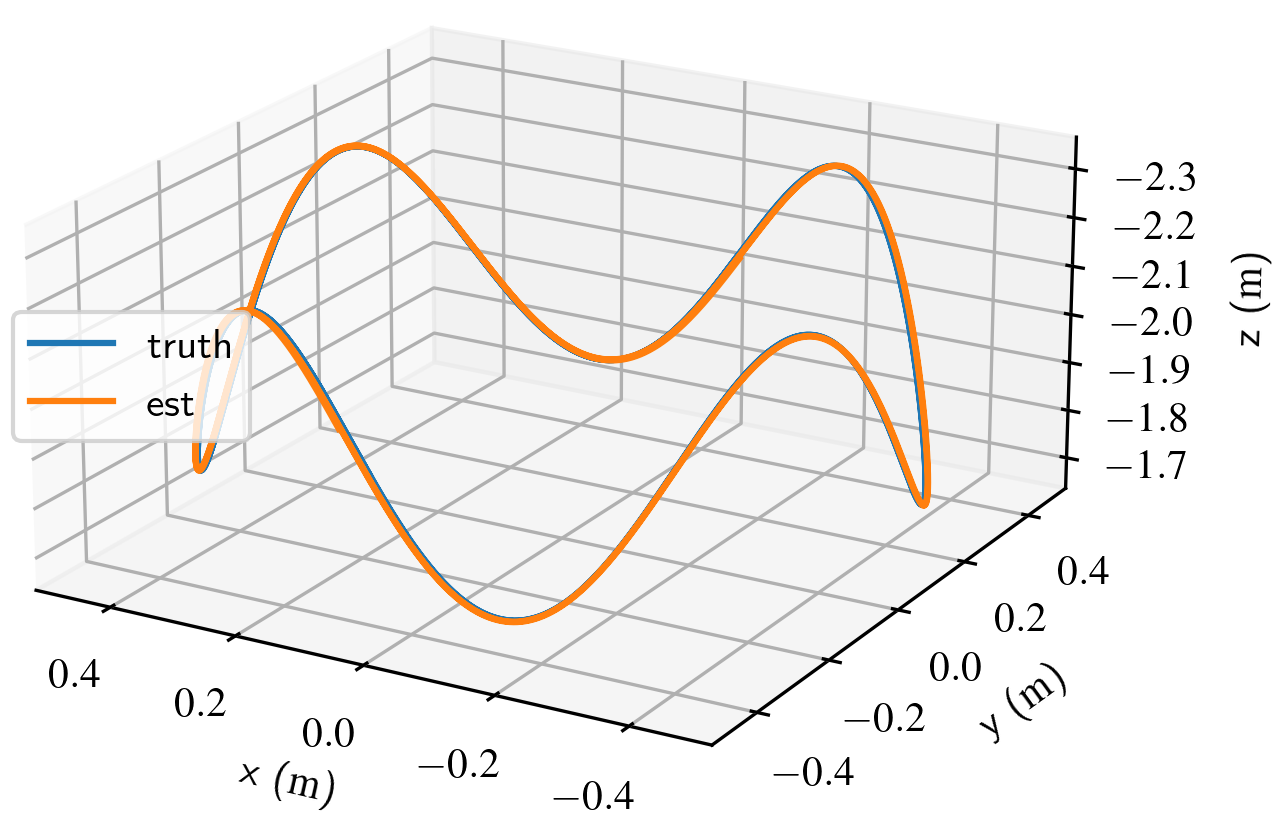}
         \label{fig:sin_gp_imu}
     }
     \hfill
     \subfigure[Motion priors + IMU]
     {
         \centering
         \includegraphics[width=0.23\textwidth]{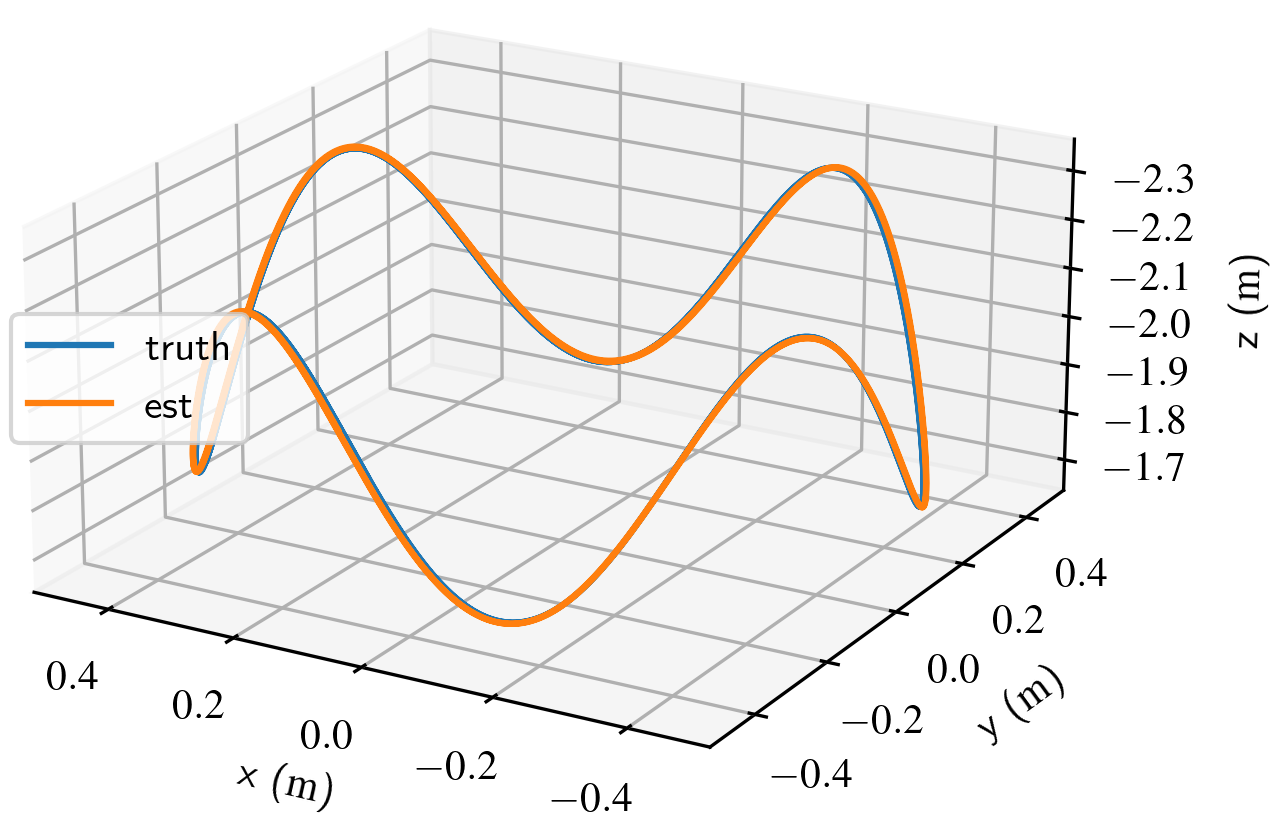}
         \label{fig:sin_gp_imump}
     }
    \caption{SE(3) GP trajectory estimate for the sinusoidal trajectory with each type of regularization.}
    \label{fig:sin_gp_traj}
\end{figure*}

We additionally simulated a sinusoidal trajectory to compare each of these methods when the assumed motion model does not match the true motion of the system. The camera was moved along a circular trajectory such that the fiducial marker was always within the simulated FOV. In this scenario the variable fiducial pose covariance was used when simulating the camera measurements. Figures~\ref{fig:sin_spline_traj} and~\ref{fig:sin_gp_traj} show the resulting estimated trajectories with each type of regularization for a spline on SE(3) with $k=4$ and a GP on SE(3), respectively. It is evident that the motion priors have a profound impact on the smoothness of the estimate. However, the estimated trajectory is much more accurate when using IMU measurements. As we noted previously, using motion priors in addition to the IMU measurements does not seem to offer any improvement. It is interesting to note that, when using the same type of regularization, the spline and GP estimates appear to be very similar.

\begin{figure*}
    \centering
    \includegraphics[width=0.98\textwidth]{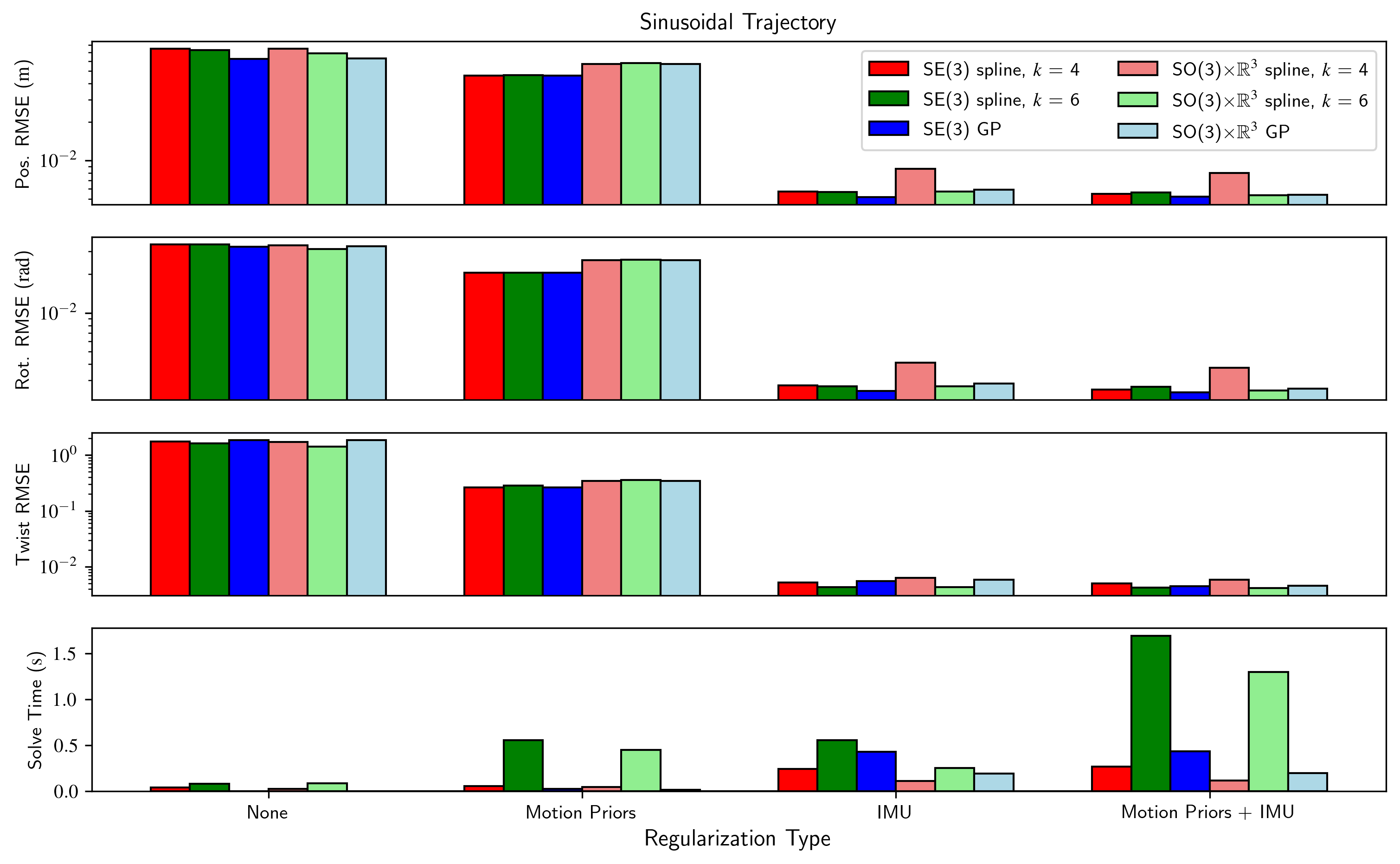}
    \caption{GP and spline trajectory error and solve time for the sinusoidal trajectory.}
    \label{fig:sin_mc_cam_imu}
\end{figure*}

We ran 100 Monte Carlo trials for each estimation scenario on this sinusoidal trajectory. The results are shown in Figure~\ref{fig:sin_mc_cam_imu}. These results are similar and consistent with Figure~\ref{fig:wnoj_mc_cam_imu}. However, the motion priors did not have as significant an effect in lowering the position and rotation RMSE, likely because they did not match the true motion of the system. They did, however, cause a significant reduction in the twist RMSE because of their effect on the smoothness of the trajectory. Interestingly, the SO(3)$\times \mathbb{R}^3$ trajectories often had lower accuracy than the SE(3) trajectories, especially for splines with $k=4$ when using IMU measurements.

    \section{IMU Hardware Comparison}
\label{sec:imu_hw}

\begin{figure*}
     \centering
     \subfigure[None]
     {
         \centering
         \includegraphics[width=0.23\textwidth]{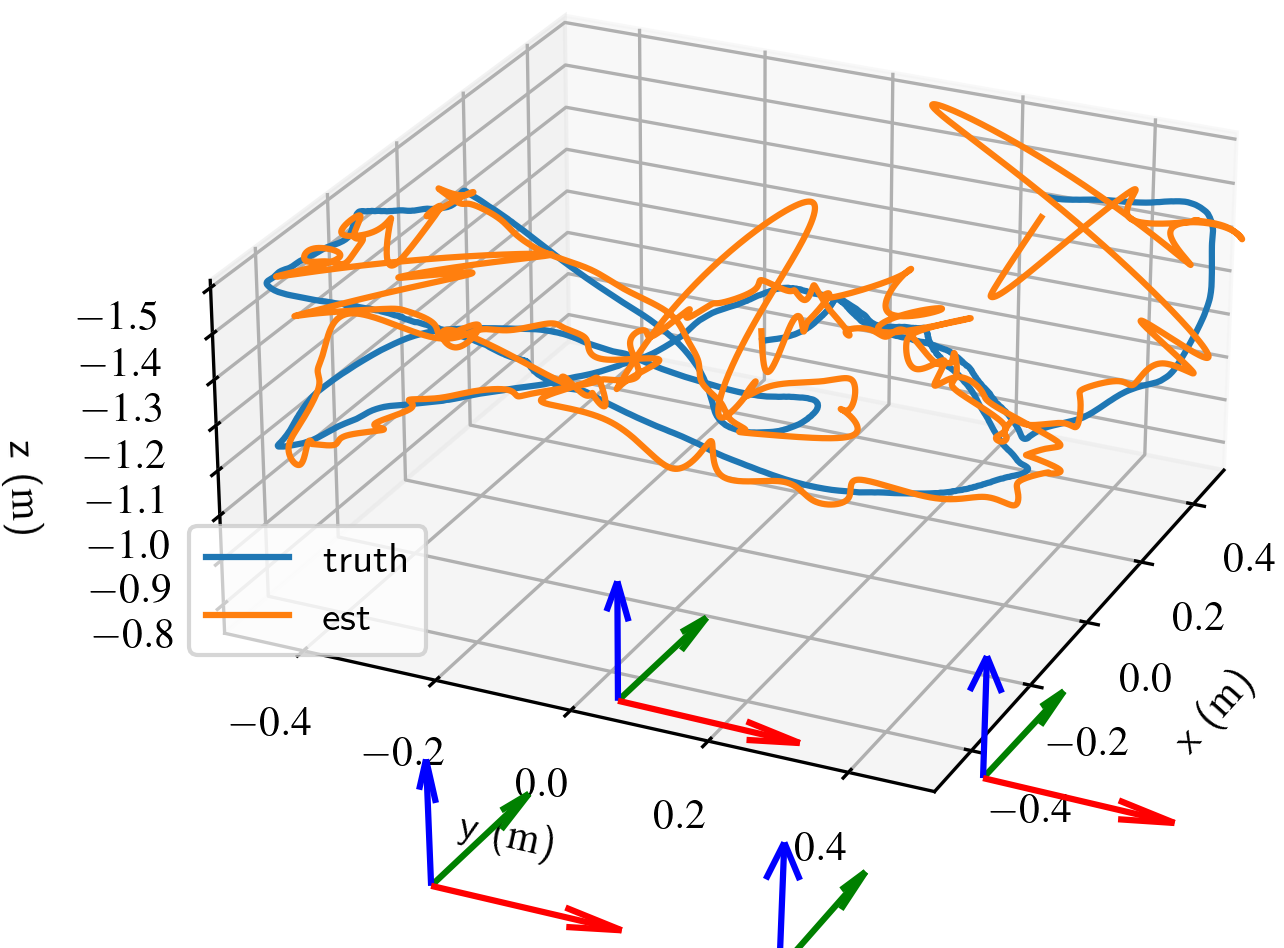}
         \label{fig:hw_spline_none}
     }
     \hfill
     \subfigure[Motion priors]
     {
         \centering
         \includegraphics[width=0.23\textwidth]{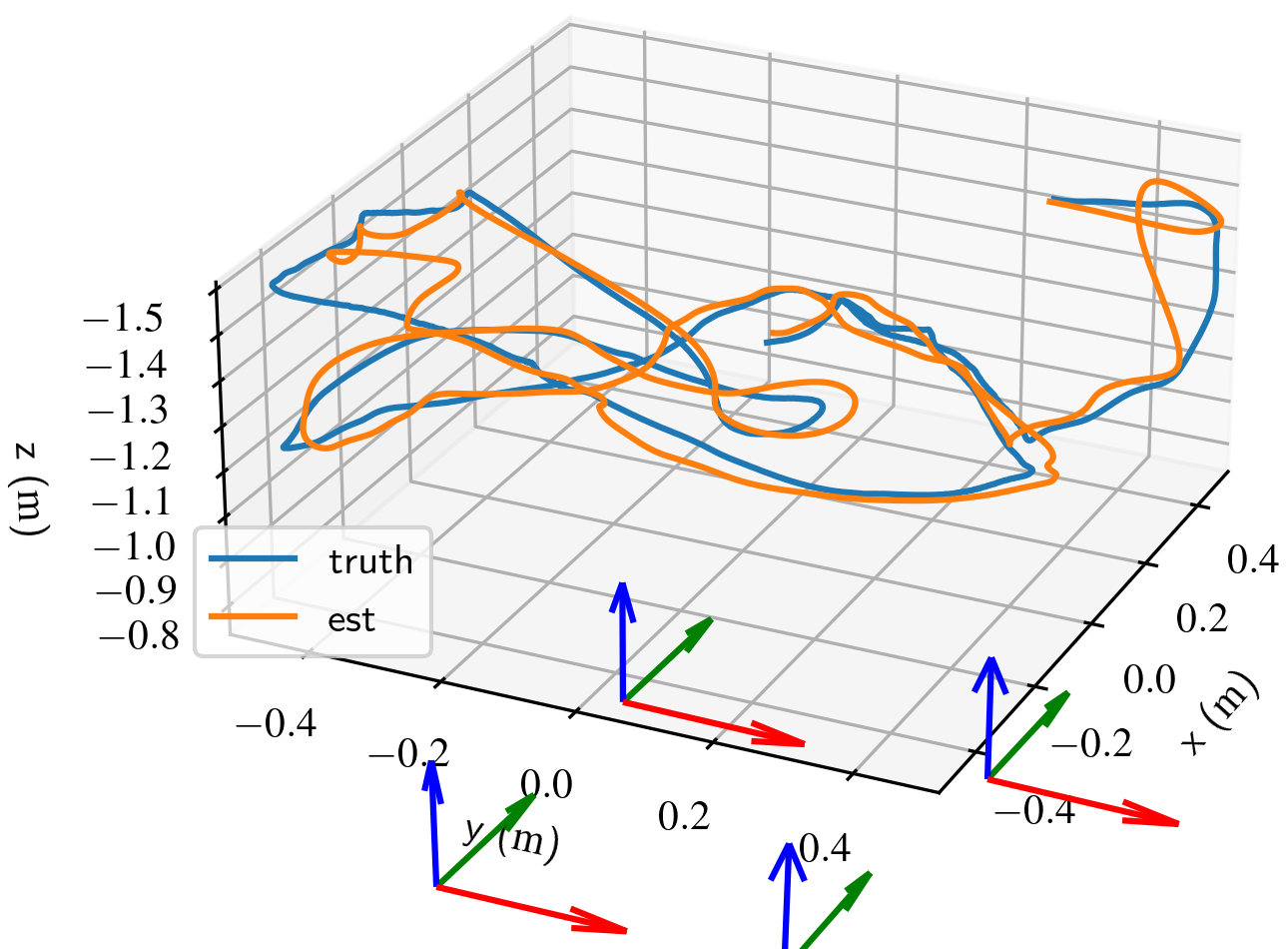}
         \label{fig:hw_spline_mp}
     }
     \hfill
     \subfigure[IMU]
     {
         \centering
         \includegraphics[width=0.23\textwidth]{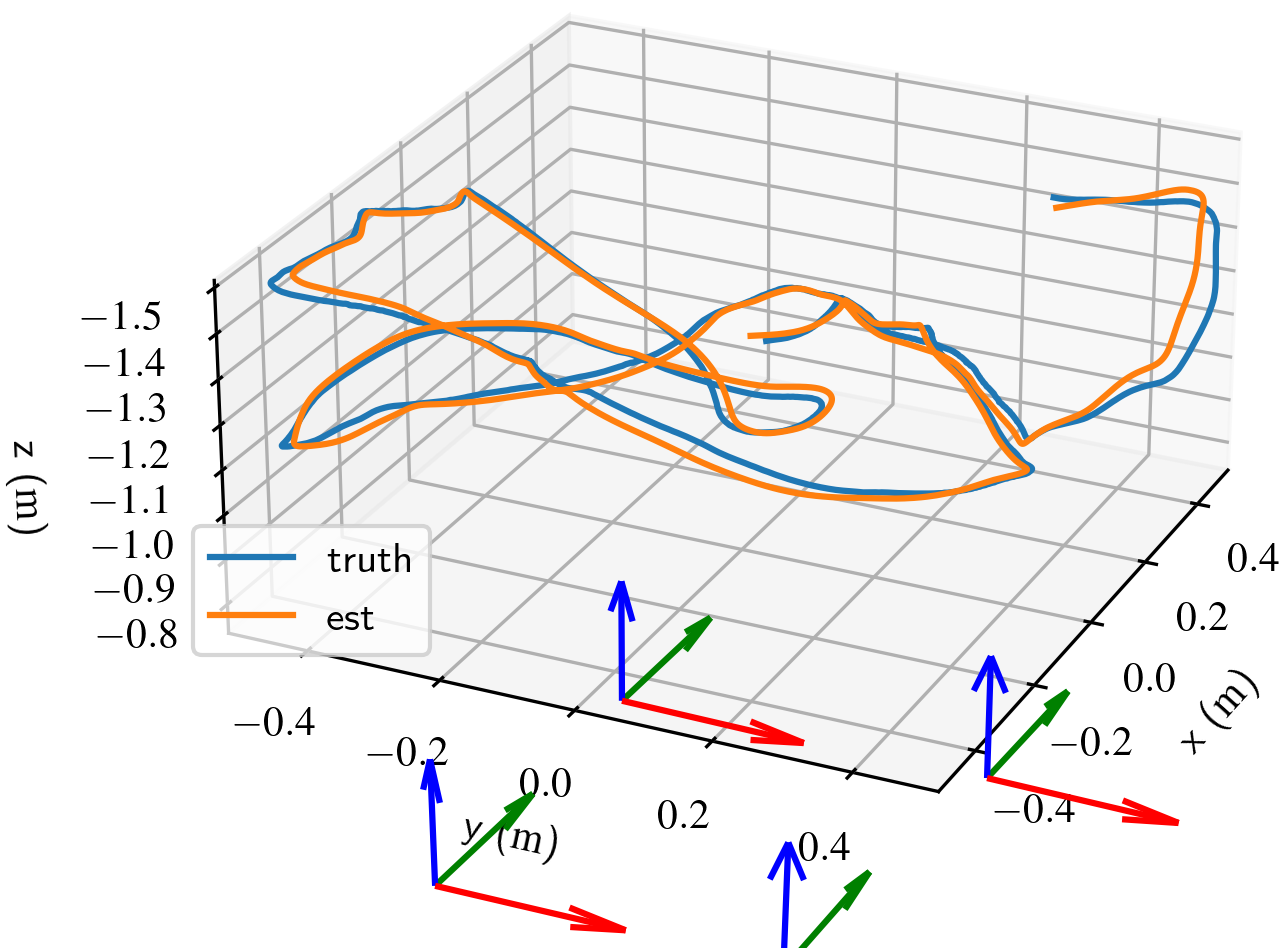}
         \label{fig:hw_spline_imu}
     }
     \hfill
     \subfigure[Motion priors + IMU]
     {
         \centering
         \includegraphics[width=0.23\textwidth]{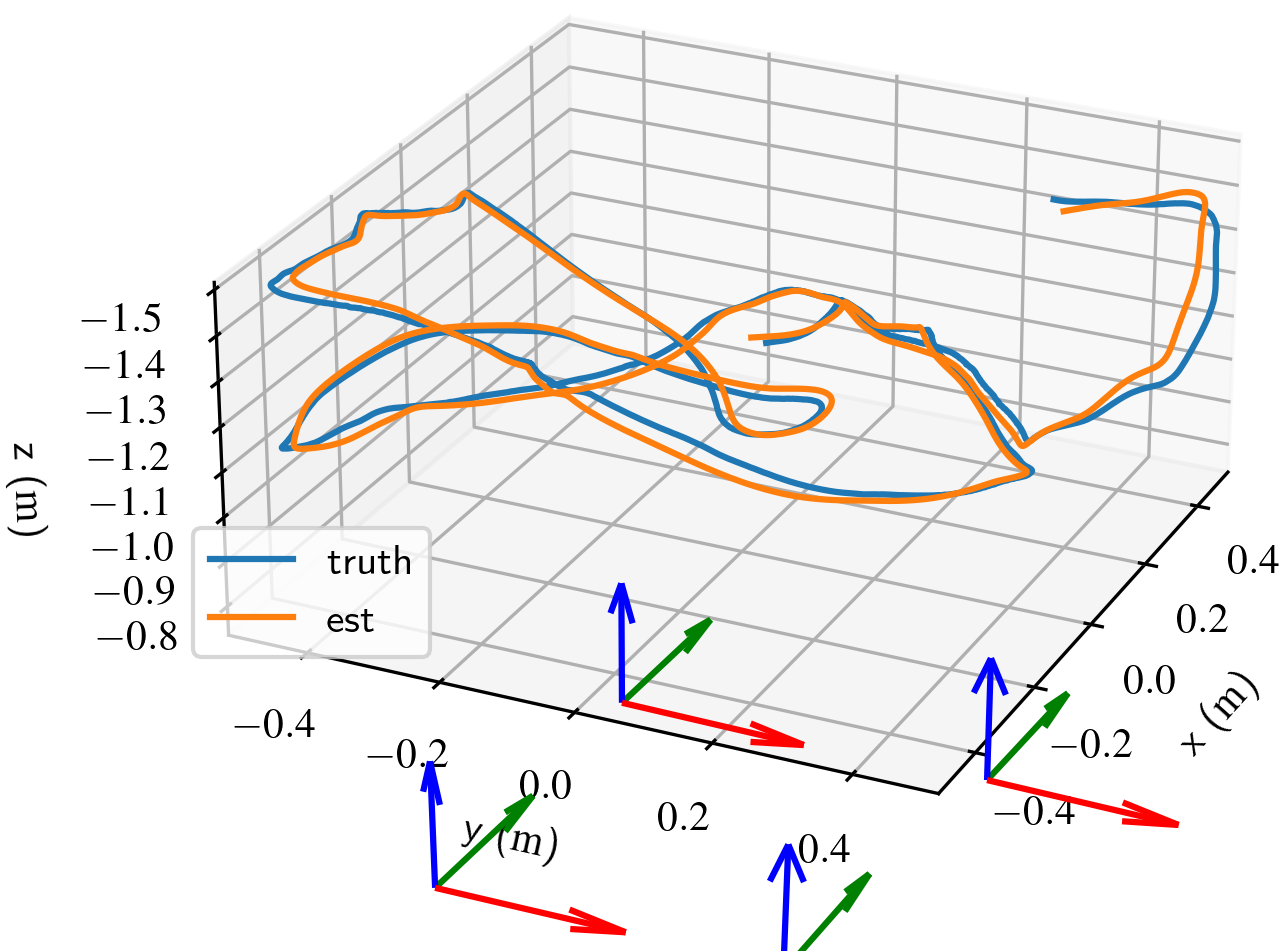}
         \label{fig:hw_spline_imump}
     }
    \caption{SE(3) spline ($k=4$) trajectory estimate for a hardware trajectory with each type of regularization.}
    \label{fig:hw_spline_traj}
\end{figure*}

\begin{figure*}
     \centering
     \subfigure[None]
     {
         \centering
         \includegraphics[width=0.23\textwidth]{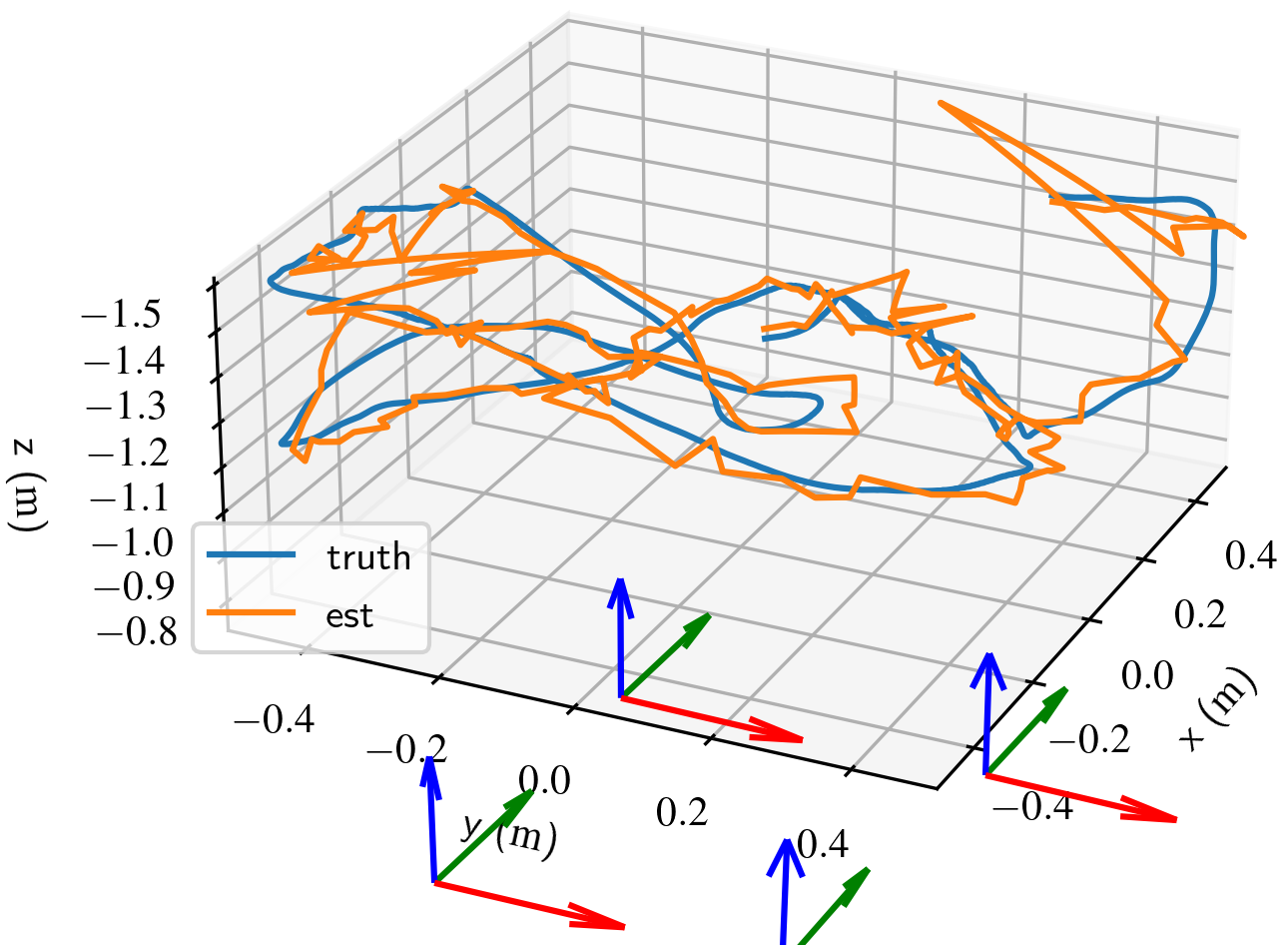}
         \label{fig:hw_gp_none}
     }
     \hfill
     \subfigure[Motion priors]
     {
         \centering
         \includegraphics[width=0.23\textwidth]{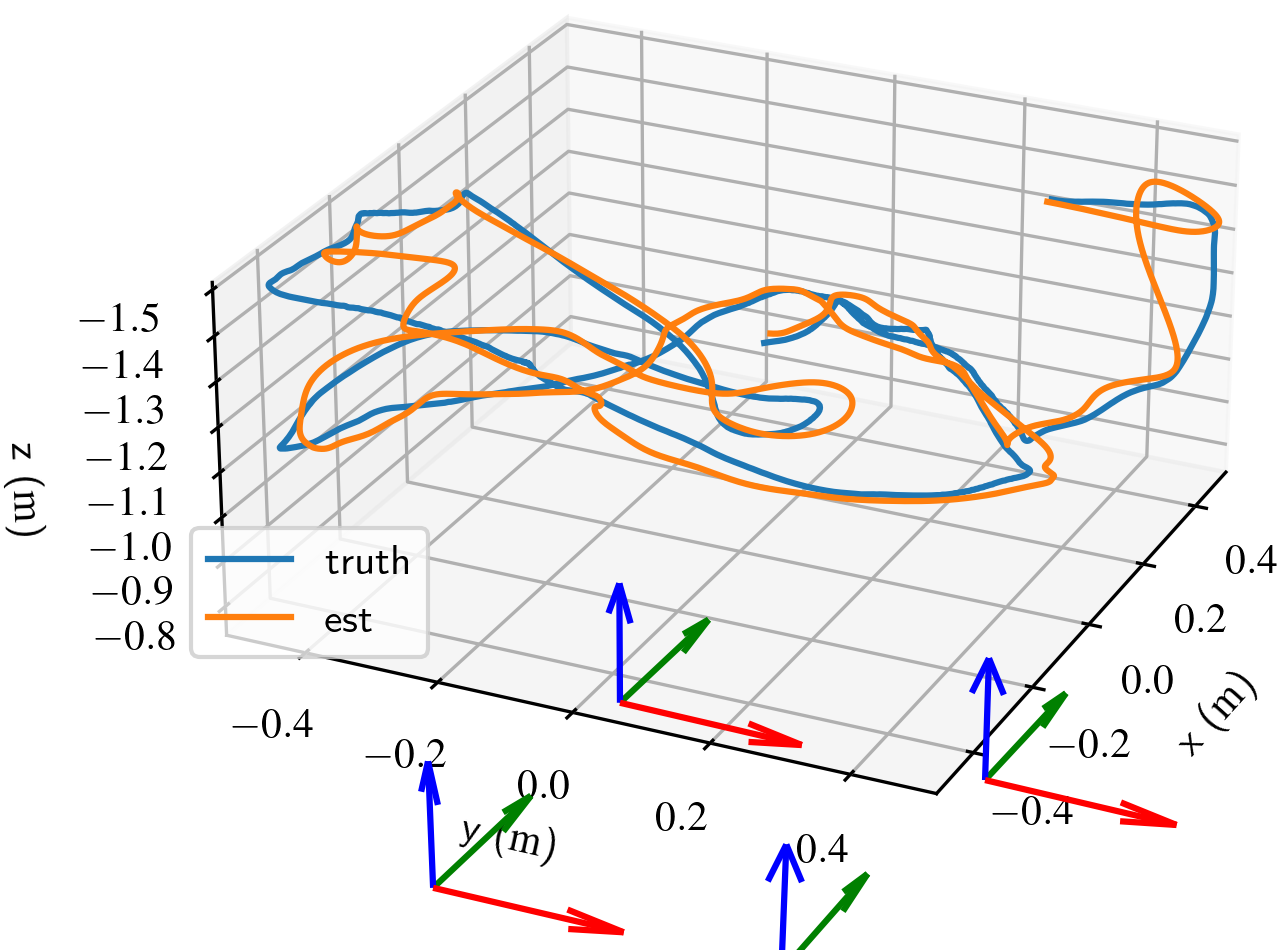}
         \label{fig:hw_gp_mp}
     }
     \hfill
     \subfigure[IMU]
     {
         \centering
         \includegraphics[width=0.23\textwidth]{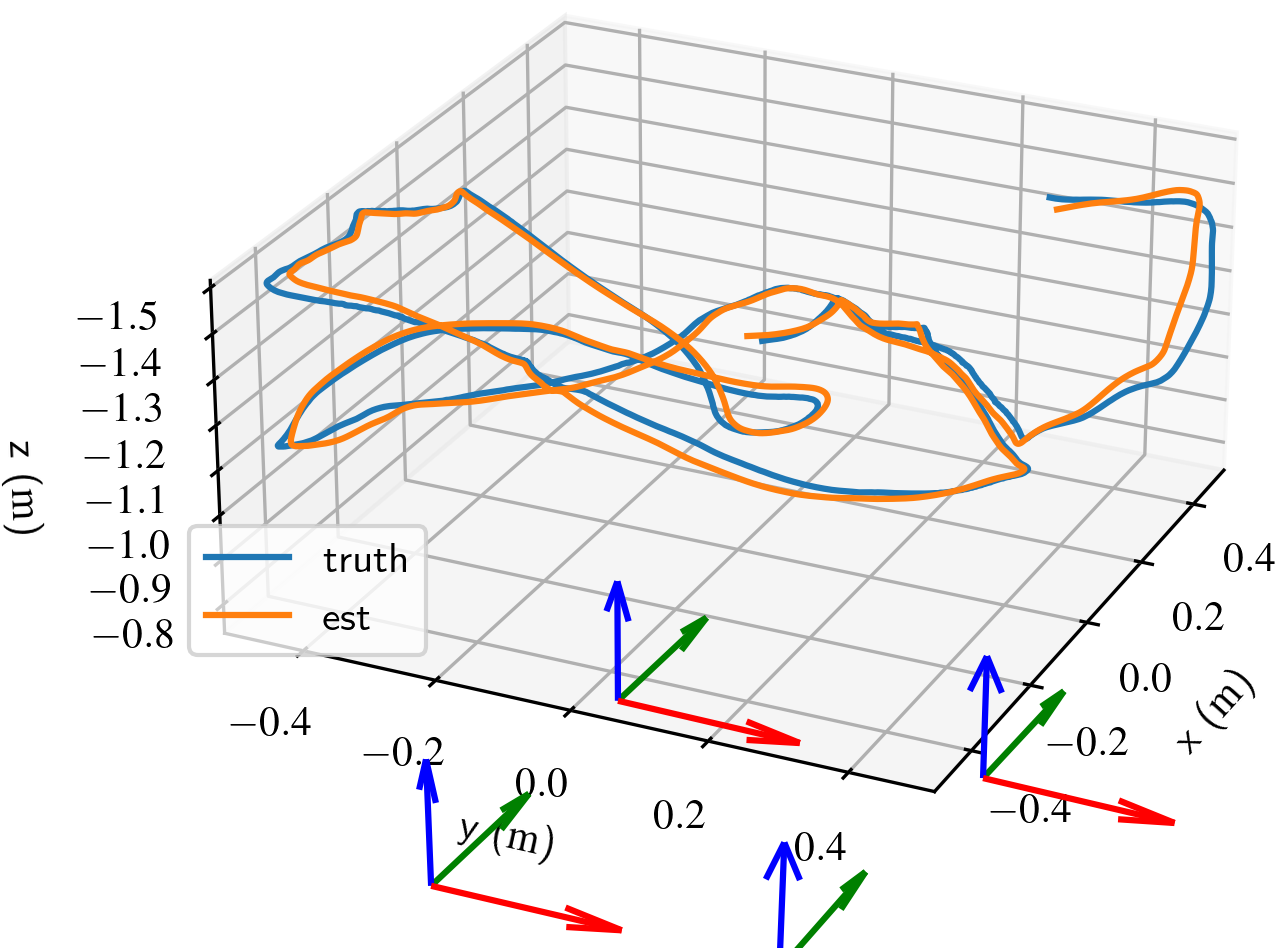}
         \label{fig:hw_gp_imu}
     }
     \hfill
     \subfigure[Motion priors + IMU]
     {
         \centering
         \includegraphics[width=0.23\textwidth]{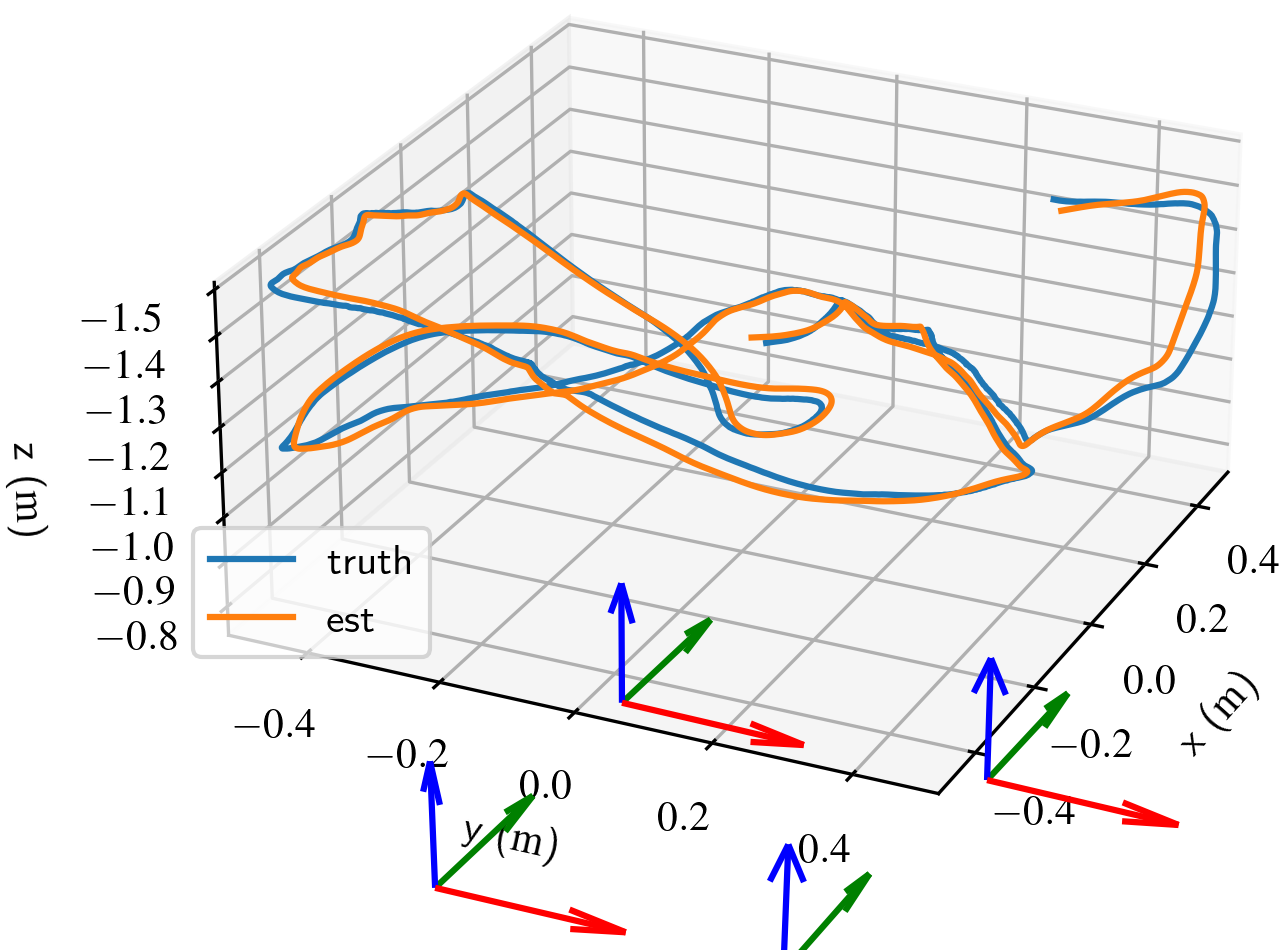}
         \label{fig:hw_gp_imump}
     }
    \caption{SE(3) GP trajectory estimate for a hardware trajectory with each type of regularization.}
    \label{fig:hw_gp_traj}
\end{figure*}

For our final comparison scenario, we again performed the experiment described in Section~\ref{sec:cam_imu_est}, but used actual data from a camera and IMU rather than simulated data. We placed four AprilTag~\cite{Wang2016} fiducial markers in a room equipped with an Optitrack\footnote{\texttt{https://optitrack.com/}} motion capture system that was used to obtain ground truth data. We maneuvered an Intel RealSense\footnote{\texttt{https://www.intelrealsense.com/}} d435i camera/IMU in the environment while observing and detecting the fiducial markers. Relative pose measurements were collected at 10~\si{Hz} and IMU measurements were collected at 100~\si{Hz}. We gathered data from several trajectories and performed estimation using each of the methods described in Section~\ref{sec:imu_sim} (again with $\delta t = 0.1$~\si{s} and $n_\text{GP} = 1$). After estimation, the relative transformation and time offset between the motion capture data and the estimated trajectory was optimized so that the ground truth and estimated trajectories could be directly compared. Figures~\ref{fig:hw_spline_traj} and~\ref{fig:hw_gp_traj} show resulting trajectories estimated using a spline on SE(3) (with $k=4$) and a GP on SE(3), respectively, using each type of regularization. Qualitatively, the results for the two methods are nearly identical when the same type of regularization is used. The motion priors had a profound impact on the smoothness of the trajectory, but the IMU enabled greater accuracy. Incorporating motion priors in addition to the IMU measurements appears to have had little to no effect.

\begin{figure*}
    \centering
    \includegraphics[width=0.98\textwidth]{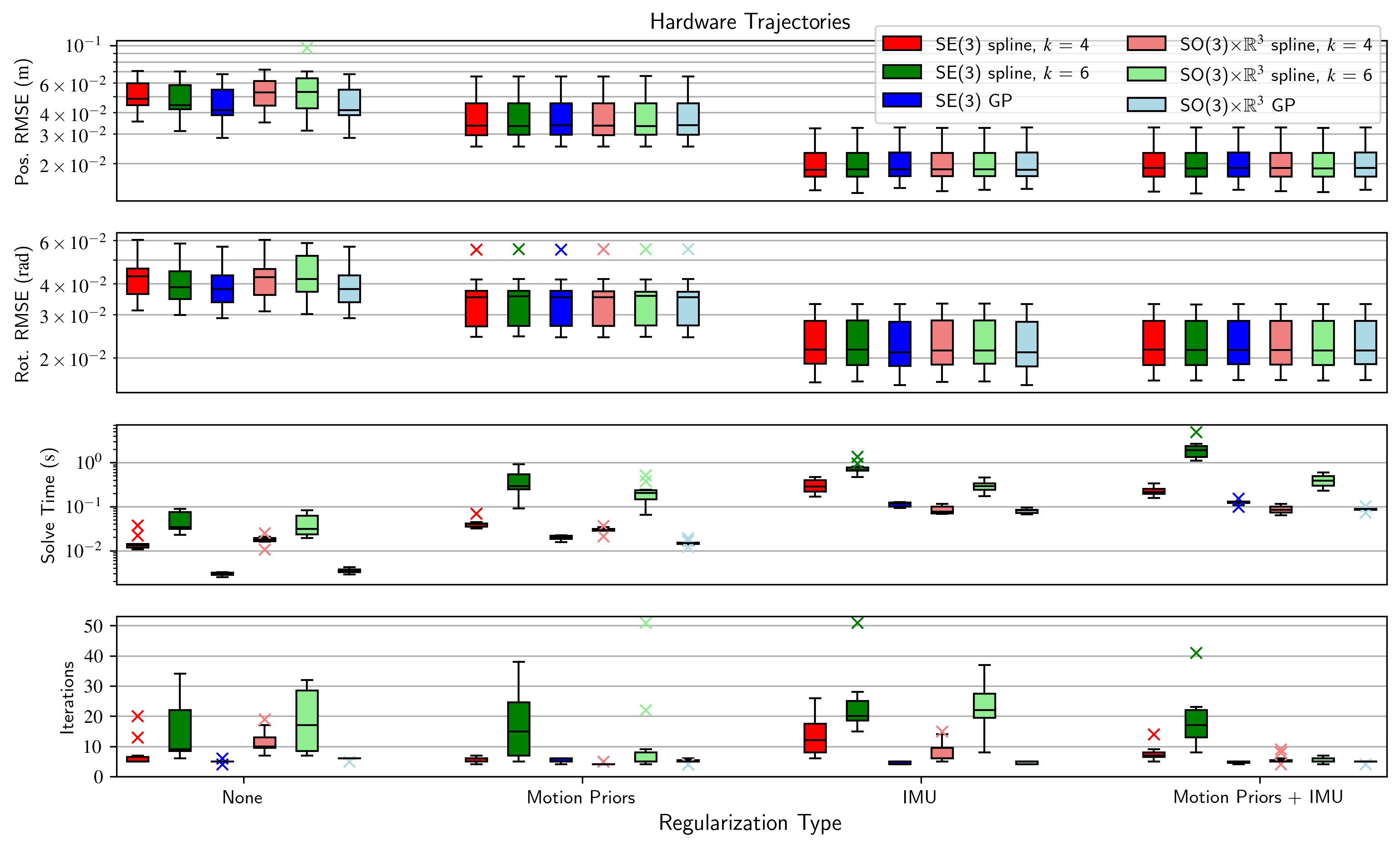}
    \caption{GP and spline trajectory error and solve time for the eleven hardware trajectories.}
    \label{fig:hw_mc}
\end{figure*}

Figure~\ref{fig:hw_mc} shows position and rotation RMSE\footnote{The motion capture system did not provide ground truth velocity data, so the twist RMSE is not shown.}, solve time, and number of Levenburg-Marquardt iterations required to converge for each method using each type of regularization for eleven distinct 10~\si{s} trajectories. These results follow a similar trend as Figures~\ref{fig:wnoj_mc_cam_imu} and~\ref{fig:sin_mc_cam_imu}. Accuracy improved significantly when motion priors were incorporated, and even more so when IMU measurements were used, but including motion priors in addition to the IMU measurements did not offer improvement. All methods had similar trajectory accuracy when using the same regularization type (although the GP methods had slightly better accuracy than the spline methods when no regularization was used). There were significant differences in computation time and number of optimizer iterations between the methods. The methods performed on $\text{SO}(3)\times \mathbb{R}^3$ had lower computation times than their SE(3) counterparts in all scenarios except when no regularization was used. GPs had faster solve times when IMU measurements were not used (again because no interpolation had to be done in these cases). However, when IMU measurements were used the solve times were mostly comparable between GPs and splines with $k=4$. The only exception to this is the case where motion priors were not used and estimation was done on SE(3), where splines with $k=4$ required a significantly higher number of iterations than GPs. Splines with $k=6$ had the highest solve times in all scenarios. They also required a  large number of iterations to converge to a solution. It is interesting to note that using motion priors often allowed the spline-based methods to converge with fewer iterations (although overall solve time did not improve because of the inclusion of additional residual terms to evaluate). We will elaborate more on these results in Section~\ref{sec:discussion}.

    \section{Discussion}
\label{sec:discussion}

% \JJcomment{
% \begin{itemize}
%     \item Purpose is to draw conclusions about when to use GPs versus splines \\
%     \item Increasing degree-of-differentiability is difficult using GPs, easy using splines \\
%     \item Increasing $k$ dramatically increases required compute time, because residual/jacobian evaluation goes up on order of $k^2$ and the linear system of equations get filled in more \\
%     \item Higher spline orders help with measurement fitting, but may not be worth it \\
%     \item Faster to solve on $\text{SO}(n)\times \mathbb{R}^n$ than SE($n$), consistent with literature \\
%     \item If you use the same model and the same measurements, GPs and splines give you the same accuracy. If you use GPs and splines with same degree-of-differentiability, computation times are similar \\
%     \item GPs are directly motivated by the chosen dynamic model, splines are not \\
%     \item Motion priors don't add additional help if IMU is in use, and can unnecessarily increase solve time (particularly for splines) \\
%     \item Point out fact that splines took longer when IMU was not in use. This is because GP doesn't have to interpolate, so estimation is basically discrete \\
%     \item Conclusion: if you need differentiability higher than 2 on lie groups, splines might be better because it's easy to increase differentiability. Otherwise, GPs and splines will probably give you similar results, provided that they have the same degree-of-differentiability.
% \end{itemize}
% }

\begin{figure*}
     \centering
     \subfigure[Spline, $k=4$]
     {
         \centering
         \includegraphics[width=0.2\textwidth]{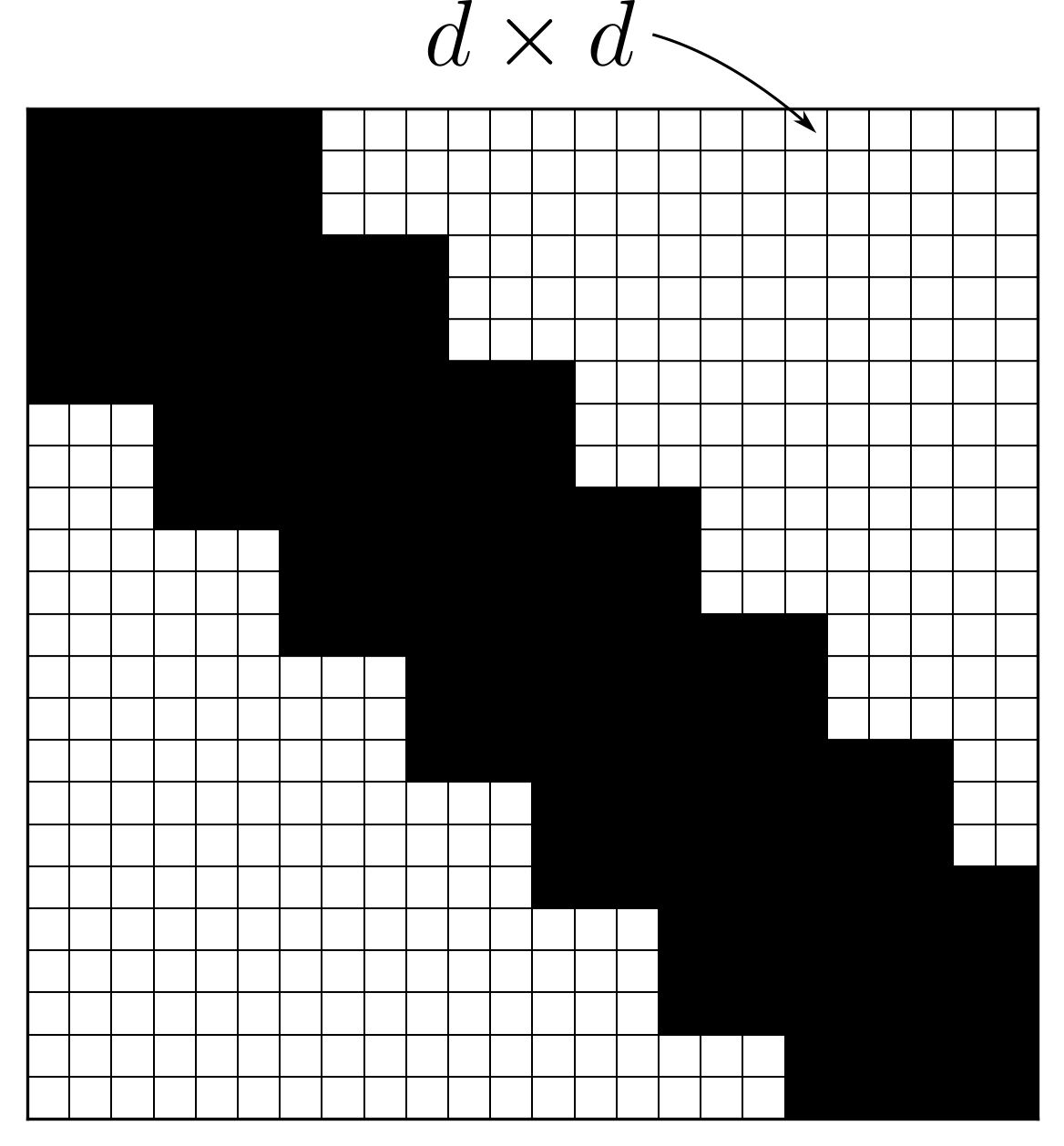}
         \label{fig:spline_4_info_mat}
     }
     \hfill
     \subfigure[Spline, $k=6$]
     {
         \centering
         \includegraphics[width=0.2\textwidth]{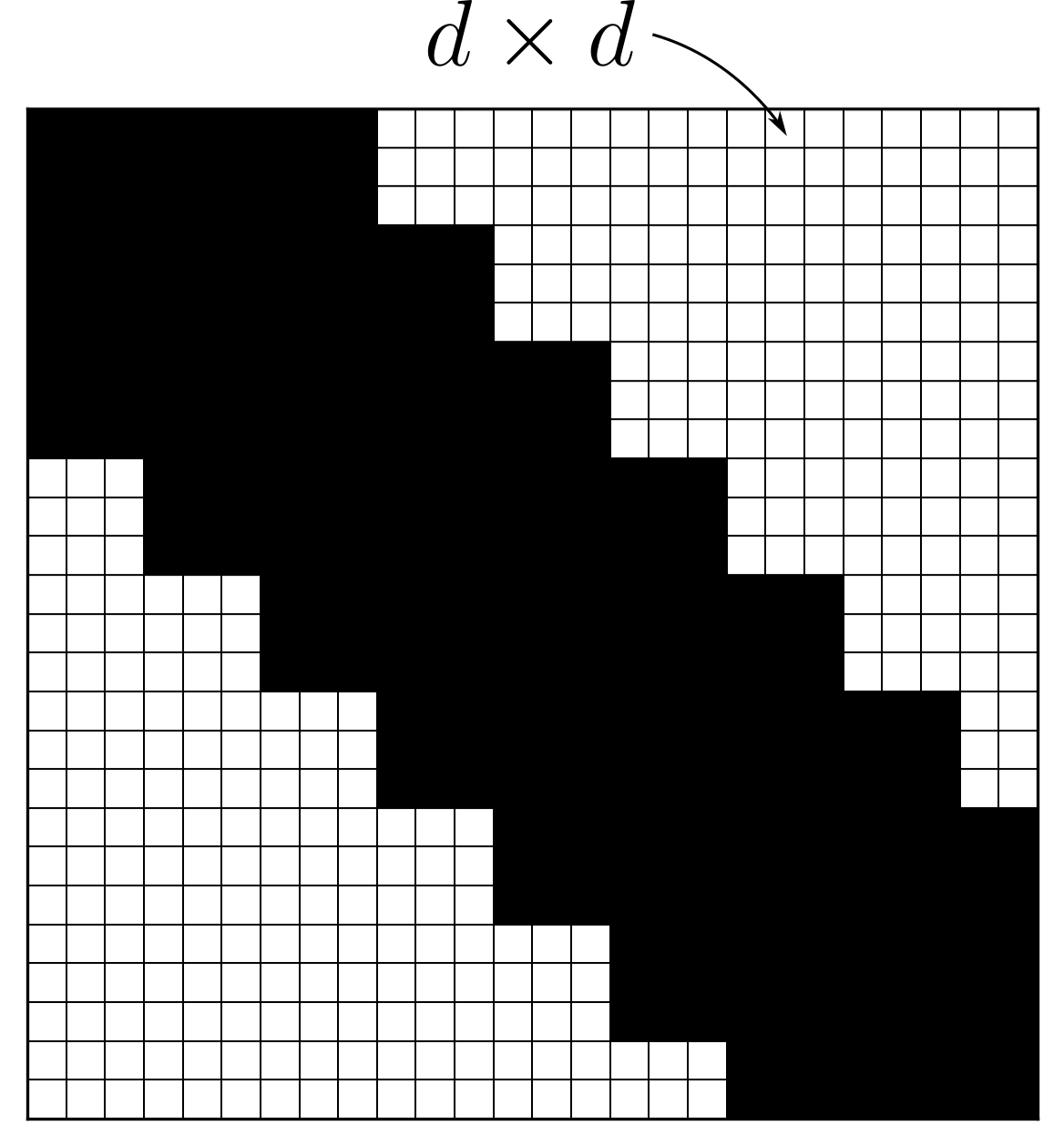}
         \label{fig:spline_6_info_mat}
     }
     \hfill
     \subfigure[GP]
     {
         \centering
         \includegraphics[width=0.2\textwidth]{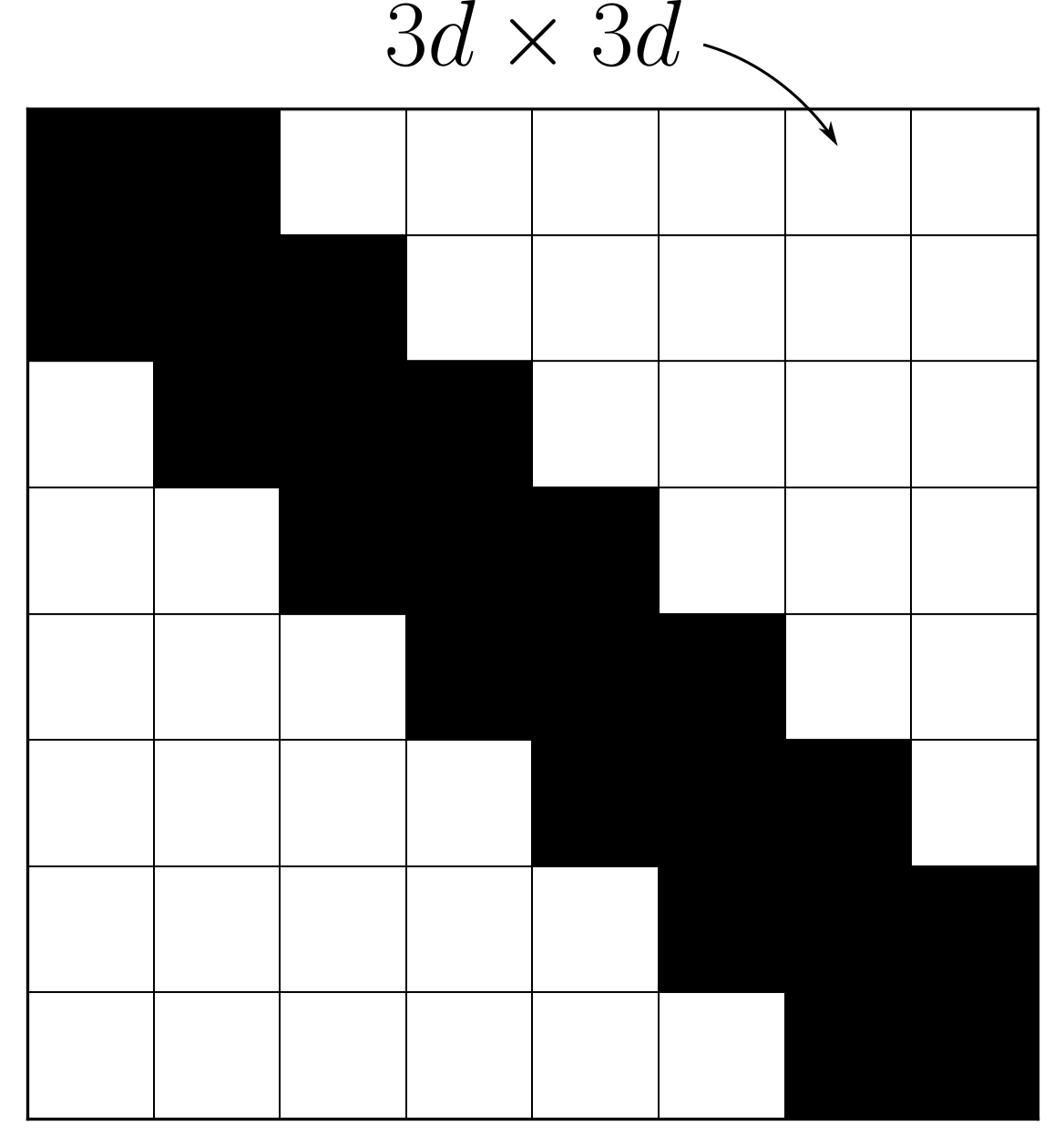}
         \label{fig:gp_info_mat}
     }
    \caption{Example information matrix structures for spline estimation with orders 4 and 6 and for GP estimation, using a WNOJ model.}
    \label{fig:info_mats}
\end{figure*}

The overall purpose of this paper is to understand when one of the two continuous-time estimation methods is preferable to the other. However, the results of the comparisons in Sections~\ref{sec:lin_sim},~\ref{sec:imu_sim}, and~\ref{sec:imu_hw} indicate that the two methods give similar results. When utilizing the same measurements and motion model, the resulting trajectory estimates are nearly identical. There are some significant differences in computation time, however. The most obvious is that splines with high order (e.g., $k=6$) take significantly longer to solve than GPs or splines with low order. The reason for this is straightforward. The state interpolation equation~\eqref{eq:spl_lie} for splines is dependent on $k$ control points, and Jacobian evaluation for each control point is $O(k)$ (see Appendix~\ref{sec:spl_derivs}). Thus the time complexity for evaluating each measurement residual is $O(k^2)$. In addition, the estimation problem becomes more dense with high spline orders. These factors are evident in Figure~\ref{fig:solve_time_breakdown}, where the Jacobian and residual evaluation time more than doubled and the linear solve time increased dramatically with $k=6$. Figure~\ref{fig:info_mats} shows the structure of the information matrix for splines with $k=4$ and $k=6$ (with $\delta t/ \delta t^\prime = 1/3$), and GPs when using a WNOJ model. A large amount of fill-in occurs as the spline order increases. Conversely, the information matrix for GPs is always block-tridiagonal (although the individual blocks are three times the size of the spline estimation blocks because they each contain pose, velocity, and acceleration information). 

\begin{figure}
    \centering
    \includegraphics[width=0.48\textwidth]{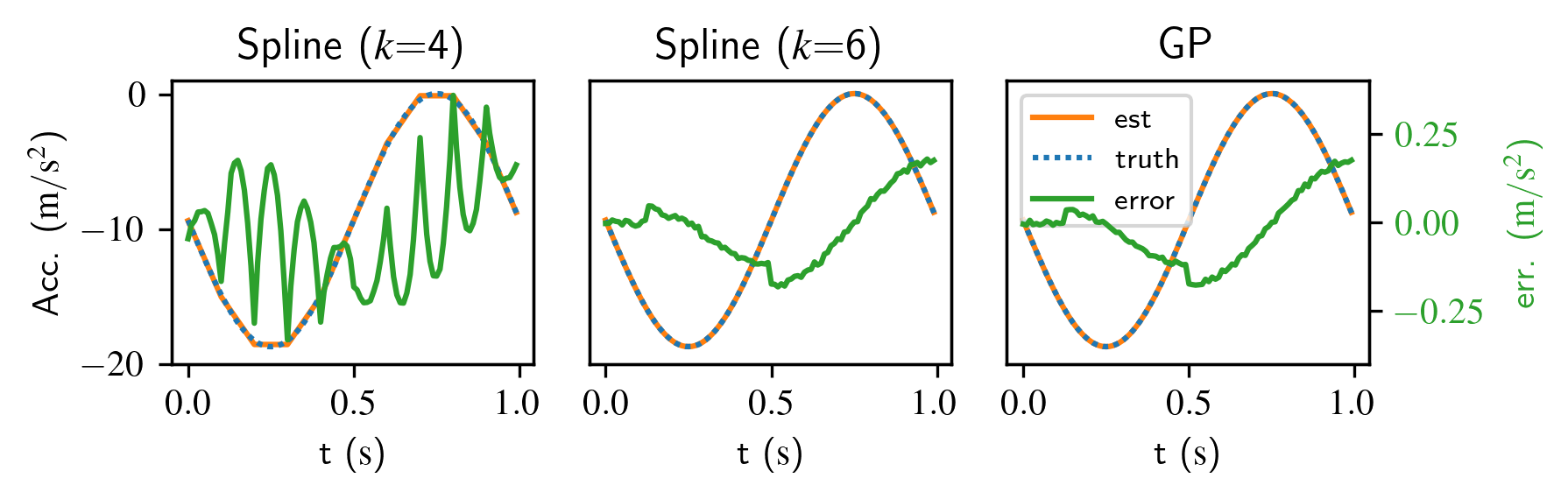}
    \caption{Post-fit accelerometer data for (from left to right) splines with $k=4$, splines with $k=6$, and GPs on SE(3). Error is shown in green using the scale on the right.}
    \label{fig:post_fit_acc}
\end{figure}

There is one benefit to choosing a higher spline order, however. The resulting spline has a higher degree-of-differentiability. Figure~\ref{fig:post_fit_acc} shows the true and estimated accelerometer measurements from the simulated sinusoidal trajectory of Section~\ref{sec:imu_sim} for (from left to right) splines on SE(3) with $k=4$, $k=6$, and GPs on SE(3). Because splines with $k=4$ are twice continuously differentiable, the estimator essentially fits straight lines to the measurement data, whereas for $k=6$ it fits cubic polynomials. This factor did not significantly degrade the quality of the trajectory estimate in this case, so the higher order may not have been worth the additional computational cost. Figure~\ref{fig:post_fit_acc} also shows a striking similarity between GP and spline ($k=6$) post-fit accelerometer residuals. This is because the acceleration component of the WNOJ GP interpolation equation~\eqref{eq:gp_interp} is cubic in $\tau$, so both splines with $k=6$ and GPs fit cubic polynomials to the accelerometer measurements. The main difference is that for GPs these polynomial segments are not continuously differentiable at the estimation times.

There are a few additional points we would also like to bring to attention.
\begin{itemize}
    \item In almost all cases, decoupling the translation and rotation trajectories gave similar accuracy with less computation time. This is consistent with previous findings from~\cite{Ovren2019, Haarbach2018}.
    \item GPs required more time to query the trajectory after optimization than splines with $k=4$ did (see Figure~\ref{fig:sampling_time}). This may not be a significant factor when only the state at the most recent measurement time is needed (as is often the case in online estimation), because the time to query the trajectory once is several orders of magnitude lower than the optimization solve time. However, the difference between GPs and splines is significant enough to bring to attention. Note that despite the longer query times, GP trajectory solve times are comparable to splines with $k = 4$ because they require fewer state interpolation Jacobians to evaluate per measurement time (2 versus 4).
    \item The trajectory representation for GPs is directly motivated by the chosen motion model. This is not the case for splines. Our experiments show that this fact does not impact the trajectory quality. However, it does make GP interpolation seem more intuitive than spline interpolation.
    \item In the case where the measurement times are all identical to the estimation times, GPs should be used instead of splines. This is evident from the solve times in Figure~\ref{fig:hw_mc} when IMU measurements were not used. In these cases the GP interpolation function becomes identity for each measurement time and the problem is equivalent to a discrete-time estimation problem (while still allowing for interpolation after the solve).
    \item The spline estimation parameters have a higher length of correlation in time than GPs. What we mean by this is that each control point will be correlated with the $k-1$ control points immediately preceding and following it in the information matrix, whereas each GP parameter will only be correlated with the one parameter that  precedes and follows it (see Figures~\ref{fig:gp_factor_graph},~\ref{fig:spline_factor_graph}, and~\ref{fig:info_mats}). This higher amount of time correlation will become problematic, for example, in SLAM scenarios where a feature is viewed at multiple time instances during the estimation window $\mathcal{W}$ and is later marginalized. In this case, the spline information matrix will fill in more than the GP information matrix will. Our tests do not include such a scenario, but we acknowledge that this is a potential case where GPs may outperform splines in terms of computation time. More investigation is needed to verify this.
    \item One potential benefit of using splines over GPs is that increasing the degree-of-differentiability is as simple as choosing a higher value of $k$. For GPs, this requires using a more complicated motion model.
\end{itemize}

In summary, GPs and splines give comparable results if the same measurements and motion model are used, and if the chosen spline degree-of-differentiability matches the differentiability required by the motion model (e.g., $k = 3$ for a WNOA model, $k = 4$ for a WNOJ model). However, in the case that a high degree-of-differentiability is required, splines may be more convenient, and in the case that the measurement times match the estimation times (i.e., the problem can be solved with discrete-time estimation), GPs should be used.
    \section{Conclusion}
\label{sec:conclusion}

In this paper we presented a direct comparison between GP and spline-based continuous-time trajectory estimation. We introduced continuous-time estimation generally and showed how the same motion models used in~\cite{Barfoot2014TE, Anderson2015STEAM} can be generalized for use with both GPs and splines on vector spaces and Lie groups. We then compared the two methods with a simulated linear model and in a camera/IMU sensor fusion scenario. Our results indicate that the two methods achieve similar trajectory accuracy if the same motion model and measurements are used. Solve times are also similar so long as the spline order is chosen such that the degree-of-differentiability matches that of the GP. Choosing a higher spline order did not improve trajectory accuracy, while requiring significantly higher solve times.

    \begin{appendix}
        \subsection{GP Interpolation as Optimal Control}
\label{sec:gp_optimal_control}

It is possible to arrive at the GP interpolation equation~\eqref{eq:gp_interp_prior} using an optimal control perspective. Consider the linear system in~\eqref{eq:lin_sys}, and suppose that the input $\mathbf{u}(t)$ is known for all values of $t$. The solution to the differential equation~\eqref{eq:lin_sys} is 
\begin{equation}
    \mathbf{x}(t) = \boldsymbol{\Phi}(t, t_n) \mathbf{x}_n + A_{t, t_n}[\mathbf{u}] + \int_{t_n}^t \boldsymbol{\Phi}(t, \tau) \mathbf{L} \mathbf{w}(\tau) d \tau
    \label{eq:lin_sys_sol}
\end{equation}
where $t \in [t_n, t_{n+1}]$, $\mathbf{x}_l = \mathbf{x}(t_l)$, and
\begin{equation}
    A_{\tau, \tau'}[\mathbf{u}] = \int_{\tau'}^\tau \boldsymbol{\Phi}(s, \tau') \mathbf{B} \mathbf{u}(s) ds.
\end{equation}
Our goal is to find the noise trajectory $\mathbf{w}(t)$ that satisfies
\begin{equation}
\begin{gathered}
    \min_{\mathbf{w}(t)} \quad \int_{t_n}^{t_{n+1}} \lVert \mathbf{w}(\tau) \rVert d \tau \\
\begin{aligned}
   \text{s.t.} \qquad \dot{\mathbf{x}} &= \mathbf{A}\mathbf{x} + \mathbf{B}\mathbf{u} + \mathbf{L}\mathbf{w} \\
   \mathbf{x}_n &= \bar{\mathbf{x}}_n \\
   \mathbf{x}_{n+1} &= \bar{\mathbf{x}}_{n+1}.
\end{aligned}
   \label{eq:min_norm_ctrl}
\end{gathered}
\end{equation}
Define the reachability Gramian
\begin{equation}
    W_R(t, t_n) = \int_{t_n}^t \boldsymbol{\Phi}(t, \tau) \mathbf{L} \mathbf{L}^\top \boldsymbol{\Phi}(t, \tau)^\top d \tau
\end{equation}
and note that $W_R(t, t_n) = \mathbf{Q}^{-1}\mathbf{Q}_t$, where $\mathbf{Q}$ is treated as a matrix that multiplies block-wise into $\mathbf{Q}_t$. If we assume that the system $(\mathbf{A}, \mathbf{L})$ is controllable, then the reachability Gramian is invertible. It is a well-known result that the solution to~\eqref{eq:min_norm_ctrl} is~\cite{Hespanha2018}
\begin{equation}
    \mathbf{w}^\ast(t) = \mathbf{L}^\top \boldsymbol{\Phi}(t_{n+1}, t)^\top \boldsymbol{\mu}_{n+1},
\end{equation}
where
\begin{equation}
\begin{gathered}
    \boldsymbol{\mu}_{n+1} = W_R(t_{n+1}, t_n)^{-1} \left( \mathbf{x}_{n+1} - \boldsymbol{\Phi}(t_{n+1}, t_n) \mathbf{x}_n \right. \\
    \left. \qquad \qquad \qquad \qquad - \: A_{t_{n+1}, t_n}[\mathbf{u}] \right).
\end{gathered}
\end{equation}
We can plug this noise input into~\eqref{eq:lin_sys_sol} to find the interpolated state:
\begin{equation}
\begin{split}
     \mathbf{x}(t) &= \boldsymbol{\Phi}(t, t_n) \mathbf{x}_n + A_{t, t_n}[\mathbf{u}] + \int_{t_n}^t \boldsymbol{\Phi}(t, \tau) \mathbf{L} \mathbf{w}^\ast(\tau) d \tau \\
     &= \boldsymbol{\Phi}(t, t_n) \mathbf{x}_n + A_{t, t_n}[\mathbf{u}] \\
     &\quad + \mathbf{Q} \left(\int_{t_n}^t \boldsymbol{\Phi}(t, \tau) \mathbf{L} \mathbf{L}^\top \boldsymbol{\Phi}(t_{n+1}, \tau)^\top d \tau \right) \cdot \\
     &\qquad \mathbf{Q}_{t_{n+1}}^{-1} \left( \mathbf{x}_{n+1} - \boldsymbol{\Phi}(t_{n+1}, t_n) \mathbf{x}_n  - A_{t_{n+1}, t_n}[\mathbf{u}] \right) \\
     &= \boldsymbol{\Phi}(t, t_n) \mathbf{x}_n + A_{t, t_n}[\mathbf{u}] \\
     &\quad + \mathbf{Q} \left(\int_{t_n}^t \boldsymbol{\Phi}(t, \tau) \mathbf{L} \mathbf{L}^\top \boldsymbol{\Phi}(t, \tau)^\top d \tau \right) \boldsymbol{\Phi}(t_{n+1}, t)^\top \cdot \\
     &\qquad \mathbf{Q}_{t_{n+1}}^{-1} \left( \mathbf{x}_{n+1} - \boldsymbol{\Phi}(t_{n+1}, t_n) \mathbf{x}_n  - A_{t_{n+1}, t_n}[\mathbf{u}] \right) \\
     &= \boldsymbol{\Phi}(t, t_n) \mathbf{x}_n + A_{t, t_n}[\mathbf{u}] \\
     &\quad + \mathbf{Q}_t \boldsymbol{\Phi}(t_{n+1}, t)^\top \mathbf{Q}_{t_{n+1}}^{-1} \cdot \\ 
     &\qquad \left( \mathbf{x}_{n+1} - \boldsymbol{\Phi}(t_{n+1}, t_n) \mathbf{x}_n  - A_{t_{n+1}, t_n}[\mathbf{u}] \right) \\
     &= \boldsymbol{\Lambda}(t) \mathbf{x}_n + \boldsymbol{\Omega}(t)(\mathbf{x}_{n+1} - A_{t_{n+1}, t_n}[\mathbf{u}]) + A_{t, t_n}[\mathbf{u}].
     \label{eq:min_norm_interp}
\end{split}
\end{equation}
Noting that the prior terms in~\eqref{eq:gp_interp_prior} are
\begin{equation}
    \check{\mathbf{x}}(\tau) = \boldsymbol{\Phi}(\tau, t_n) \check{\mathbf{x}}_n + A_{\tau, t_{n}}[\mathbf{u}],
\end{equation}
it can be seen that~\eqref{eq:min_norm_interp} and~\eqref{eq:gp_interp_prior} are equivalent when $\mathbf{x}_n = \bar{\mathbf{x}}_n, \ \mathbf{x}_{n+1} = \bar{\mathbf{x}}_{n+1}$, which must be the case due to the constraints in~\eqref{eq:min_norm_ctrl}.
        \subsection{Lie Group Gaussian Process Derivatives}
\label{sec:lie_gp_derivatives}

Here we derive the time derivatives of~\eqref{eq:gp_xi}.
\begin{equation}
\begin{split}
    \frac{d}{dt}(\boldsymbol{\xi}_n(t)) &= \lim_{h \to 0} \frac{\text{Log}(g(t+h) \bar{g}_n^{-1} ) - \boldsymbol{\xi}_n(t)}{h} \\
    &= \lim_{h \to 0} \frac{\text{Log}( \text{Exp}(\dot{g}(t)h) g(t) \bar{g}_n^{-1}) - \boldsymbol{\xi}_n(t) }{h} \\
    &= J_l^{-1}(\boldsymbol{\xi}_n(t)) \dot{g}(t).
\end{split}
\end{equation}
To compute the second derivative, we use the chain rule
\begin{equation}
    \frac{d}{dt} (\dot{\boldsymbol{\xi}}_n(t)) = J_l^{-1}(\boldsymbol{\xi}_n(t)) \ddot{g}(t) + \frac{d}{dt} \left( J_l^{-1}(\boldsymbol{\xi}_n(t)) \right) \dot{g}(t).
\end{equation}
We know of no simple way to compute $\frac{d}{dt} \left( J_l^{-1}(\boldsymbol{\xi}_n(t)) \right)$. Rather, we approximate it using a first-order Taylor series expansion,
\begin{equation}
    \frac{d}{dt} \left( J_l^{-1}(\boldsymbol{\xi}_n(t)) \right) \approx \frac{d}{dt} \left( \mathbf{I} - \frac{1}{2} \boldsymbol{\xi}_n(t)^\curlywedge \right) = -\frac{1}{2} \dot{\boldsymbol{\xi}}_n(t)^\curlywedge.
\end{equation}
This approximation is reasonable so long as $\dot{\boldsymbol{\xi}}_n(t)$ remains small throughout the interval $[t_n, t_{n+1}]$. Applying this approximation, we arrive at~\eqref{eq:gp_local_var}.

\subsection{Lie Group Gaussian Process Jacobians}
\label{sec:lie_gp_jacs}

\subsubsection{Global Variable Jacobians}

We need the Jacobians of the interpolated global variables $\mathbf{x}(t)$ in~\eqref{eq:gp_global_var} with respect to the estimation states $\bar{\mathbf{x}}_n, \bar{\mathbf{x}}_{n+1}$, where $t_n \leq t \leq t_{n+1}$. We will start with $g(t)$. Note that
\begin{equation}
\begin{gathered}
    \frac{\partial g(t)}{\partial \bar{g}_n} = \frac{\partial g(t)}{\partial \bar{g}_n} + \frac{\partial g(t)}{\partial \boldsymbol{\xi}_n(t)} \frac{\partial \boldsymbol{\xi}_n(t)}{\partial \bar{g}_n}, \\
    \frac{\partial g(t)}{\partial \bar{g}_{n+1}} = \frac{\partial g(t)}{\partial \boldsymbol{\xi}_n(t)} \frac{\partial \boldsymbol{\xi}_n(t)}{\partial \bar{g}_{n+1}}, \\
    \frac{\partial g(t)}{\partial \dot{\bar{g}}_n} = \frac{\partial g(t)}{\partial \boldsymbol{\xi}_n(t)} \frac{\partial \boldsymbol{\xi}_n(t)}{\partial \dot{\bar{g}}_n}, \\
    \frac{\partial g(t)}{\partial \dot{\bar{g}}_{n+1}} = \frac{\partial g(t)}{\partial \boldsymbol{\xi}_n(t)} \frac{\partial \boldsymbol{\xi}_n(t)}{\partial \dot{\bar{g}}_{n+1}}, \\
    \frac{\partial g(t)}{\partial \ddot{\bar{g}}_n} = \frac{\partial g(t)}{\partial \boldsymbol{\xi}_n(t)} \frac{\partial \boldsymbol{\xi}_n(t)}{\partial \ddot{\bar{g}}_n}, \\
    \frac{\partial g(t)}{\partial \ddot{\bar{g}}_{n+1}} = \frac{\partial g(t)}{\partial \boldsymbol{\xi}_n(t)} \frac{\partial \boldsymbol{\xi}_i(t)}{\partial \ddot{\bar{g}}_{n+1}}.
\end{gathered}
\end{equation}
These are computed as
\begin{equation}
\begin{split}
    \frac{\partial g}{\partial \bar{g}_n} &= \lim_{\boldsymbol{\tau} \to \mathbf{0}} \frac{\text{Log}\left( \text{Exp}(\boldsymbol{\xi}_n) \text{Exp}(\boldsymbol{\tau}) \bar{g}_n \left( \text{Exp}(\boldsymbol{\xi}_n) \bar{g}_n \right)^{-1} \right)}{\boldsymbol{\tau}} \\
    &= \text{Ad}_{\text{Exp}(\boldsymbol{\xi}_n)}, \\
    \frac{\partial g}{\partial \boldsymbol{\xi}_n} &= \lim_{\boldsymbol{\tau} \to \mathbf{0}} \frac{\text{Log}\left( \text{Exp}(\boldsymbol{\xi}_n + \boldsymbol{\tau}) \bar{g}_n \left( \text{Exp}(\boldsymbol{\xi}_n) \bar{g}_n \right)^{-1} \right)}{\mathbf{\tau}} \\
    &= \lim_{\boldsymbol{\tau \to \mathbf{0}}} \frac{\text{Log}\left( \text{Exp}(J_l(\boldsymbol{\xi}_n)\boldsymbol{\tau}) \text{Exp}(\boldsymbol{\xi}_n) \text{Exp}(\boldsymbol{\xi}_n)^{-1} \right)}{\boldsymbol{\tau}} \\
    &= J_l(\boldsymbol{\xi}_n).
\end{split}
\end{equation}
where $\text{Ad}_X$ is the adjoint representation of $X \in G$. Computing the Jacobians of the local variables with respect to the estimation parameters is more involved. We will do this in Section~\ref{sec:gp_loc_var_jac}.

Next, we need the Jacobians of $\dot{g}(t)$,
\begin{equation}
    \frac{\partial \dot{g}(t)}{\partial \bar{g}_n} = \frac{\partial \dot{g}(t)}{\partial \boldsymbol{\xi}_n(t)} \frac{\partial \boldsymbol{\xi}_n(t)}{\partial \bar{g}_n} + \frac{\partial \dot{g}(t)}{\partial \dot{\boldsymbol{\xi}}_n(t)} \frac{\partial \dot{\boldsymbol{\xi}}_n(t)}{\partial \bar{g}_n}.
\end{equation}
The Jacobians of $\dot{g}(t)$ with respect to all other estimation parameters have a similar form. Note that for arbitrary $\mathbf{r},\mathbf{v} \in \mathbb{R}^d$,
\begin{equation}
\begin{split}
    &\frac{\partial}{\partial \mathbf{r}} \left( J_l(\mathbf{r}) \mathbf{v} \right) \\
    &\quad= \lim_{\boldsymbol{\tau} \to \mathbf{0}} \frac{J_l(\mathbf{r} + \boldsymbol{\tau}) \mathbf{v} - J_l(\mathbf{r}) \mathbf{v}}{\boldsymbol{\tau}} \\ 
    &\quad\approx \lim_{\boldsymbol{\tau} \to \mathbf{0}} \frac{\left( \left( \mathbf{I} + \frac{1}{2} (\mathbf{r} + \boldsymbol{\tau})^\curlywedge  + O\left({\mathbf{r}^\curlywedge}^2\right)\right) - J_l(\mathbf{r}) \right) \mathbf{v}}{\boldsymbol{\tau}} \\
    &\quad= \lim_{\boldsymbol{\tau} \to \mathbf{0}} \frac{\left(\frac{1}{2}\boldsymbol{\tau}^\curlywedge + J_l(\mathbf{r})\right) \mathbf{v} - J_l(\mathbf{r}) \mathbf{v}}{\boldsymbol{\tau}} \\
    &\quad= -\frac{1}{2}\mathbf{v}^\curlywedge.
\end{split}
\label{eq:Jl_jac}
\end{equation}
Here we have used the Taylor series expansion of $J_l(\cdot)$ and the fact that the wedge map $\curlywedge$ is linear. Also note that high-order terms of $\boldsymbol{\tau}$ in the numerator go to zero in the limit. However, we have ignored cross-terms between $\mathbf{r}$ and $\boldsymbol{\tau}$ that are linear in $\boldsymbol{\tau}$, thus this is an approximation of the true Jacobian. This approximation should hold well so long as $\mathbf{r}$ and/or $\mathbf{v}$ remain small, which will almost always be the case in our segmented local GP formulation. Thus we have
\begin{equation}
\begin{gathered}
    \frac{\partial\dot{g}(t)}{\partial \boldsymbol{\xi}_n(t)} = -\frac{1}{2} \dot{\boldsymbol{\xi}}_n(t)^\curlywedge, \\
    \frac{\partial\dot{g}(t)}{\partial \dot{\boldsymbol{\xi}}_n(t)} = J_l(\boldsymbol{\xi}_n(t)).
\end{gathered}
\end{equation}

Next, for $\ddot{g}(t)$,
\begin{equation}
\begin{split}
    \frac{\partial \ddot{g}(t)}{\partial \bar{g}_n} &= \frac{\partial \ddot{g}(t)}{\partial \boldsymbol{\xi}_n(t)} \frac{{\partial \boldsymbol{\xi}_n(t)}}{\partial \bar{g}_n} + \frac{\partial \ddot{g}(t)}{\partial \dot{\boldsymbol{\xi}}_n(t)} \frac{{\partial \dot{\boldsymbol{\xi}}_n(t)}}{\partial \bar{g}_n} \\
    &\quad+ \frac{\partial \ddot{g}(t)}{\partial \ddot{\boldsymbol{\xi}}_n(t)} \frac{{\partial \ddot{\boldsymbol{\xi}}_n(t)}}{\partial \bar{g}_n} + \frac{\partial \ddot{g}(t)}{\partial \dot{g}(t)} \frac{\partial \dot{g}(t)}{\partial \bar{g}_n}.
\end{split}
\end{equation}
Similarly for all other estimation parameters. We have
\begin{equation}
\begin{gathered}
    \frac{\partial \ddot{g}(t)}{\partial \boldsymbol{\xi}_n(t)} = -\frac{1}{2} \left( \ddot{\boldsymbol{\xi}}_n(t) + \frac{1}{2} \dot{\boldsymbol{\xi}}_n(t)^\curlywedge \dot{g}(t) \right)^\curlywedge \\
    \frac{\partial \ddot{g}(t)}{\partial \dot{\boldsymbol{\xi}}_n(t)} = -\frac{1}{2} J_l(\boldsymbol{\xi}_n(t)) \dot{g}(t)^\curlywedge \\
    \frac{\partial \ddot{g}(t)}{\partial \ddot{\boldsymbol{\xi}}_n(t)} = J_l(\boldsymbol{\xi}_n(t)) \\
    \frac{\partial \ddot{g}(t)}{\partial \dot{g}(t)} = \frac{1}{2} J_l(\boldsymbol{\xi}_n(t)) \dot{\boldsymbol{\xi}}_n(t)^\curlywedge.
\end{gathered}
\end{equation}
Note that we used~\eqref{eq:Jl_jac} in the first line.

\subsubsection{Local Variable Jacobians}
\label{sec:gp_loc_var_jac}

Note that $\boldsymbol{\gamma}_n(t_n) = \begin{bmatrix} \mathbf{0}^\top & \dot{\bar{g}}_n^\top & \ddot{\bar{g}}_n^\top \end{bmatrix}^\top$. We can expand each row of~\eqref{eq:gp_interp_lie} as
\begin{equation}
\begin{split}
    &\boldsymbol{\xi}_n^{(i)}(t) = \Lambda_{i,1} \dot{\bar{g}}_n + \Lambda_{i,2} \ddot{\bar{g}}_n + \Omega_{i,0} \boldsymbol{\xi}_n(t_{n+1}) + \Omega_{i,1} \dot{\boldsymbol{\xi}}_n(t_{n+1}) \\
    &\quad+ \Omega_{i,2} \left( -\frac{1}{2} \dot{\boldsymbol{\xi}}_n(t_{n+1}))^\curlywedge \dot{\bar{g}}_{n+1} + J_l^{-1}(\boldsymbol{\xi}_n(t_{n+1})) \ddot{\bar{g}}_{n+1}\right),
\end{split}
\label{eq:xi_interp}
\end{equation}
where $(\cdot)^{(i)}, i \in \{0, 1, 2\}$ indicates the $i$-th derivative and $(\cdot)_{i,j} \in \mathbb{R}^{d\times d}$ indicates the $(i,j)$-th block. Since~\eqref{eq:xi_interp} has the same form for each derivative, we only need to compute its Jacobians for a single case. First, we will need the following:
\begin{equation}
\begin{split}
    \frac{\partial \boldsymbol{\xi}_n(t_{n+1})}{\partial \bar{g}_n} &= \lim_{\boldsymbol{\tau} \to \mathbf{0}} \frac{\text{Log} \left( \bar{g}_{n+1} \left(\text{Exp}(\boldsymbol{\tau}) \bar{g}_n \right)^{-1} \right) - \boldsymbol{\xi}_n(t_{n+1})}{\boldsymbol{\tau}} \\
    &= - J_r^{-1} (\boldsymbol{\xi}_n(t_{n+1})),
\end{split}
\end{equation}
\begin{equation}
\begin{split}
    \frac{\partial \boldsymbol{\xi}_n(t_{n+1})}{\partial \bar{g}_{n+1}} &= \lim_{\boldsymbol{\tau} \to \mathbf{0}} \frac{\text{Log} \left( \text{Exp}(\boldsymbol{\tau}) \bar{g}_{n+1} \bar{g}_n^{-1} \right) - \boldsymbol{\xi}_n(t_{n+1})}{\boldsymbol{\tau}} \\
    &= J_l^{-1}(\boldsymbol{\xi}_n(t_{n+1}),
\end{split}
\end{equation}
where $J_r(\cdot)$ is the right Jacobian of $G$.
Next, we will need the following. For arbitrary $\mathbf{r}, \mathbf{v} \in \mathbb{R}^d$ and $\mathbf{A} \in \mathbb{R}^{d \times d}$,
\begin{equation}
    \frac{\partial}{\partial \mathbf{r}} \left(J_l^{-1}(\mathbf{r}) \mathbf{v} \right) = \frac{1}{2}\mathbf{v}^\curlywedge,
\end{equation}
following similar logic (and using a similar approximation) as~\eqref{eq:Jl_jac}, and
\begin{equation}
\begin{split}
    &\frac{\partial}{\partial \mathbf{v}} \left( (\mathbf{A}\mathbf{v})^\curlywedge \mathbf{v}\right) \\ 
    &\quad= \lim_{\boldsymbol{\tau} \to \mathbf{0}} \frac{\left(\mathbf{A} (\mathbf{v} + \boldsymbol{\tau}) \right)^\curlywedge (\mathbf{v} + \boldsymbol{\tau}) - (\mathbf{A}\mathbf{v})^\curlywedge \mathbf{v}}{\boldsymbol{\tau}} \\
    &\quad= \lim_{\boldsymbol{\tau} \to \mathbf{0}} \frac{ \left( (\mathbf{A}\mathbf{v})^\curlywedge + (\mathbf{A}\boldsymbol{\tau})^\curlywedge\right)(\mathbf{v} + \boldsymbol{\tau}) - (\mathbf{A}\mathbf{v})^\curlywedge \mathbf{v}}{\boldsymbol{\tau}} \\
    &\quad= \lim_{\boldsymbol{\tau} \to \mathbf{0}} \frac{  (\mathbf{A}\mathbf{v})^\curlywedge \boldsymbol{\tau} + (\mathbf{A}\boldsymbol{\tau})^\curlywedge\mathbf{v}}{\boldsymbol{\tau}} \\
    &\quad= (\mathbf{Av})^\curlywedge - \mathbf{v}^\curlywedge \mathbf{A}.
\end{split}
\label{eq:mat_vec_twist_vec_jac}%
\end{equation}
We then have
\begin{gather}
    \frac{\partial \dot{\boldsymbol{\xi}}_n(t_{n+1})}{\partial \bar{g}_n} = \frac{1}{2} \dot{\bar{g}}_{n+1}^\curlywedge \frac{\partial \boldsymbol{\xi}_n(t_{n+1})}{\partial \bar{g}_n}, \\
    \frac{\partial \dot{\boldsymbol{\xi}}_n(t_{n+1})}{\partial \bar{g}_{n+1}} = \frac{1}{2} \dot{\bar{g}}_{n+1}^\curlywedge \frac{\partial \boldsymbol{\xi}_n(t_{n+1})}{\partial \bar{g}_{n+1}}, \\
    \frac{\partial \dot{\boldsymbol{\xi}}_n(t_{n+1})}{\partial \dot{\bar{g}}_{n+1}} = J_l^{-1}(\boldsymbol{\xi}_n(t_{n+1})).
\end{gather}

We are now ready to compute the Jacobians of~\eqref{eq:xi_interp}.
\begin{equation}
\begin{split}
    &\frac{\partial \boldsymbol{\xi}_n^{(i)}(t)}{\partial \bar{g}_n} = \Omega_{n,0} \frac{\partial \boldsymbol{\xi}_n(t_{n+1})}{\partial \bar{g}_n} + \Omega_{n,1} \frac{\partial \dot{\boldsymbol{\xi}}_n(t_{n+1})}{\partial \bar{g}_n} \\
    &\quad+ \frac{1}{2} \Omega_{n,2} \left( \dot{\bar{g}}_{n+1}^\curlywedge \frac{\partial \dot{\boldsymbol{\xi}}_n(t_{n+1})}{\partial \bar{g}_n} + \ddot{\bar{g}}_{n+1}^\curlywedge \frac{\partial \boldsymbol{\xi}_n(t_{n+1})}{\partial \bar{g}_n} \right) \\
    &= \left( \Omega_{n,0} + \frac{1}{2} \Omega_{n,1} \dot{\bar{g}}_{n+1}^\curlywedge \right. \\
    &\quad+ \left.\frac{1}{2} \Omega_{n,2} \left( \frac{1}{2} {\dot{\bar{g}}_{n+1}^\curlywedge}^2 + \ddot{\bar{g}}_{n+1}^\curlywedge \right) \right) \frac{\partial \boldsymbol{\xi}_n(t_{n+1})}{\partial \bar{g}_n}.
\end{split}
\end{equation}
Similarly for $\frac{\partial \boldsymbol{\xi}_n^{(i)}(t)}{\partial \bar{g}_{n+1}}$, and
\begin{equation}
\begin{gathered}
    \frac{\partial \boldsymbol{\xi}_n^{(i)}(t)}{\partial \dot{\bar{g}}_n} = \Lambda_{i,1}, \quad \frac{\partial \boldsymbol{\xi}_n^{(i)}(t)}{\partial \ddot{\bar{g}}_n} = \Lambda_{i,2}, \\
    \begin{aligned}
    \frac{\partial \boldsymbol{\xi}_n^{(i)}(t)}{\partial \dot{\bar{g}}_{n+1}} &= \Omega_{i,1} J_l^{-1}(\boldsymbol{\xi}_n(t_{n+1})) \\
    &\quad- \frac{1}{2} \Omega_{i,2} \left( \dot{\boldsymbol{\xi}}_n(t_{n+1})^\curlywedge - \dot{\bar{g}}_{n+1}^\curlywedge J_l^{-1}(\boldsymbol{\xi}_n(t_{n+1})) \right) ,
    \end{aligned}\\
    \frac{\partial \boldsymbol{\xi}_n^{(i)}(t)}{\partial \ddot{\bar{g}}_{n+1}} = \Omega_{n,3}(t) J_l^{-1}( \boldsymbol{\xi}_n(t_{n+1})).
\end{gathered}
\end{equation}
Here we have used~\eqref{eq:mat_vec_twist_vec_jac} to compute $\frac{\partial \boldsymbol{\xi}_n^{(i)}(t)}{\partial \dot{\bar{g}}_{n+1}}$.

        \subsection{Lie Group Spline Derivatives}
\label{sec:spl_derivs}

Here we compute the first two time derivatives of~\eqref{eq:spl_lie}. We will first need to express~\eqref{eq:spl_lie} in a recursive form that is cleaner to work with (similar to what was done in~\cite{Sommer2020}),
\begin{equation}
\begin{split}
    g^{(0)} &= g_{n-k+1}, \\
    g^{(j)} &= g^{(j-1)}\mathbf{A}_j(u),
    \label{eq:spl_recurs}
\end{split}
\end{equation}
where $\mathbf{A}_j(u) = \text{Exp}(b_j(u) \boldsymbol{\Omega}_{n+j-k+1})$. Then $g(t) = g^{(k-1)}$.

\subsubsection{First Derivative}

First we need $\frac{d}{dt} \mathbf{A}_j(u)$.
\begin{equation}
\begin{split}
    \frac{d}{dt} \mathbf{A}_j(u) &= \dot{b}_j(u) \frac{d}{db_j} \text{Exp}(b_j \boldsymbol{\Omega}_{n+j-k+1}) \\
    &= \dot{b}_j(u) \boldsymbol{\Omega}_{n+j-k+1},
\end{split}
\end{equation}
where the second equality holds because $b_j \in \mathbb{R}$ induces a one parameter subgroup of $G$.
Next, we need the product rule on Lie groups. Given $X(t), Y(t) \in G$, we use the chain rule to find
\begin{equation}
    \frac{d}{dt} X(t) Y(t) = \frac{\partial}{\partial X} (XY) \frac{dX}{dt} + \frac{\partial}{\partial Y} (XY) \frac{dY}{dt}.
\end{equation}
We can calculate $\frac{\partial}{\partial X} (XY)$ and $\frac{\partial}{\partial Y} (XY)$:

\begin{align}
&\begin{split}
    \frac{\partial}{\partial X} (XY) &= \lim_{\boldsymbol{\tau} \to 0} \frac{\text{Log}(\text{Exp}(\boldsymbol{\tau})XY (XY)^{-1})}{\boldsymbol{\tau}} \\
    &= \mathbf{I},
\end{split} \\
&\begin{split}
    \frac{\partial}{\partial Y} (XY) &= \lim_{\boldsymbol{\tau} \to 0} \frac{\text{Log}(X\text{Exp}(\boldsymbol{\tau})Y (XY)^{-1})}{\boldsymbol{\tau}} \\
    &= \lim_{\boldsymbol{\tau} \to 0} \frac{\text{Log}(X\text{Exp}(\boldsymbol{\tau}) X^{-1})}{\boldsymbol{\tau}} \\ 
    &= \lim_{\boldsymbol{\tau} \to 0} \frac{\text{Ad}_X \boldsymbol{\tau}}{\boldsymbol{\tau}} \\
    &= \text{Ad}_X.
\end{split}
\end{align}
Thus
\begin{equation}
    \frac{d}{dt} X(t) Y(t) = \frac{dX}{dt} + \text{Ad}_X \frac{dY}{dt}.
    \label{eq:lie_prod_rule}
\end{equation}
Now we can take the first time derivative of~\eqref{eq:spl_recurs}.
\begin{equation}
\begin{split}
    \frac{d}{dt} g^{(0)} &= 0, \\
    \frac{d}{dt} g^{(j)} &= \frac{d}{dt} g^{(j-1)} + \text{Ad}_{g^{(j-1)}} \frac{d}{dt} \mathbf{A}_j(u).
\end{split}
\end{equation}

\subsubsection{Second Derivative}

We will need
\begin{equation}
    \frac{d^2}{dt^2} \mathbf{A}_j(u) = \ddot{b}_j(u) \boldsymbol{\Omega}_{n+j-k+1}.
\end{equation}
We will additionally need the derivative of the adjoint. Given $X \in G, \mathbf{p} \in \mathbb{R}^d$,
\begin{equation}
\begin{split}
    \frac{\partial}{\partial X} \text{Ad}_X \mathbf{p} &= \lim_{\boldsymbol{\tau} \to 0} \frac{\text{Ad}_{\text{Exp}(\boldsymbol{\tau})X} \mathbf{p} - \text{Ad}_{X} \mathbf{p}}{\boldsymbol{\tau}} \\
    &= \lim_{\boldsymbol{\tau} \to 0} \frac{\text{exp}(\boldsymbol{\tau}^\curlywedge) \text{Ad}_X \mathbf{p} - \text{Ad}_{X} \mathbf{p}}{\boldsymbol{\tau}} \\
    &= \lim_{\boldsymbol{\tau} \to 0} \frac{(\mathbf{I} + \boldsymbol{\tau}^\curlywedge) \text{Ad}_X \mathbf{p} - \text{Ad}_{X} \mathbf{p}}{\boldsymbol{\tau}} \\
    &= -(\text{Ad}_X \mathbf{p})^\curlywedge.
    \label{eq:adjoint_deriv}
\end{split}
\end{equation}
Here we have used the identities
\begin{gather*}
    \text{Ad}_{XY} = \text{Ad}_X \text{Ad}_Y, \\
    \text{Ad}_{\text{Exp}(\boldsymbol{\tau})} = \text{exp}(\boldsymbol{\tau}^\curlywedge).
\end{gather*}
We also have
\begin{equation}
    \frac{\partial}{\partial \mathbf{p}} \text{Ad}_X \mathbf{p} = \text{Ad}_X.
\end{equation}

We can finally take the second derivative of \eqref{eq:spl_recurs}:
\begin{equation}
\begin{split}
    \frac{d^2}{dt^2} g^{(0)} &= 0, \\
    \frac{d^2}{dt^2} g^{(j)} &= \frac{d^2}{dt^2} g^{(j-1)} - \left(\text{Ad}_{g^{(j-1)}} \frac{d}{dt} \mathbf{A}_j(u)\right)^\curlywedge \frac{d}{dt} g^{(j-1)}\\
    &\quad+ \text{Ad}_{g^{(j-1)}} \frac{d^2}{dt^2} \mathbf{A}_j(u).
\end{split}
\label{eq:d2_unsimp}
\end{equation}
This can be simplified by noting that 
\begin{equation}
    \text{Ad}_{g^{(j-1)}} \frac{d}{dt} \mathbf{A}_j(u) = \frac{d}{dt} g^{(j)} - \frac{d}{dt} g^{(j-1)}
    \label{eq:ad_T_adot}
\end{equation}
and using the fact that, from the alternating property of Lie algebras, for $a \in \mathbb{R}^d \cong \text{ad}(\mathfrak{g})$ (noting that $\text{ad}(\mathfrak{g})$ is itself a Lie algebra),
\begin{equation}
    a^\curlywedge a = 0,
\end{equation}
thus~\eqref{eq:d2_unsimp} simplifies to 
\begin{equation}
\begin{split}
    \frac{d^2}{dt^2} g^{(0)} &= 0, \\
    \frac{d^2}{dt^2} g^{(j)} &= \frac{d^2}{dt^2} g^{(j-1)} - \left(\frac{d}{dt} g^{(j)}\right)^\curlywedge \frac{d}{dt} g^{(j-1)}\\
    &\quad+ \text{Ad}_{g^{(j-1)}} \frac{d^2}{dt^2} \mathbf{A}_j(u).
\end{split}
\end{equation}

\subsection{Lie Group Spline Jacobians}
\label{sec:spl_jacs}

In order to perform optimization on Lie Group splines, we will need to compute the Jacobians of~\eqref{eq:spl_lie} and its derivatives with respect to the control points. We show how these are computed in this section, and note that these Jacobians make no approximations.

\subsubsection{Zeroth Derivative Jacobians}

We seek to calculate the Jacobians $\frac{\partial}{\partial \bar{g}_l} g(t)$ for all values of $l$. We first note that due to the local support property of B-splines, if $l < n - k + 1$ or $l > n$, $\frac{\partial}{\partial \bar{g}_l} g(t) = 0$. Otherwise, we have
\begin{equation}
\begin{split}
    \frac{\partial}{\partial \bar{g}_l} g^{(0)} &= \frac{\partial}{\partial \bar{g}_l} \bar{g}_{n-k+1} \qquad (\mathbf{I}\text{ if } l = n-k+1, \: 0 \text{ else)} \\
    \frac{\partial}{\partial \bar{g}_l} g^{(j)} &= \frac{\partial}{\partial \bar{g}_l} g^{(j-1)} + \text{Ad}_{g^{(j-1)}} \frac{\partial}{\partial \bar{g}_l} \mathbf{A}_j
\end{split}
\end{equation}
using~\eqref{eq:lie_prod_rule}. We need to calculate $\frac{\partial}{\partial \bar{g}_l} \mathbf{A}_j = \frac{\partial \mathbf{A}_j}{\partial \boldsymbol{\Omega}_{n+j-k+1}} \frac{\partial \boldsymbol{\Omega}_{n+j-k+1}}{\partial \bar{g}_l}$. First,
\begin{equation}
\begin{split}
    \frac{\partial \mathbf{A}_j}{\partial \boldsymbol{\Omega}} &= \lim_{\boldsymbol{\tau} \to 0} \frac{\text{Log}\left( \text{Exp}(b_j(\boldsymbol{\Omega} + \boldsymbol{\tau})) \text{Exp}(b_j \boldsymbol{\Omega})^{-1} \right)}{\boldsymbol{\tau}} \\
    &= \lim_{\boldsymbol{\tau} \to 0} \frac{\text{Log}\left( \text{Exp}(b_j J_l(b_j \boldsymbol{\Omega})\boldsymbol{\tau})\text{Exp}(b_j\boldsymbol{\Omega}) \text{Exp}(b_j \boldsymbol{\Omega})^{-1} \right)}{\boldsymbol{\tau}} \\ 
    &= b_j J_l(b_j \boldsymbol{\Omega}),
\end{split}
\end{equation}
where we have omitted the subscript on $\boldsymbol{\Omega}$ for conciseness. Next, there are three cases for $\frac{\partial \boldsymbol{\Omega}_{n+j-k+1}}{\partial \bar{g}_l}$.

Case 1, $l = n+j-k$:
\begin{equation}
\begin{split}
    \frac{\partial}{\partial \bar{g}_l} \boldsymbol{\Omega}_{l+1} &= \lim_{\boldsymbol{\tau} \to 0} \frac{\text{Log}\left( (\text{Exp}(\boldsymbol{\tau}) \bar{g}_l)^{-1} \bar{g}_{l+1} \right) - \boldsymbol{\Omega}_{l+1}}{\boldsymbol{\tau}} \\
    &= \lim_{\boldsymbol{\tau} \to 0} \frac{\text{Log}\left( \bar{g}_l^{-1} \text{Exp}(-\boldsymbol{\tau}) \bar{g}_{l+1} \right) - \boldsymbol{\Omega}_{l+1}}{\boldsymbol{\tau}} \\
    &= \lim_{\boldsymbol{\tau} \to 0} \frac{\text{Log}\left( \text{Exp}(- \text{Ad}_{\bar{g}_l^{-1}} \boldsymbol{\tau}) \bar{g}_l^{-1} \bar{g}_{l+1} \right) - \boldsymbol{\Omega}_{l+1}}{\boldsymbol{\tau}} \\
    &= \lim_{\boldsymbol{\tau} \to 0} \frac{-J_l^{-1}(\boldsymbol{\Omega}_{l+1}) \text{Ad}_{\bar{g}_l^{-1}} \boldsymbol{\tau} + \boldsymbol{\Omega}_{l+1} - \boldsymbol{\Omega}_{l+1}}{\boldsymbol{\tau}} \\
    &= -J_l^{-1}(\boldsymbol{\Omega}_{l+1}) \text{Ad}_{\bar{g}_l^{-1}}.
\end{split}
\end{equation}
Case 2, $l = n+j-k+1$:
\begin{equation}
\begin{split}
    \frac{\partial}{\partial \bar{g}_l} \boldsymbol{\Omega}_{l} &= \lim_{\boldsymbol{\tau} \to 0} \frac{\text{Log}\left( \bar{g}_{l-1}^{-1} \text{Exp}(\boldsymbol{\tau}) \bar{g}_{l} \right) - \boldsymbol{\Omega}_{l}}{\boldsymbol{\tau}} \\
    &= J_l^{-1}(\boldsymbol{\Omega}_{l}) \text{Ad}_{\bar{g}_{l-1}^{-1}}.
\end{split}
\end{equation}
Case 3, otherwise:
\begin{equation}
    \frac{\partial}{\partial \bar{g}_l} \boldsymbol{\Omega} = 0.
\end{equation}
Hence
\begin{equation}
    \frac{\partial}{\partial \bar{g}_l} \boldsymbol{\Omega}_{n+j-k+1} = \begin{cases}
    -J_l^{-1}(\boldsymbol{\Omega}) \text{Ad}_{\bar{g}_l^{-1}}, & l = n+j-k \\
    J_l^{-1}(\boldsymbol{\Omega}) \text{Ad}_{\bar{g}_{l-1}^{-1}}, & l = n+j-k+1 \\
    0 & \text{else}.
    \end{cases}
    \label{eq:om_struct}
\end{equation}

\subsubsection{First Derivative Jacobians}

We will first need $\frac{\partial}{\partial \bar{g}_l} \dot{\mathbf{A}}_j$. This is trivial given \eqref{eq:om_struct}:
\begin{equation}
    \frac{\partial}{\partial \bar{g}_l} \dot{\mathbf{A}}_j = \dot{b}_j \frac{\partial}{\partial \bar{g}_l} \boldsymbol{\Omega}_{n+j-k+1}.
\end{equation}
Then, applying~\eqref{eq:adjoint_deriv} and~\eqref{eq:ad_T_adot},
\begin{equation}
\begin{split}
    \frac{\partial}{\partial \bar{g}_l} \dot{g}^{(0)} &= 0 \\
    \frac{\partial}{\partial \bar{g}_l} \dot{g}^{(j)} &= \frac{\partial}{\partial \bar{g}_l} \dot{g}^{(j-1)} - \left( \dot{g}^{(j)} - \dot{g}^{(j-1)} \right)^\curlywedge \frac{\partial}{\partial \bar{g}_l} g^{(j-1)} \\
    &\quad+ \text{Ad}_{g^{(j-1)}} \frac{\partial}{\partial \bar{g}_l} \dot{\mathbf{A}}_j.
\end{split}
\end{equation}

\subsubsection{Second Derivative Jacobians}

We will need the Jacobians of $\mathbf{p}^\curlywedge \mathbf{v}$ with respect to $\mathbf{p} \in \mathbb{R}^d$ and $\mathbf{v} \in \mathbb{R}^d$:
\begin{align}
    \frac{\partial}{\partial \mathbf{v}} \left(\mathbf{p}^\curlywedge \mathbf{v}\right) &= \mathbf{p}^\curlywedge \\ 
    \frac{\partial}{\partial \mathbf{p}} \left(\mathbf{p}^\curlywedge \mathbf{v}\right) &= \frac{\partial}{\partial \mathbf{p}} \left(-\mathbf{v}^\curlywedge \mathbf{p}\right) = -\mathbf{v}^\curlywedge
\end{align}
Then, for arbitrary $Y$ in any space,
\begin{equation}
    \frac{\partial}{\partial Y} \left[ \mathbf{p}(Y)^\curlywedge \mathbf{v}(Y) \right] =  -\mathbf{v}^\curlywedge \frac{\partial \mathbf{p}}{\partial Y} + \mathbf{p}^\curlywedge \frac{\partial \mathbf{v}}{\partial Y}.
\end{equation}

We will also need $\frac{\partial}{\partial \bar{g}_l} \ddot{\mathbf{A}}_j$:
\begin{equation}
    \frac{\partial}{\partial \bar{g}_l} \ddot{\mathbf{A}}_j = \ddot{b}_j \frac{\partial}{\partial \bar{g}_l} \boldsymbol{\Omega}_{n+j-k+1}.
\end{equation}

Finally,
\begin{equation}
\begin{split}
    \frac{\partial}{\partial \bar{g}_l} \ddot{g}^{(0)} &= 0 \\
    \frac{\partial}{\partial \bar{g}_l} \ddot{g}^{(j)} &= \frac{\partial}{\partial \bar{g}_l} \ddot{g}^{(j-1)} - \left( \dot{g}^{(j)} \right)^\curlywedge \frac{\partial}{\partial \bar{g}_l} \dot{g}^{(j-1)} \\
    &\quad+ \left( \dot{g}^{(j-1)} \right)^\curlywedge \frac{\partial}{\partial \bar{g}_l} \dot{g}^{(j)} \\
    &\quad- \left( \text{Ad}_{g^{(j-1)}} \ddot{\mathbf{A}}_j \right)^\curlywedge \frac{\partial}{\partial \bar{g}_l} g^{(j-1)} \\
    &\quad+ \text{Ad}_{g^{(j-1)}} \frac{\partial}{\partial \bar{g}_l} \ddot{\mathbf{A}}_j.
\end{split}
\end{equation}
        \subsection{Lie Group Motion Prior Jacobians}
\label{sec:mp_jacs}

We need to compute the Jacobians of $\mathbf{r}_{j, j+1} = m(\mathbf{x}(t_j^\prime), \mathbf{x}(t_{j+1}^\prime))$ w.r.t. $\mathbf{x}(t_j^\prime)$, $\mathbf{x}(t_{j+1}^\prime)$ for Lie groups. Everything we need to derive these Jacobians was presented in Appendix~\ref{sec:lie_gp_derivatives}. Let $g_j = g(t_j^\prime)$, $\dot{g}_j = \dot{g}(t_j^\prime)$, and $\ddot{g}_j = \ddot{g}(t_j^\prime)$. We have
\begin{equation}
    \frac{\partial \mathbf{r}_{j, j+1}}{\partial g_j} = \begin{bmatrix} \frac{\partial \boldsymbol{\xi}_j(t_{j+1}^\prime)}{\partial g_j} \\ \frac{\partial \dot{\boldsymbol{\xi}}_j(t_{j+1}^\prime)}{\partial g_j} \\ \frac{1}{2} \ddot{g}_{j+1}^\curlywedge \frac{\partial \boldsymbol{\xi}_j(t_{j+1}^\prime)}{\partial g_j} + \frac{1}{2} \dot{g}_{j+1}^\curlywedge \frac{\partial \dot{\boldsymbol{\xi}}_j(t_{j+1}^\prime)}{\partial g_j} \end{bmatrix}.
\end{equation}
Similarly for $\frac{\partial \mathbf{r}_{j, j+1}}{\partial g_{j+1}}$, and
\begin{equation}
\begin{gathered}
    \frac{\partial \mathbf{r}_{j, j+1}}{\partial \dot{\bar{g}}_j} = \begin{bmatrix}
    -\Delta t \mathbf{I} \\ -\mathbf{I} \\ \mathbf{0} \end{bmatrix}, \quad
    \frac{\partial \mathbf{r}_{j, j+1}}{\partial \ddot{\bar{g}}_j} = \begin{bmatrix} -\frac{1}{2}\Delta t^2 \mathbf{I} \\ -\Delta t \mathbf{I} \\ -\mathbf{I} \end{bmatrix}, \\
    \frac{\partial \mathbf{r}_{j, j+1}}{\partial \dot{g}_{j+1}} = \begin{bmatrix} \mathbf{0} \\ J_l^{-1}(\boldsymbol{\xi}_j(t_{j+1}^\prime)) \\ -\frac{1}{2} \left( \dot{\boldsymbol{\xi}}_j(t_{j+1}^\prime)^\curlywedge - \dot{g}_{j+1}^\curlywedge J_l^{-1}(\boldsymbol{\xi}_j(t_{j+1}^\prime)) \right) \end{bmatrix}, \\
    \frac{\partial \mathbf{r}_{j, j+1}}{\partial \ddot{g}_{j+1}} = \begin{bmatrix} \mathbf{0} \\ \mathbf{0} \\ J_l^{-1}(\boldsymbol{\xi}_j(t_{j+1}^\prime)) \end{bmatrix}.
\end{gathered}
\end{equation}
    \end{appendix}

    \bibliographystyle{IEEEtran}
    \bibliography{refs}

\end{document}